\documentclass[11pt, a4paper, onecolumn]{article}

\usepackage{amsmath,amssymb}
\usepackage{bm}
\usepackage{graphicx}
\usepackage{algorithm,algorithmic}
\usepackage{color}
\usepackage{url}
\usepackage[pagebackref=true,breaklinks=true,colorlinks,bookmarks=false]{hyperref}
\usepackage{cite}
\usepackage[whole]{bxcjkjatype}
\usepackage[margin=40mm]{geometry}
\usepackage{stmaryrd}
\usepackage[]{hyperref}
\hypersetup{colorlinks=true,}

\newcommand{\vect}[1]{\mathbf{#1}}
\newcommand{\mat}[1]{\mathbf{#1}}
\newcommand{\ten}[1]{\boldsymbol{\mathcal{#1}}}
\newcommand{\bbR}[1]{\mathbb{R}^{#1}}
\newcommand{\perm}{\text{permute}}
\newcommand{\unvec}{\text{vec}}
\newcommand{\fold}{\text{fold}}
\newcommand{\tr}{\text{trace}}
\newcommand{\diag}{\text{diag}}
\newcommand{\argmin}{\mathop{\text{argmin}}}

\newcommand{\minimize}{\mathop{\text{minimize}}}

\newcommand{\rank}{\text{rank}}

\newcommand{\inp}[2]{\left\langle #1, #2 \right\rangle}
\newcommand{\ang}[1]{\left\langle #1 \right\rangle}

\newcommand{\llangle}{\left\langle\hspace{-2.5pt}\left\langle}
\newcommand{\rrangle}{\right\rangle\hspace{-2.5pt}\right\rangle}

\begin{document}

\title{Very Basics of Tensors with Graphical Notations: Unfolding, Calculations, and Decompositions}
\author{Tatsuya Yokota
\thanks{Tatsuya Yokota is with Nagoya Institute of Technology, Japan and with RIKEN Center for Advanced Intelligence Project, Japan. e-mail: t.yokota@nitech.ac.jp.}}

\maketitle

\hypersetup{linkcolor=blue}
\tableofcontents

\newpage

\section{Introduction}
Tensor network diagram (graphical notation) is a useful tool that graphically represents multiplications between multiple tensors using nodes and edges. Using the graphical notation, complex multiplications between tensors can be described simply and intuitively, and it also helps to understand the essence of tensor products.
In fact, most of matrix/tensor products including inner product, outer product, Hadamard product, Kronecker product, and Khatri-Rao product can be written in graphical notation.
These matrix/tensor operations are essential building blocks for the use of matrix/tensor decompositions in signal processing and machine learning.

Although both the graphical notation by Penrose \cite{penrose1971applications} and the original methods of CP decomposition by Harshman, Carroll, and Chang \cite{harshman1970foundations,carroll1970analysis} have been proposed in the 1970s, graphical notation has been used for tensor decompositions since the 2010s \cite{holtz2012alternating}.
At least, the graphical notation was not used in the review papers of tensor decompositions \cite{kolda2009tensor,cichocki2015tensor,sidiropoulos2017tensor}.
Until the 2010s, the main tensor decomposition models used in signal processing and machine learning were relatively simple models such as CP and Tucker decompositions, then there would have been no big problems.
However, it seems that things started to change around this time.
Tensor networks, which have been used in quantum physics \cite{ran2020tensor,taylor2024introduction,fernandez2024learning}, have been applied to tensor decompositions in signal processing and machine learning \cite{oseledets2011tensor,holtz2012alternating,zhao2016tensor,stoudenmire2016supervised,cichocki2016tensor,cichocki2017tensor,novikov2015tensorizing,novikov2021tensor,amiridi2022lowI,amiridi2022lowII,sengupta2022tensor,memmel2024position}.

\begin{figure}[p]
    \centering
    \includegraphics[width=0.99\textwidth]{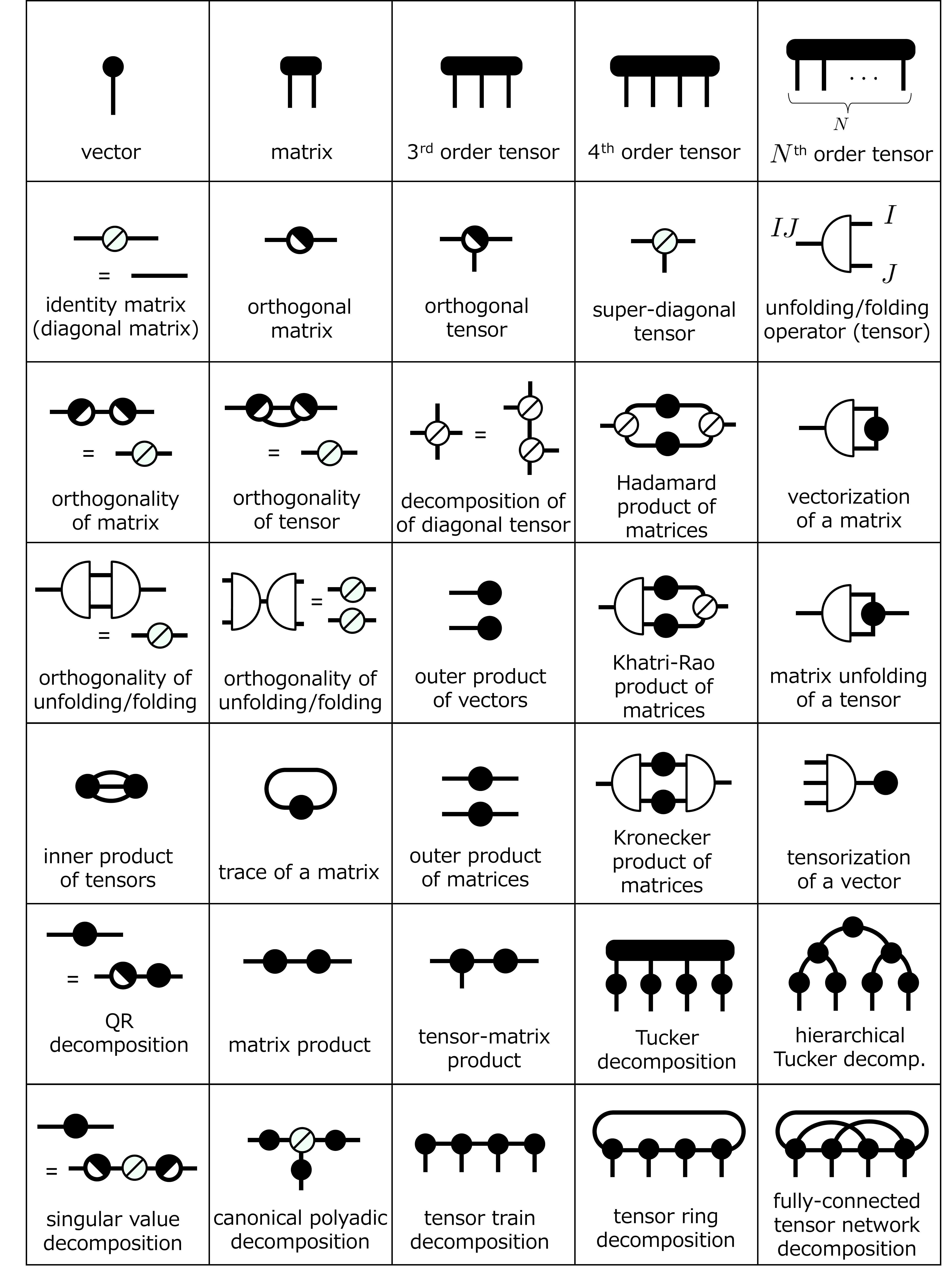}
    \caption{List of tensor network diagrams.}\label{fig:tensor_network_diagrams}
\end{figure}

Although tensor decomposition is a very promising technique, understanding its building blocks can seem like a relatively long journey for a beginner.
However, we now have tensor network diagrams, and learning tensor decompositions with network diagrams can significantly reduce the journey.
The purpose of this lecture note is to learn the very basics of tensors and how to represent them in mathematical symbols and graphical notation.  For example, one of the goals is to allow the reader to understand some specific mathematical symbols for the tensors included in Table~\ref{tab:symbols}, and the graphical notation shown in Figure~\ref{fig:tensor_network_diagrams}.
Many papers using tensors omit these detailed definitions and explanations, which can be difficult for the reader.
I hope this note will be of help to such readers.

\begin{table}[p]
\caption{List of symbols} \label{tab:symbols}
\centering
\begin{tabular}{ll}\hline 
symbols         & explanations \\ \hline
$\bbR{}$ & set of real numbers\\
$\bbR{I}$ & $I$-dimensional real vector space\\
$\bbR{I \times J}$ & $(I,J)$-dimensional real matrix space\\
$\bbR{I \times J \times K}$ & $(I,J,K)$-dimensional real tensor space\\
$\vect{a}$      & vector\\
$\vect{e}_i$ & one-hot vector whose $i$-th entry is 1 \\
$\mat{A}$     & matrix \\
$\ten{A}$     & tensor \\
$a_i$ or $a(i)$ & $i$-th entry of vector $\vect{a}$\\
$A_{ij}$ or $A(i,j)$ & $(i,j)$-th entry of matrix $\mat{A}$ \\
$\mathcal{A}_{ijk}$ or $\mathcal{A}(i,j,k)$ & $(i,j,k)$-th entry of tensor $\ten{A}$\\
$\ten{A}(:,i,j)$ & $(j,k)$-th column fiber of tensor $\ten{A}$ \\
$\ten{A}(i,:,k)$ & $(i,k)$-th row fiber of tensor $\ten{A}$ \\
$\ten{A}(i,j,:)$ & $(i,j)$-th tube fiber of tensor $\ten{A}$ \\
$\ten{A}(:,:,k)$ & $k$-th slice of tensor $\ten{A}$ along $(1,2)$-modes \\
$\ten{A}(:,j,:)$ & $j$-th slice of tensor $\ten{A}$ along $(1,3)$-modes \\
$\ten{A}(i,:,:)$ & $i$-th slice of tensor $\ten{A}$ along $(2,3)$-modes \\
$\mat{A}^\top$ & matrix transposition ($\mat{A}^\top=\text{permute}_{[2,1]}(\mat{A})$) \\
$\text{permute}_{\vect{p}}(\ten{A})$ & mode permutation of tensor $\ten{A}$ with order $\vect{p}$ \\
$[I]$  & set of natural numbers up to $I$ (i.e., $[I]=\{1, 2, ..., I\}$) \\
$\overline{ij}$ & $(j-1)I + i$ for $i \in [I]$ and $j \in [J]$ \\
$\overline{ijk}$ & $(k-1)IJ + (j-1)I + i$ for $i \in [I]$, $j \in [J]$ and $k \in [K]$\\
$\text{vec}(\ten{A})$ & vectorization of tensor $\ten{A}$\\
$\mat{A}_{(n)}$ & mode-$n$ unfolding of tensor $\ten{A}$\\
$\text{fold}_{(I,J)}(\vect{a})$ & matricization of vector $\vect{a} \in \mathbb{R}^{IJ}$ to a matrix of size $(I,J)$ \\
$\text{fold}_{(I,J,K)}(\vect{a})$ & tensorization of vector $\vect{a} \in \mathbb{R}^{IJK}$ to a tensor of size $(I,J,K)$ \\ 
$\boxdot$ & Hadamard product (including broadcasting option\cite{matsui2024broadcast}) \\
$\boxslash$ & entry-wise division (including broadcasting option\cite{matsui2024broadcast}) \\
$\inp{\cdot}{\cdot}$ & inner product \\
$\circ$ & outer product \\
$\otimes$ & Kronecker product \\
$\odot$ & Khatri-Rao product \\
$\times_n$ & mode product of a tensor by matrix \\
$\ten{G} \times \{\mat{A}\}$ & $\ten{G} \times_1 \mat{A}^{(1)} \times_2 \mat{A}^{(2)} \cdots \times_N \mat{A}^{(N)}$ \\
$\ten{G} \times_{-n} \{\mat{A}\}$ & $\ten{G} \times_1 \mat{A}^{(1)} \cdots \times_{n-1} \mat{A}^{(n-1)} \times_{n+1} \mat{A}^{(n+1)} \cdots \times_N \mat{A}^{(N)}$\\
$\ten{A} \mathbin{_{n}\times^{m}} \ten{B} $ & tensor-product (e.g., $\mat{W}\mat{H} = \mat{W} \mathbin{_{2}\times^{1}} \mat{H} $) \\
\hline
\end{tabular}
\end{table}

\newpage 

\section{Vectors, Matrices and Tensors}\label{sec:tensors}

\subsection{Vectors}
A {\em vector} $\vect{a} \in \bbR{I}$ is an array of numbers which $I$ scalar values $\{a_1, a_2, ..., a_I\}$ arranged in one direction: 
\begin{equation}
    \vect{a} = \begin{bmatrix} a_1 \\ a_2 \\ \vdots \\ a_I \end{bmatrix}. \label{eq:column_vector}
\end{equation}
In this note, we denote vectors by lowercase bold.
The $i$-th entry of a vector is represented as $a_i$ or $a(i)$. 
For example, $b_5$ represents the $5$th entry of the vector $\vect{b}$.
Furthermore, $a_k(i)$ represents the $i$th entry of the vector $\vect{a}_k$.

\paragraph{Column vectors and row vectors}
In most cases, vectors are often arranged vertically as \eqref{eq:column_vector}, in which case they are specifically called {\em column vectors}.
On the other hand, when they are arranged horizontally, they are called {\em row vectors}.
In this note, the column vectors will be represented by $\vect{a}$, and the row vectors will be represented by $\mathbf{a}^\top$ with transposition $\cdot^\top$.

\paragraph{One-hot vectors}
A vector with only one entry is $1$ and the rest of the entries are all $0$ is called a {\em one-hot vector}.
For example, three-dimensional one-hot vectors can be enumerated as follows:
\begin{equation}
\vect{e}_1 = \begin{bmatrix} 1 \\ 0 \\ 0 \end{bmatrix}, \ \vect{e}_2 = \begin{bmatrix} 0 \\ 1 \\ 0 \end{bmatrix}, \ \vect{e}_3 = \begin{bmatrix} 0 \\ 0 \\ 1 \end{bmatrix}. \label{eq:one_hot_vector}
\end{equation}
In this note, a one-hot vector whose $i$-th entry is 1 is denoted $\vect{e}_i$.

\subsection{Matrices}\label{sec:matrix}
A {\em matrix} $\mat{A} \in \bbR{I \times J}$ is a rectangle array which $J$ column vectors $\{\vect{a}_{1}, \vect{a}_{2}, ..., \vect{a}_{J}\}$ arranged in horizontal direction or $I$ row vectors $\{\tilde{\vect{a}}_1^\top, \tilde{\vect{a}}_2^\top, ..., \tilde{\vect{a}}_I^\top\}$ arranged in vertical direction: 
\begin{align}
    \mat{A} &= \begin{bmatrix} \ \vect{a}_1 & \ \ \vect{a}_2 & \  \cdots & \ \vect{a}_J \ \end{bmatrix} \notag \\ 
    &= \begin{bmatrix}
        A_{11} & A_{12} & \cdots & A_{1J} \\
        A_{21} & A_{22} & \cdots & A_{2J} \\
        \vdots & \vdots & \ddots & \vdots \\
        A_{I1} & A_{I2} & \cdots & A_{IJ} 
    \end{bmatrix}
    =\begin{bmatrix} \tilde{\vect{a}}_1^\top \\ \tilde{\vect{a}}_2^\top \\ \vdots \\ \tilde{\vect{a}}_I^\top \end{bmatrix}.
    \label{eq:matrix}
\end{align}
In this note, we denote matrices by bold capital letters.
$A_{ij}=A(i,j)$ represents the $(i,j)$th entry of the matrix $\mat{A}$.

When $I=J$, the matrix is called a {\em square matrix}.
When $I>J$, the matrix is called a {\em tall matrix}.
When $I<J$, the matrix is called a {\em wide matrix}.
A matrix of size $(I,1)$ can be said to be a column vector of size $I$.

\paragraph{Diagonal matrices and identity matrices}
A square matrix with all off-diagonal entries being 0 is called a {\em diagonal matrix}.
Among diagonal matrices, a matrix with all diagonal entries being $1$ is called an {\em identity matrix}:
\begin{align}
    \mat{I} = \begin{bmatrix}
        1 & 0 & \cdots & 0 \\
        0 & 1 & \cdots & 0 \\
        \vdots & \vdots & \ddots & \vdots \\
        0 & 0 & \cdots & 1 
    \end{bmatrix}.
\end{align}
Each entry of the identity matrix is {\em Kronecker delta}:
\begin{align}
I(i, j) = \delta_{ij} = \left\{ \begin{array}{ll} 1 & \text{if } i=j, \\ 0 & \text{if } i\neq j. \end{array} \right.
\end{align}
In addition, each column (and each row) of the identity matrix is a one-hot vector.
For ease of reading, an identity matrix of any size is often simply denoted $\mat{I}$, but sometimes it may be necessary to denote the identity matrix while keeping in mind the size of the matrix.
In such cases, an identity matrix of size $(N,N)$ is denoted as $\mat{I}_N$.

\paragraph{Matrix units and one-hot vectors}
When only one entry in a matrix is $1$ and the others are all $0$, it is called a {\em matrix unit}.
For example, the matrix units of size $(2,2)$ are
\begin{align}
    \begin{bmatrix}
        1 & 0 \\
        0 & 0 
    \end{bmatrix}, \ 
    \begin{bmatrix}
        0 & 0 \\
        1 & 0 
    \end{bmatrix}, \ 
    \begin{bmatrix}
        0 & 1 \\
        0 & 0 
    \end{bmatrix}, \ 
    \begin{bmatrix}
        0 & 0 \\
        0 & 1 
    \end{bmatrix}.
\end{align}
It can be thought of as a matrix version of a one-hot vector.
A simple way to enumerate matrix units of size $(I,J)$ is to reshape each column of an identity matrix $\mat{I}_{IJ}$ into a matrix of size $(I,J)$.

\begin{figure}[t]
    \centering
    \includegraphics[width=0.95\textwidth]{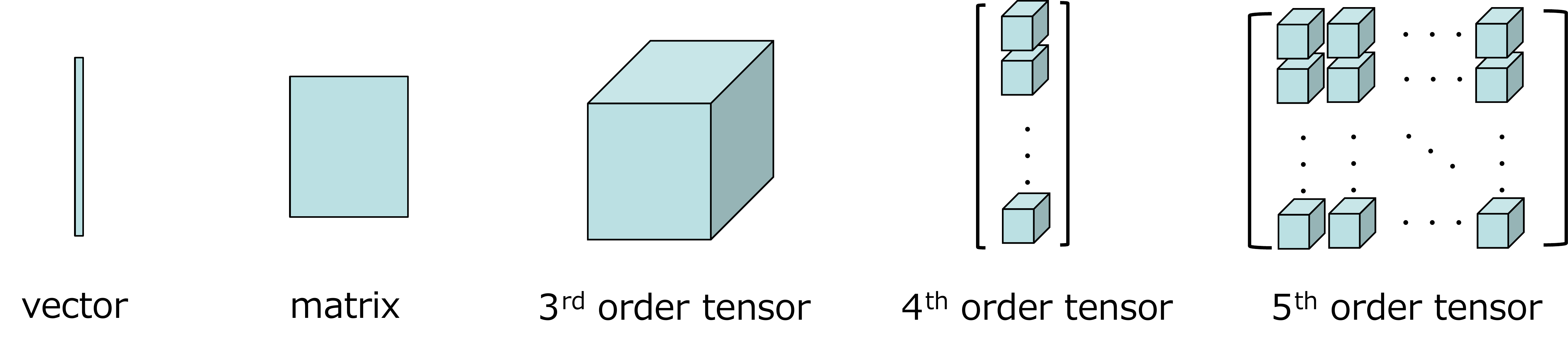}
    \vspace{-5mm}
    \caption{Tensors}\label{fig:tensor}
\end{figure}

\paragraph{All-ones matrices}
A matrix in which all entries are $1$ is called an {\em all-ones matrix} and is written as $\mat{1}$. When adding size information to the notation, it is written as $\mat{1}_{I \times J}$.
This concept applies directly to vectors and tensors such as {\em all-ones vectors} and {\em all-ones tensors}, respectively.

\subsection{Tensors}
{\em Tensors} are generalizations of vectors and matrices.
Vectors are {\em first-order tensors} and matrices are {\em second-order tensors}\footnote{In a case, vectors, matrices and third-order tensors are also called 1-way tensors，2-way tensors，and 3-way tensors, respectively.}.

A {\em third-order tensor} $\ten{A} \in \bbR{I \times J \times K}$ is a cubic array which $K$ matrices $\{\mat{A}_1, \mat{A}_2, ..., \mat{A}_K\}$ arranged in depth direction.
Similarly, arranging third-order tensors results in a fourth-order tensor, arranging fourth-order tensors results in a fifth-order tensor, and so on. 
We can recursively introduce tensors of the next higher order.
In other words, we can say that an $N$th order tensor is a $N$-dimensional array which $(N-1)$-order tensors are arranged in some direction.

\begin{figure}[t]
    \centering
    \includegraphics[width=0.8\textwidth]{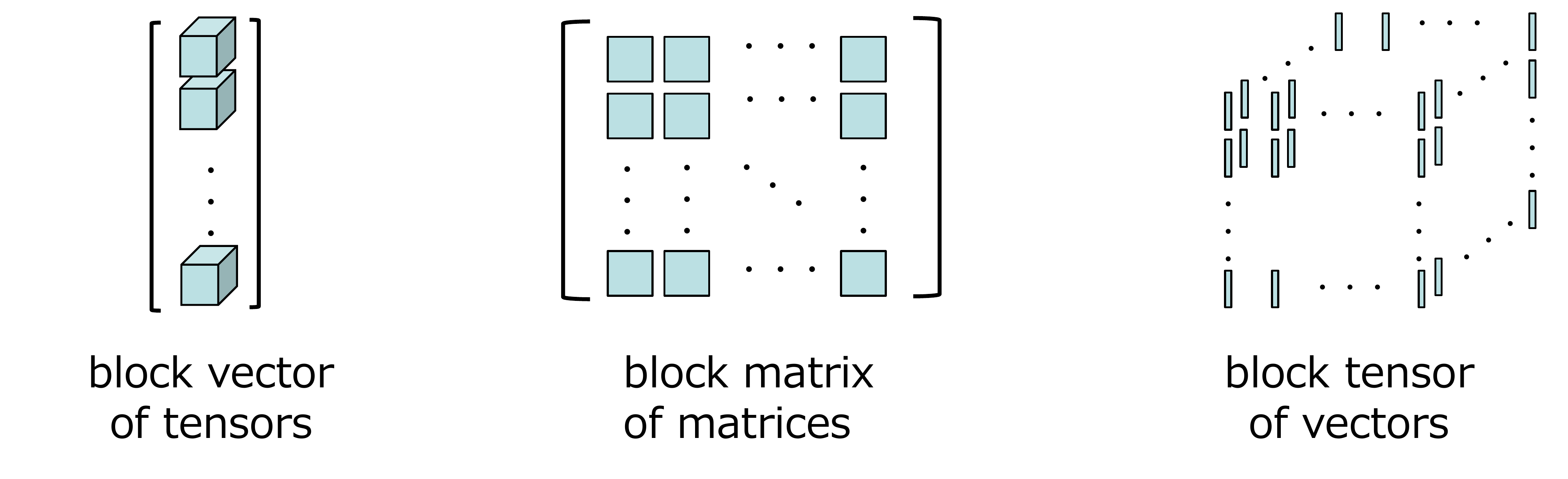}
    \vspace{-5mm}
    \caption{Different views of fourth-order tensors}\label{fig:tensor_view}
    \vspace{5mm}
    \includegraphics[width=0.9\textwidth]{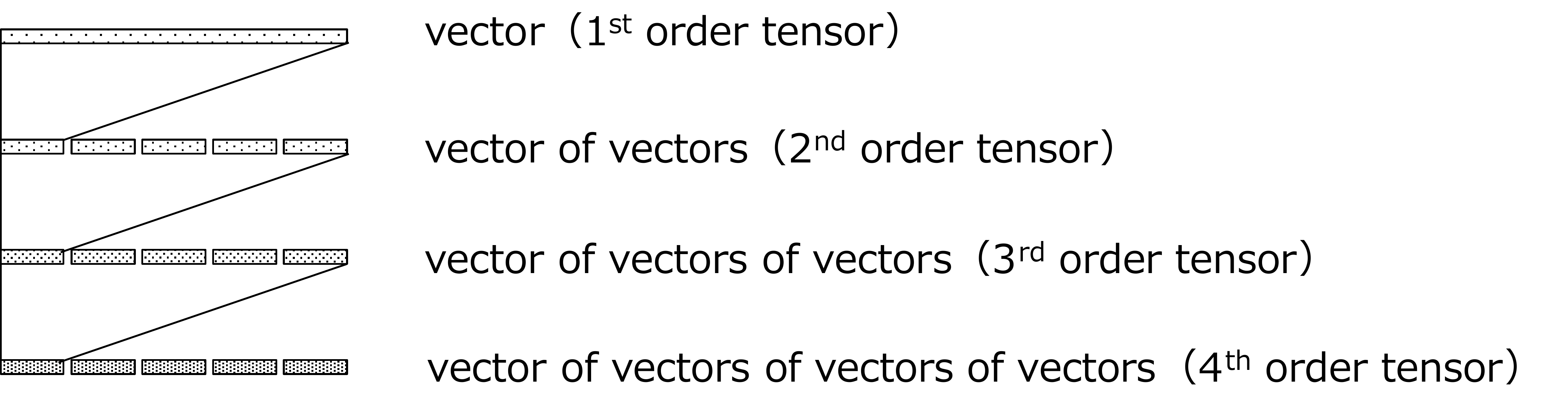}
    \vspace{-5mm}
    \caption{Hierarchical views of tensors}\label{fig:tensor_Hview}    
\end{figure}

$N$th-order tensors with $N \geq 3$ are written in bold calligraphy\footnote{Traditionally, tensors were often written with underlines $ \underline{\mathbf{A}} \in \bbR{I_1 \times I_2 \times \cdots \times I_N}$, but this is rarely seen these days. } $\ten{A} \in \bbR{I_1 \times I_2 \times \cdots \times I_N}$.
$\mathcal{A}_{i_1i_2...i_N}=\mathcal{A}(i_1,i_2,...,i_N)$ denotes the $(i_1,i_2,...,i_N)$th entry of the tensor $\ten{A}$.

Figure~\ref{fig:tensor} shows an illustration of the tensors.
A vector can be imagined as a straight line, a matrix as a rectangle, and a third-order tensor as a cube.
Furthermore, a fourth-order tensor can be imagined as an array of cubes arranged in a vector fashion, and a fifth-order tensor can be imagined as an array of cubes arranged in a matrix fashion.
In addition, there are many other ways to imagine tensors as shown in Figures~\ref{fig:tensor_view} and \ref{fig:tensor_Hview}.
If we divide the four axes of a fourth-order tensor into $1+3$, $2+2$, and $3+1$, then a fourth-order tensor can be viewed in three ways: as a block vector whose elements are tensors, a block matrix whose elements are matrices, and a block tensor whose elements are vectors (Figure~\ref{fig:tensor_view}).
If we divide the four axes of a fourth-order tensor into $1+1+1+1$, then a fourth-order tensor can be viewed as a multilevel block vectors by hierarchical loops like a vector of vectors of vectors of vectors (Figure~\ref{fig:tensor_Hview}).

\begin{figure}[t]
    \centering
    \includegraphics[width=0.5\textwidth]{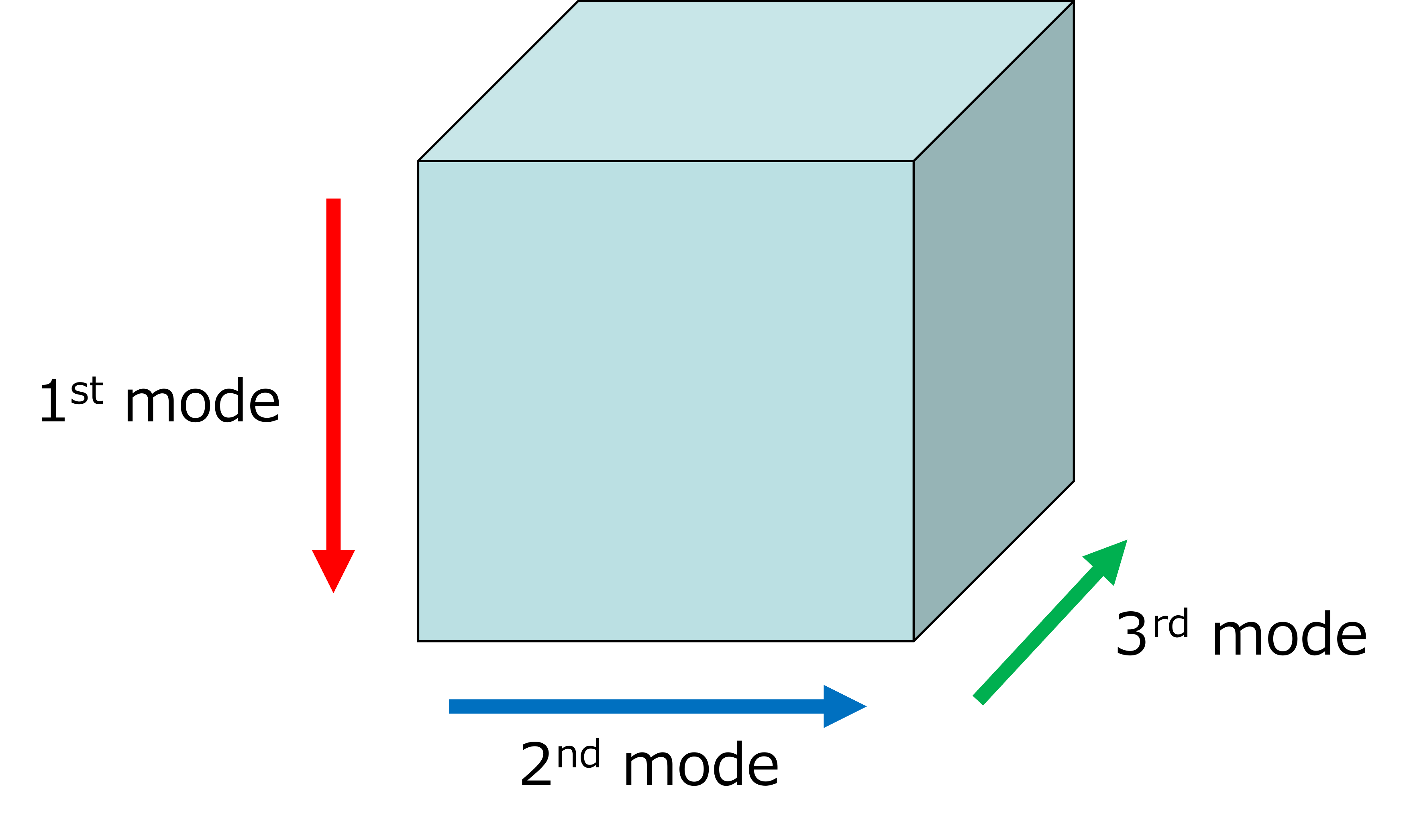}
    \caption{Modes for tensors.}\label{fig:tensor_mode}
\end{figure}

\subsection{Modes of tensors}
The directions (or axes) of the arrangement of the tensors are called the {\em modes} (Figure~\ref{fig:tensor_mode}).
The number of arranged entries is called the {\em length} or {\em size}.
For example, the vertical direction of a matrix of size $(I,J)$ is usually the {\em first mode} (length $I$), and the horizontal direction is the {\em second mode} (length $J$).
When $K$ matrices of the same size are arranged to construct a third-order tensor, the added depth direction is the {\em third mode} (length $K$).
An $N$-order tensor has $N$ modes.

When a third-order tensor is written as a formula such as $\ten{A} \in \bbR{I \times J \times K}$ without any special context, it is usually numbered from left to right, with the direction of length $I$ being called the first mode, the direction of length $J$ the second mode, and the direction of length $K$ the third mode.
In this case, there is no ranking by numbers, such as the first mode being important and the second mode being unimportant. There is no meaning other than the symbols to distinguish each one.

\subsection{Tensor network diagrams}
In Figure~\ref{fig:tensor}, we explain how to imagine tensors as lines, rectangles, cubes, and their hierarchical structures.
This image is concrete and useful for relating tensors as data arrays or any other target that you are actually trying to handle in some analysis.
However, it is still difficult to imagine higher-order tensors.
Also, it is time-consuming to draw such images of tensors on a notebook or blackboard.

\begin{figure}[t]
    \centering
    \includegraphics[width=0.95\textwidth]{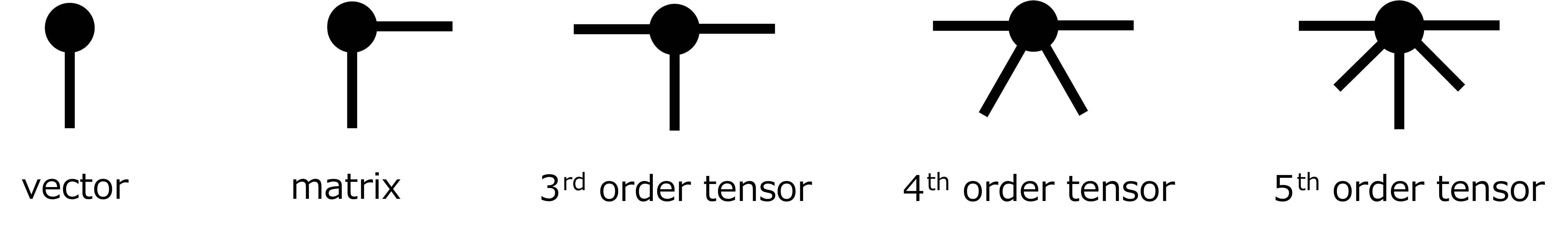}
    \caption{Tensors in graphical notation.}\label{fig:tensor_diagram}
\end{figure}

Here, we will introduce convenient notations called {\em tensor network diagrams} or {\em graphical notations}\footnote{These are also called {\em Penrose graphical notations} because these were proposed by Roger Penrose in 1971 \cite{penrose1971applications}. In this note, we refer to these simply as {\em graphical notations} or {\em diagrams}.} which allow you to imagine tensors in more abstract ways and that can be easily drawn.
In Figure~\ref{fig:tensor_diagram}, tensors of each order are shown in diagrams.
Each tensor is represented by a vertex and its modes are represented by edges that extend from the vertex.
A scalar has 0 edges, a vector has 1 edge, a matrix has 2 edges, and an $N$th-order tensor has $N$ edges\footnote{Edges are also called as legs \cite{taylor2024introduction,fernandez2024learning}.}.

\newpage

\section{Entries, Fibers, Slices, and Subtensors}

\subsection{Entries}
\begin{figure}[t]
    \centering
    \includegraphics[width=0.9\textwidth]{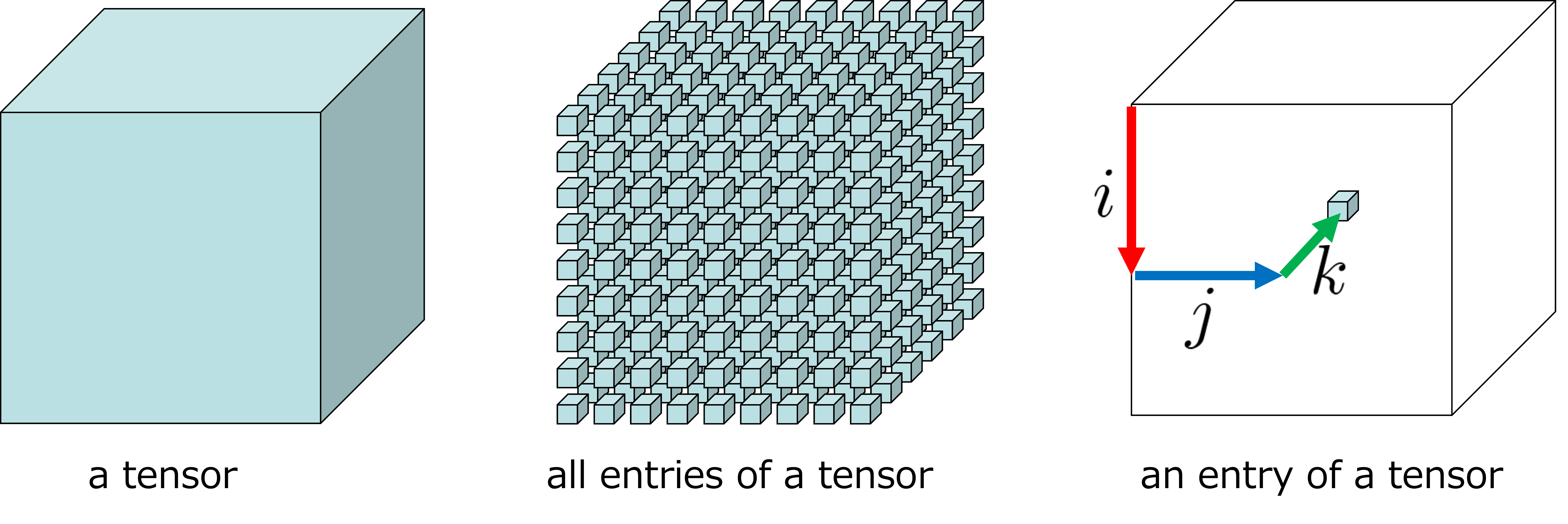}
    \vspace{-5mm}
    \caption{Entry of a third-order tensor.}\label{fig:entry}
\end{figure}

Since a tensor is a multidimensional array whose scalars are arranged in multiple directions, scalars can be said to be the smallest unit of a tensor.
Each scalar contained in a tensor is called an {\em entry} of the tensor.
Each entry is assigned an index, which is like an ``addres'' for a tensor.
For example, the location of the $(i,j,k)$-th entry of a third-order tensor $\ten{A} \in \bbR{I \times J \times K}$ is the $i$-th location along the first mode, the $j$-th location along the second mode, and the $k$-th location along the third mode.
Figure~\ref{fig:entry} shows the concept of entries for a third-order tensor.

\begin{figure}[t]
    \centering
    \vspace{5mm}
    \includegraphics[width=0.7\textwidth]{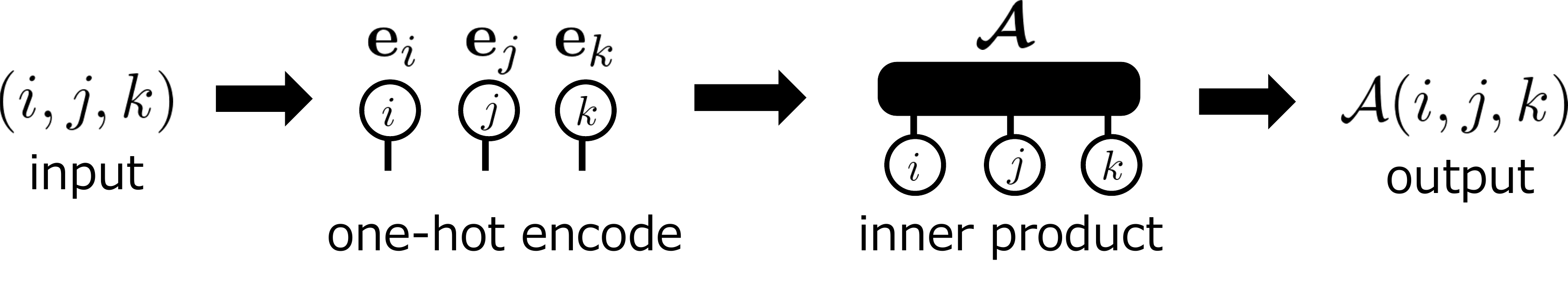}
    \vspace{-5mm}
    \caption{Entry as a function.}\label{fig:ext_entry}
\end{figure}

\paragraph{Entry as a function}
The value of the $(i,j,k)$-th entry of the third-order tensor $\ten{A} \in \bbR{I \times J \times K}$ is denoted as $\mathcal{A}(i,j,k)$ or $\mathcal{A}_{ijk}$.
This can be thought of as a function that outputs a scalar value when an index is input.
In other words, it is $\mathcal{A}: [I] \times [J] \times [K] \rightarrow \bbR{}$, where the sets of natural numbers are defined by $[C] = \{1, 2, ..., C\}$ for any positive integer $C>0$.
The third-order tensor $\ten{A}$ can be thought of as a comprehensive representation of all outputs $\mathcal{A}(i,j,k)$ for the input $(i,j,k) \in [I] \times [J] \times [K] $.

As shown in Figure~\ref{fig:ext_entry}, using the inner product $\inp{\cdot}{\cdot}$, outer product $\circ$, and one-hot vectors $(\vect{e}_i, \vect{e}_j, \vect{e}_k)$, the $(i,j,k)$-th entry of a tensor $\ten{A}$ can be represented by
\begin{align}
  \mathcal{A}(i,j,k) = \inp{\ten{A}}{\vect{e}_i \circ \vect{e}_j \circ \vect{e}_k}.
\end{align}
It can be said that a tensor-based parametric model of a continuous multivariate function studied in \cite{stoudenmire2016supervised,novikov2021tensor} can be obtained by extending this representation.

\begin{figure}[t]
    \centering
    \includegraphics[width=0.85\textwidth]{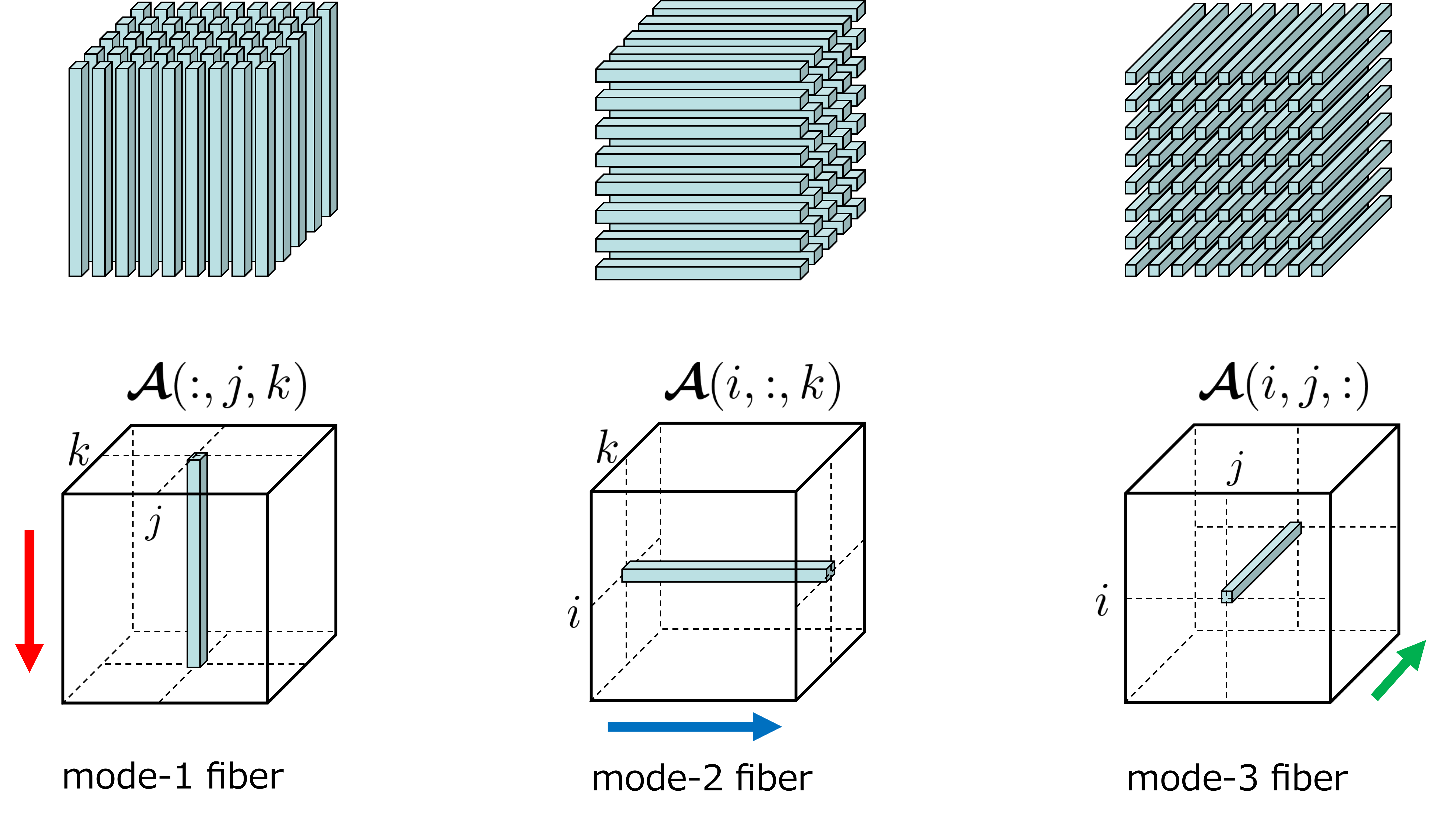}
    \caption{Fibers of third-order tensor (column, row, and tube).}\label{fig:fiber}
\end{figure}

\subsection{Fibers}
When there exists a matrix $\mat{A}=[\vect{a}_1, \vect{a}_2, ..., \vect{a}_J]=[\tilde{\vect{a}}_1, \tilde{\vect{a}}_2, ..., \tilde{\vect{a}}_I]^\top \in \bbR{I \times J}$, the vector $\vect{a}_j$ and the vector $\tilde{\vect{a}}_i$ are called {\em fibers} of matrix $\mat{A}$.
The fibers of the matrix $\mat{A}$ can be written in 
\begin{align}
\mat{A}(:,j) = \vect{a}_j \in \bbR{I}, \\
\mat{A}(i,:) = \tilde{\vect{a}}_i \in \bbR{J},
\end{align}
where the colon `` $:$ '' is used to mean like ``all''.
Unless otherwise specified, a single fiber as a formula is considered a column vector\footnote{Note that the direction may also be preserved when extracting fibers with some software. For example, in MATLAB, \texttt{A(i,:)} represents a row vector. }.

\paragraph{Columns, rows, and tubes}
For a third-order tensor $\ten{A} \in \bbR{I \times J \times K}$, we can consider three types of fiber: $\ten{A}(:,j,k)$ that follow the first mode, $\ten{A}(i,:,k)$ that follow the second mode, and $\ten{A}(i,j,:)$ that follow the third mode.
This image is shown in Figure~\ref{fig:fiber}.
In particular, fibers that follow the first mode are called {\em columns}, fibers that follow the second mode are called {\em rows}, and fibers that follow the third mode are called {\em tubes}.

\paragraph{Fiber as a function}
As in the discussion of entries, fibers can be thought of as functions that take an index as input and output a vector.
For example, a fiber along the first mode takes $(j,k)$ as input and outputs the vector $\ten{A}(:,j,k)$, so it can be expressed as $\ten{A}(:,\cdot,\cdot): [J] \times [K] \rightarrow \bbR{I}$.

\begin{figure}[t]
    \centering
    \includegraphics[width=0.85\textwidth]{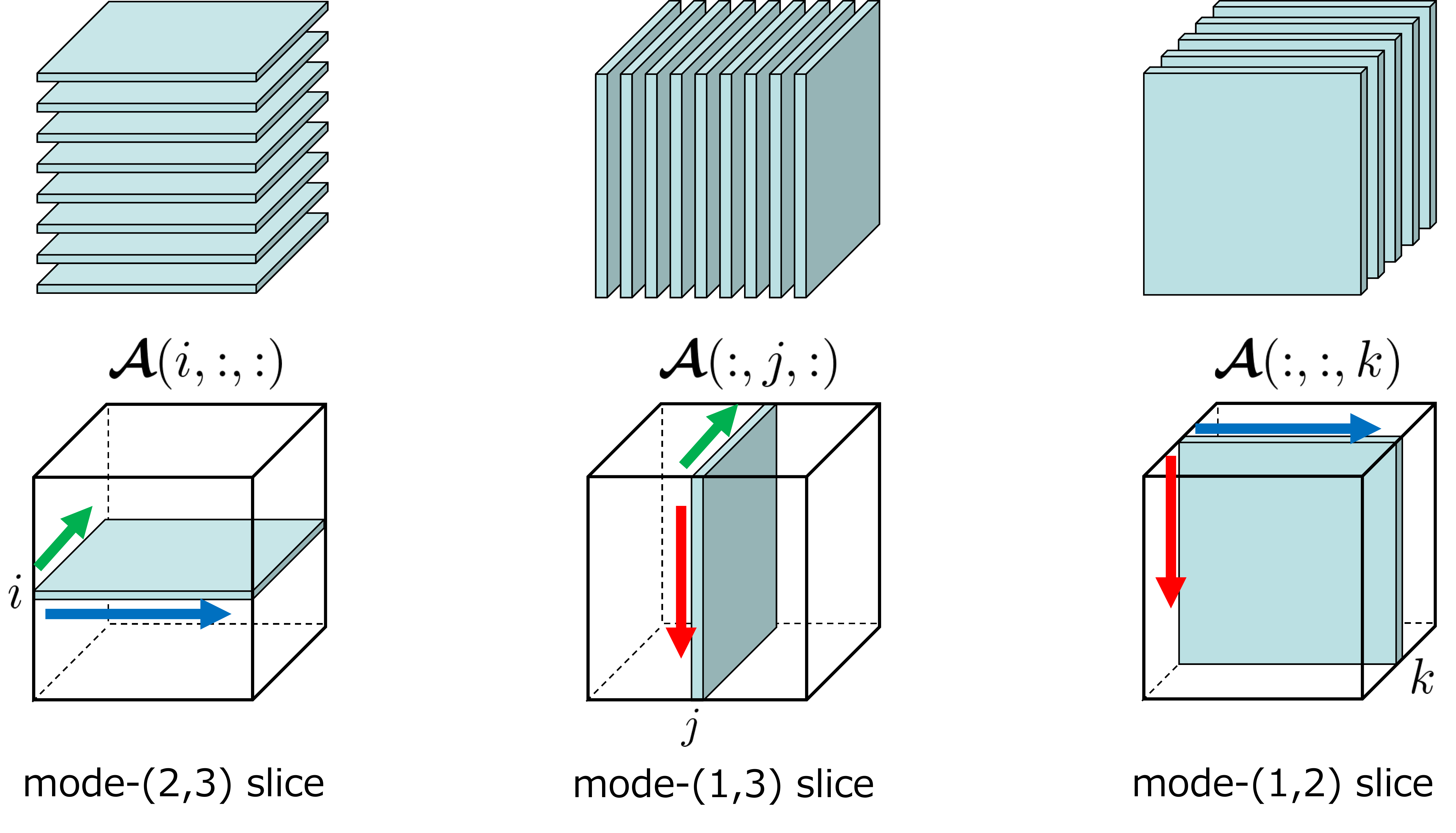}
    \caption{Slices of third-order tensor.}\label{fig:slice}
\end{figure}

\subsection{Slices}
The extension of a fiber to a matrix is called a {\em slice}.
Whereas a fiber chooses one mode and extracts entries along it, a slice chooses two modes and extracts entries along them.

For a third-order tensor $\ten{A} \in \bbR{I \times J \times K}$, we can consider three slices: $\ten{A}(:,:,k) \in \bbR{I \times J}$ along the first and second modes, $\ten{A}(:,j,:) \in \bbR{I \times K}$ along the first and third modes, and $\ten{A}(i,:,:) \in \bbR{J \times K}$ along the second and third modes.
When expressing a slice as a formula, it is a matrix of the lower mode $\times$ the other mode.
An image of a slice is shown in Figure~\ref{fig:slice}.

\paragraph{Slice as a function}
As in the discussion of entries and fibers, slices can be thought of as functions that take an index as input and output the matrix.
For example, consider the fourth-order tensor $\ten{T} \in \bbR{I_1 \times I_2 \times J \times K}$.
Since the slices along the first and second modes take $(j,k)$ as input and output the matrix $\ten{T}(:,:,j,k)$, they can be expressed as $\ten{T}(:,:,\cdot,\cdot): [J] \times [K] \rightarrow \bbR{I_1 \times I_2}$.

\paragraph{Extensions of slices (generalized slices)}
We have explained that slices are an extension of fibers, but we would like to extend this further.
For an $N$th-order tensor, we consider an integer $M$ such that $3 \leq M < N$, and by selecting $M$ modes, we extract an $M$th-order tensor.
This extracted $M$th-order tensor can be considered as an extended concept of slice.
When $M=1$, it is a fiber and when $M=2$, it is a slice.

For example, for a fourth-order tensor $\ten{T} \in \bbR{I \times J \times K \times L}$, we can consider it with $M=3$ such as
\begin{align}
\ten{T}(:,:,:,l) \in \bbR{I \times J \times K}, \notag\\
\ten{T}(:,:,k,:) \in \bbR{I \times J \times L}, \notag\\
\ten{T}(:,j,:,:) \in \bbR{I \times K \times L}, \notag\\
\ten{T}(i,:,:,:) \in \bbR{J \times K \times L}. \notag
\end{align}

\subsection{Sub-tensors}
Finally, we will introduce the {\em sub-tensors} as a generalized concept of fibers, slices, and their extensions.

In the explanations so far, the colon has been used to mean ``all,'' but this does not change the meaning if we say ``from beginning to end.''
Here, we introduce ``$m:n$'' as a usage of the colon to mean ``from $m$ to $n$.''
For example, using this notation, we can write the {\em sub-matrix} obtained by taking columns 1 to $R$ of the matrix $\mat{A} = [\vect a_1, ..., \vect a_J] \in \bbR{I \times J}$ as follows:
\begin{align}
\mat{A}(:,1:R) = \mat{A}_{:,1:R} = [\vect a_1, ..., \vect a_R] \in \bbR{I \times R}.
\end{align}

The same applies to tensors. For example, for the tensor $\ten{A} \in \bbR{I \times J \times K}$, the sub-tensor that combines the first mode $m_1$ to $n_1$, the second mode $m_2$ to $n_2$, and the third mode $m_3$ to $n_3$ can be written as follows:
\begin{align}
&\ten{A}(m_1:n_1, m_2:n_2, m_3:n_3) \notag \\ 
&\ \ = \ten{A}_{m_1:n_1, m_2:n_2, m_3:n_3} \in \bbR{(n_1-m_1+1) \times (n_2-m_2+1) \times (n_3 -m_3+1)}.
\end{align}

\newpage

\section{Reshaping Tensors}

\subsection{Mode permutation (transposition)}

\begin{figure}[t]
    \centering
    \includegraphics[width=0.6\textwidth]{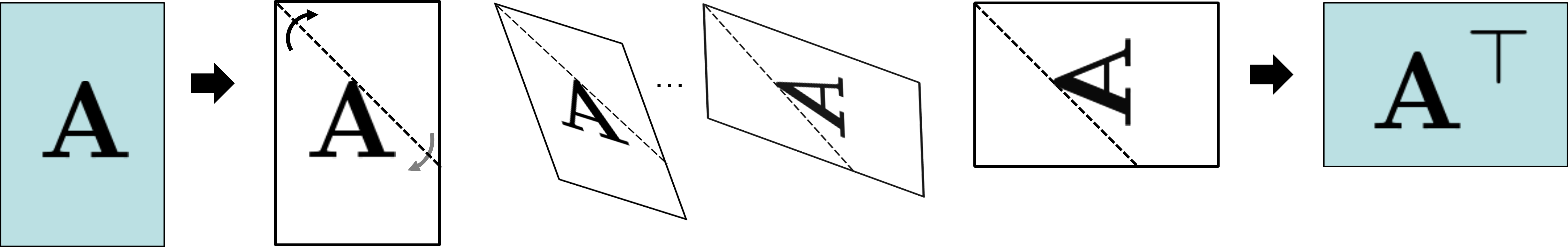}
    \caption{Image of matrix transposition.}\label{fig:matrix_transpose}
\end{figure}

\paragraph{Matrix transposition}
{\em Mode permutation} of tensor is a generalization of {\em matrix transposition}\footnote{In MATLAB, this is the \texttt{permute} function. }.
Figure~\ref{fig:matrix_transpose} shows the image of matrix transposition.
First, a matrix can be imagined as a rectangle, its transposition can be understood to flip it around a diagonal line.
The vertical direction of the rectangle becomes the horizontal direction and the horizontal direction becomes the vertical direction.
When the transposition of a matrix $\mat{A} \in \bbR{I \times J}$ is expressed as $\mat{B} = \mat{A}^\top \in \bbR{J \times I}$, we have
\begin{align}
B(j,i) = A(i,j)
\end{align}
for all $i \in [I]$ and $j \in [J]$.
In other words, the transposition of a matrix is the exchange of $i$ and $j$.
In general terms, for the second-order tensor $\mat{A}$, $i$ is the first mode and $j$ is the second mode, so the transposition of a matrix can be thought of as replacing the order of ``the first mode $\rightarrow$ the second mode'' with the order of ``the second mode $\rightarrow$ the first mode.''

\begin{figure}[t]
    \centering
    \vspace{5mm}
    \includegraphics[width=0.95\textwidth]{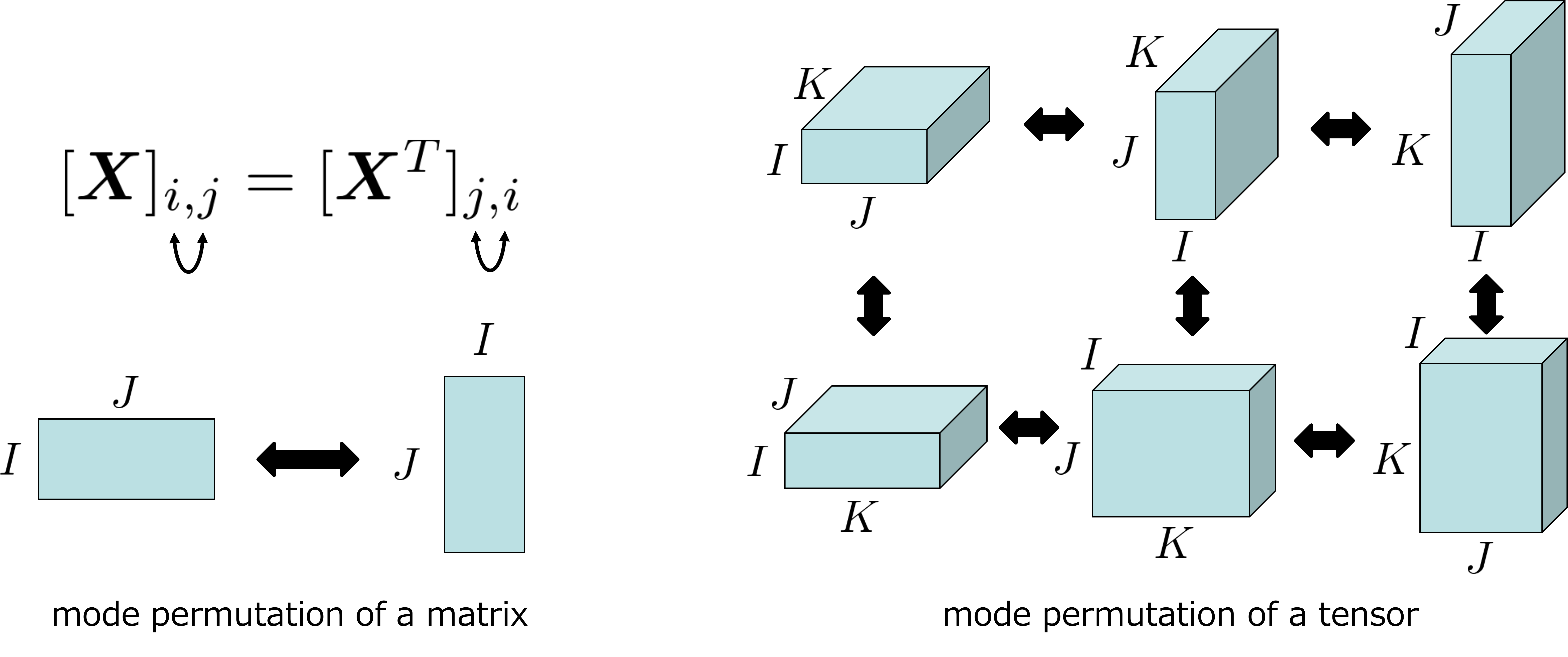}
    \vspace{-5mm}
    \caption{Image of mode permutation of tensor.}\label{fig:tensor_mode_permutation}
\end{figure}

\paragraph{Mode permutation}
Next, we consider the mode permutation of third-order tensors $\ten{A} \in \bbR{I \times J \times K}$.
Obviously, if we do nothing, the order of the modes will be ``the first mode$\rightarrow$ the second mode$\rightarrow$ the third mode'', then it expresses this as a permutation $[1,2,3]$. The mode permutation is a rearrangement of this.  For third-order tensors, we can enumerate six possible patterns: $[1,2,3]$, $[1,3,2]$, $[2,1,3]$, $[2,3,1]$, $[3,1,2]$, and $[3,2,1]$.
The sizes of the corresponding tensors are $I \times J \times K$, $I \times K \times J$, $J \times I \times K$, $J \times K \times I$, $K \times I \times J$, and $K \times J \times I$, respectively.
An illustration of this operation is shown in Figure~\ref{fig:tensor_mode_permutation}.

For $N$th-order tensors, there are $_N P_N$ possible combinations.
Let any permutation made from $\{1, 2, ..., N\}$ be $\vect{p} = [p(1), p(2), ..., p(N)]$, and denote the mode permutation operation of tensor $\ten{X} \in \bbR{I_1 \times I_2 \times \cdots \times I_N}$ as 
\begin{align}
 \ten{Y} = \perm_\vect{p}(\ten{X}).
 \end{align}
The size of the resulting tensor is $\ten{Y} \in \bbR{I_{p(1)} \times I_{p(2)} \times \cdots \times I_{p(N)}}$, and
\begin{align}
\mathcal{Y}(i_{p(1)}, i_{p(2)}, ..., i_{p(N)}) = \mathcal{X}(i_1, i_2, ..., i_N)
\end{align}
holds for all entries.

\paragraph{Inverse of mode permutation}
The inverse operation of the tensor $\ten{Y}$ to the original tensor $\ten{X}$ is written as $\ten{X} = \perm_\vect{p}^{-1}(\ten{Y})$.
For example, when performing mode permutation with $[2,3,1]$ for third-order tensor $\ten{X} \in \bbR{I \times J \times K}$, it is expressed as 
\begin{align}
\ten{Y}&=\perm_{[2,3,1]}(\ten{X}) \in \bbR{J \times K \times I}, \\
\ten{X}&=\perm_{[2,3,1]}^{-1}(\ten{Y}) = \perm_{[3,1,2]}(\ten{Y}) \in \bbR{I \times J \times K}.
\end{align}

\subsection{Vectorization}
{\em Vectorization}\footnote{In MATLAB, vectorization can be done by \texttt{a=A(:)}. } is the operation of transforming an $N$th-order tensor $\ten{A} \in \bbR{I_1 \times I_2 \times \cdots \times I_N}$ into a vector $\vect a$ of length $\prod_{n=1}^N I_n$. The entries of both correspond as follows:
\begin{equation}
a(\overline{i_1 i_2 \cdots i_N}) = \mathcal{A}(i_1, i_2, ..., i_N).
\end{equation}
Readers who understand everything from the above explanation can skip this section and move on to the next section.
From now on, we will give an explanation of vectorization from the very basics.

\paragraph{Abstract explanation of vectorization}
In Section~\ref{sec:tensors}, we explained that arranging scalars makes a vector, arranging vectors makes a matrix, arranging matrices makes a third-order tensor, and arranging third-order tensors makes a fourth-order tensor $\cdots$. We also assumed that the ``arrangement directions'' are different at each stage.
So what happens if they are all in the same direction?
First, arranging $I$ scalars makes a vector (length $I$).
Next, arranging $J$ of those vectors in the same direction\footnote{It might be better to say ``connect'' them. It is like connecting straight lines. } makes a vector, but the resulting vector is a ``long vector'' of length $IJ$.
Then, arranging $K$ of those ``long vectors'' in the same direction again makes an ``very long vector'' of length $IJK$.
Next, the ``very long vector'' becomes an ``extremely long vector''.
This is the image of vectorization of a tensor.

\paragraph{Vectorization of a matrix}
First, we consider the vectorization of a matrix.
Vectorization of a matrix is the operation of converting any $(I,J)$-matrix into a vector of length $IJ$.
For a matrix $\mat{A} = [\vect{a}_1, \vect{a}_2, ..., \vect{a}_J] \in \bbR{I \times J}$, its vectorrization is defined as follows:
\begin{align}
\unvec(\mat{A}) = \begin{bmatrix} \vect{a}_1 \\ \vect{a}_2 \\ \vdots \\ \vect{a}_J \end{bmatrix} \in \bbR{IJ}.
\end{align}
If we set $\vect{a} = \unvec(\mat{A})$, then we have 
\begin{align}
a( \overline{ij} ) = A(i,j),
\end{align}
where $\overline{ij} = (j-1)I + i$.

\begin{figure}[t]
    \centering
    \includegraphics[width=0.95\textwidth]{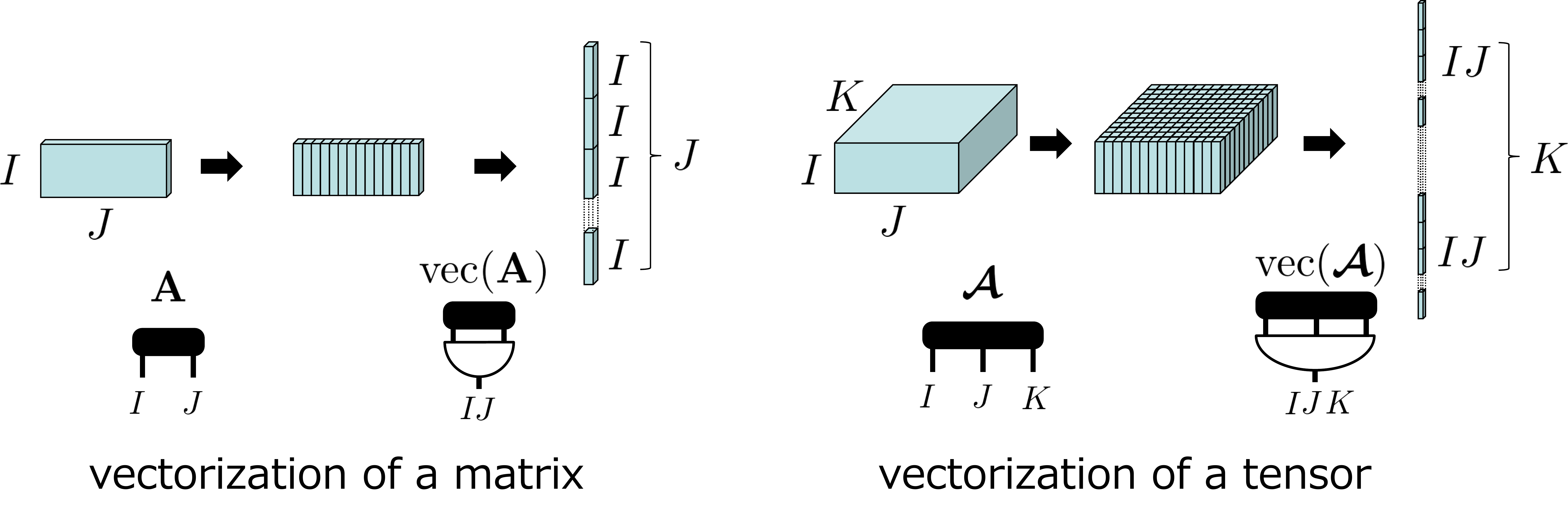}
    \vspace{-5mm}
    \caption{Vectorization of a matrix and a tensor.}\label{fig:vectorization}
\end{figure}

\begin{figure}[t]
    \centering
    \vspace{5mm}
    \includegraphics[width=0.6\textwidth]{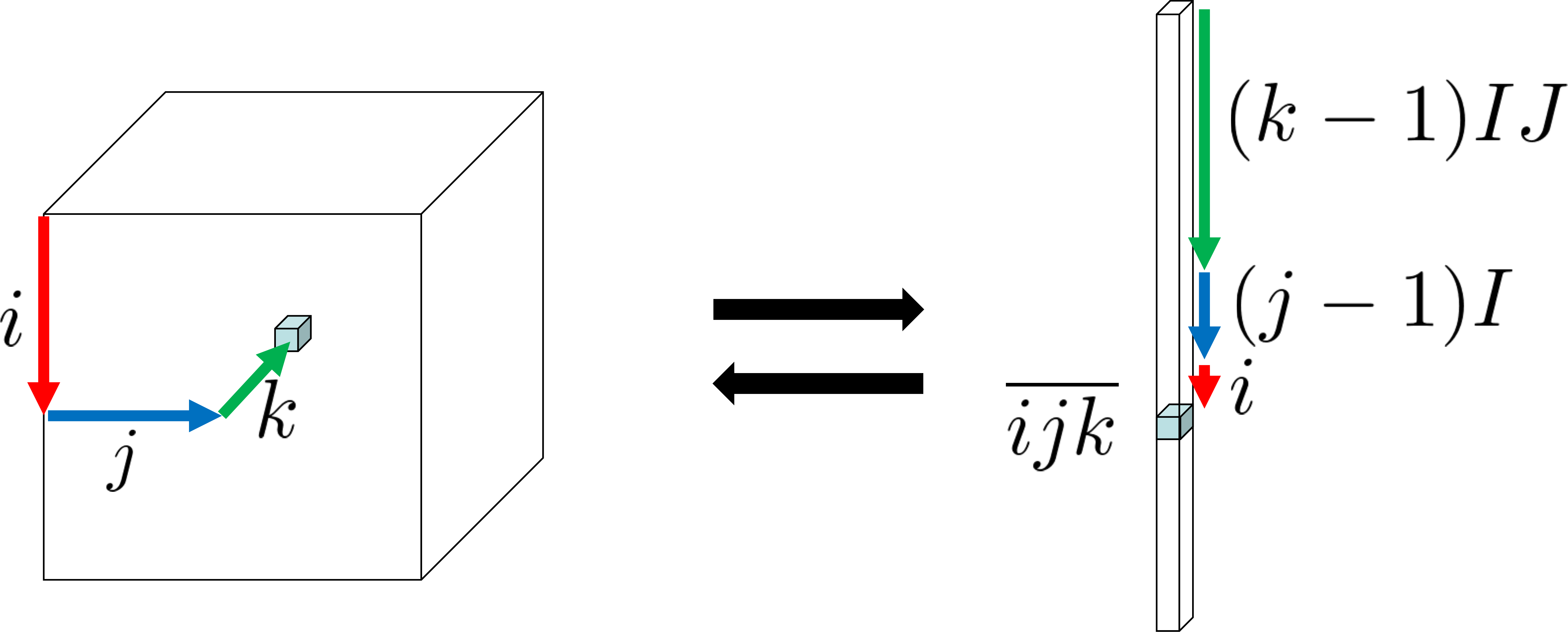}
    \vspace{-5mm}
    \caption{Entries of a tensor and its vectorization.}\label{fig:entry_of_vectorization}
\end{figure}

\paragraph{Vectorization of a third-order tensor}
Next, we consider vectorization of a third-order tensor.
It can be defined by using the vectorization of matrices.
For the third-order tensor $\ten{A} \in \bbR{I \times J \times K}$, if we enumerate all slices along the first and second modes and their vectorization, we can construct a matrix below:
\begin{align}
[\unvec(\ten{A}(:,:,1)), \cdots, \unvec(\ten{A}(:,:,K))] \in \bbR{IJ \times K}. \label{eq:unvec_slices}
\end{align}
Vectorization of a third-order tensor can be obtained by vectorizing the matrix \eqref{eq:unvec_slices} as follows:
\begin{align}
\unvec(\ten{A}) = \unvec\left( [\unvec(\ten{A}(:,:,1)), ..., \unvec(\ten{A}(:,:,K))
] \right) \in \bbR{IJK}. \label{eq:unvec_3ten}
\end{align}
If we set $\vect{a} = \unvec(\ten{A})$, then
\begin{align}
a( \overline{ijk} ) = \mathcal{A}(i,j,k)
\end{align}
holds, where $\overline{ijk} = (k-1)IJ + (j-1)I + i$.
Eq.~\eqref{eq:unvec_3ten} defines $\unvec(\cdot)$ using $\unvec(\cdot)$, which may seem strange at first glance, but strictly speaking, it is not a recursive definition because the left side $\unvec(\cdot)$ is for the third-order tensors and the right side $\unvec(\cdot)$ is for the matrices.
To avoid complicating the notation here, we use $\unvec(\cdot)$ in the same way whether the input is a matrix or a third-order tensor.

Figure~\ref{fig:vectorization} shows an image of the vectorization of a matrix and a third-order tensor.
The figure also shows diagrams for the vectorization of a matrix and a tensor.  
In this diagram, the vectorization operator is represented by a semicircular shape, and we can imagine combining multiple modes into one mode.
We can see a hierarchical structure from the inside to the outside, such as $I$, $J$, $K$.
The correspondence of the entries before and after vectorization is shown in Figure~\ref{fig:entry_of_vectorization}.

\begin{figure}[t]
    \centering
    \vspace{5mm}
    \includegraphics[width=0.99\textwidth]{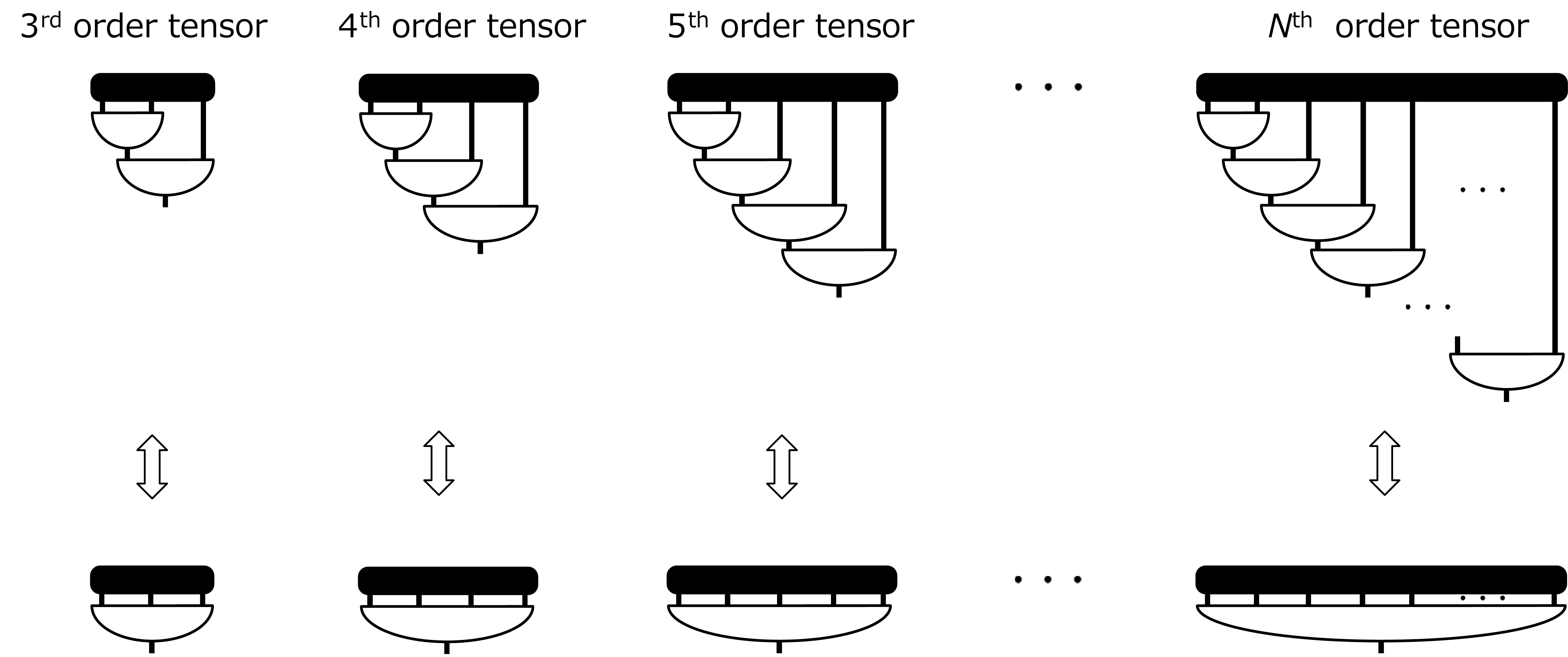}
    \vspace{-5mm}
    \caption{Vectorization of tensors.}\label{fig:vectorization_of_tensors}
\end{figure}

\paragraph{Vectorization of an $N$th-order tensor}
Finally, we consider vectorization of an $N$th-order tensor.
For any $N$th-order tensor $\ten{T} \in \bbR{I_1 \times I_2 \times \cdots \times I_N}$, consider a ``generalized slice'' that becomes an $(N-1)$th-order tensor:
\begin{align}
\ten{T}_{i_N} = \ten{T}(:,:,...,:,i_N) \in \bbR{I_1 \times I_2 \times \cdots \times I_{N-1}}.
\end{align}
Using this, the vectorization of an $N$th-order tensor ($2 < N$) can be defined as follows:
\begin{align}
\unvec(\ten{T}) = \unvec( [\unvec(\ten{T}_{1}), ..., \unvec(\ten{T}_{I_N})] ) \in \bbR{\prod_{n=1}^N I_n}. \label{eq:unvec_tensor}
\end{align}
Although it is not written explicitly, Eq.~\eqref{eq:unvec_tensor} defines the vectorization of an $N$th-order tensor using the inner vectorization for $(N-1)$th-order tensors and the outer vectorization of a matrix.
When $N=3$, the definition is equivalent to Eq.~\eqref{eq:unvec_3ten}.
Using the definition of $N=3$, we can define $N=4$.
Using the definition of $N=4$, we can further define $N=5$.
Thinking about it this way, we can see that the formula \eqref{eq:unvec_tensor} can be applied to any $N > 2$.

When we call for a vectorization of an $N$th-order tensor, it calls for vectorization of $(N-1)$th-order tensors, which in turn calls for vectorization of $(N-2)$th-order tensors, $\cdots$, and the innermost, calls for vectorization of matrices. Then, in reverse order (from the inside to the outside), the vectorization of the matrices are completed and the results are used to complete the vectorization of the third-order tensors, $\cdots$, the vectorization of the $(N-1)$th-order tensors, and finally, the vectorization of the $N$th-order tensor is completed using their results.
While this may be different from the actual implementation of computational software, this definition allows us to correctly understand the input and output of vectorization of an $N$th-order tensor.
Figure~\ref{fig:vectorization_of_tensors} shows how the vectorization of an $N$th-order tensor is defined by recursively applying the matrix vectorization operator.

\subsection{Matricization}\label{sec:matricization}

\paragraph{Mode-$n$ matricization}
{\em Matricization} is an operation of converting an $N$th-order tensor into a matrix.
In particular, {\em mode-$n$ matricization} is an operation of selecting the $n$-th mode as the first mode and combining all the remaining modes into the second mode to a matrix.
For example, there are three patterns of the mode-$n$ matricization of the third-order tensor $\ten{A} \in \bbR{I \times J \times K}$ as follows:
\begin{align}
   \mat A_{(1)} \in \bbR{I \times JK}, \\
   \mat A_{(2)} \in \bbR{J \times IK}, \\
   \mat A_{(3)} \in \bbR{K \times IJ}. 
\end{align}
Each entry satisfies the following relationship:
\begin{align}
 A_{(1)}(i,\overline{jk}) = A_{(2)}(j,\overline{ik}) = A_{(3)}(k,\overline{ij}) = \mathcal{A}(i,j,k).
\end{align}

\begin{figure}[t]
    \centering
    \includegraphics[width=0.99\textwidth]{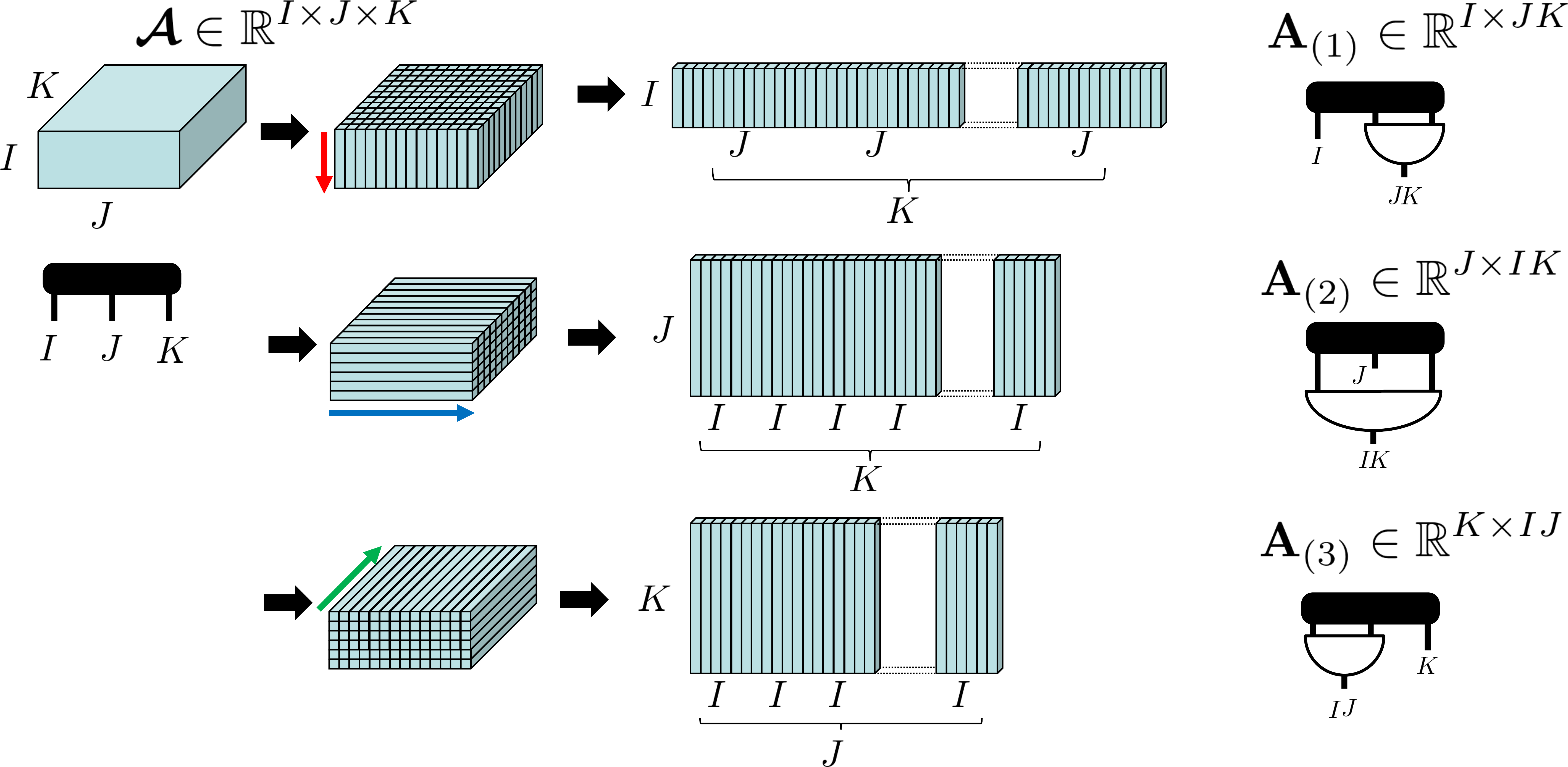}
    \caption{Matricization of a third-order tensor.}\label{fig:matrix_unfolding}
\end{figure}

Figure~\ref{fig:matrix_unfolding} shows an image of matricization and its diagrams. 
From this image, it can also be understood as the operation of ``extracting and arranging all fibers''.
The horizontal arrangement is the same order as vectorization.

For the third-order tensor $\ten{A} \in \bbR{I \times J \times K}$, the mode-1 matricization $\mat{A}_{(1)} \in \bbR{I \times JK}$, the mode-2 matricization $\mat{A}_{(2)} \in \bbR{J \times IK}$, and the mode-3 matricization $\mat{A}_{(3)} \in \bbR{K \times IJ}$ can be written as follows:
\begin{align}
    &\mat{A}_{(1)} = [\unvec(\ten{A}(1,:,:)), \unvec(\ten{A}(2,:,:)), ..., \unvec(\ten{A}(I,:,:))]^\top, \\
    &\mat{A}_{(2)} = [\unvec(\ten{A}(:,1,:)), \unvec(\ten{A}(:,2,:)), ..., \unvec(\ten{A}(:,J,:))]^\top, \\
    &\mat{A}_{(3)} = [\unvec(\ten{A}(:,:,1)), \unvec(\ten{A}(:,:,2)), ..., \unvec(\ten{A}(:,:,K))]^\top.
\end{align}
This shows that the vectorizations of slices of tensors are arranged for the matricization.

The mode-$n$ matricization for an $N$th-order tensor $\ten{T} \in \bbR{I_1 \times I_2 \times \cdots \times I_N}$ can also be written in general as follows:
\begin{align}
    &\mat{T}_{(n)} = \begin{bmatrix}
        \unvec(\mathrm P_n\ten{T}(1,:,...,:))^\top \\
        \unvec(\mathrm P_n\ten{T}(2,:,...,:))^\top \\
        \vdots \\
        \unvec(\mathrm P_n\ten{T}(I_n,:,...,:))^\top
    \end{bmatrix} \in \bbR{I_n \times \prod_{k \neq n} I_k}, \\
    &\mathrm P_n\ten{T} = \perm_{[n,1,...,n-1,n+1,...,N]}(\ten{T}) \in \bbR{I_n \times I_1 \times \cdots \times I_{n-1} \times I_{n+1} \times \cdots \times I_N}. \notag
\end{align}
For each entry, we have $T_{(n)}(i_n, \overline{i_1\cdots i_{n-1} i_{n+1} \cdots i_N}) = \mathcal{T}(i_1, i_2, ..., i_N)$.
Here, $\mathrm P_n$ represents the mode permutation that brings the $n$-th mode of the input tensor to the first mode.
In other words, when a tensor of size $(I_1, ..., I_N)$ is input, a tensor of size $(I_n,I_1, ..., I_{n-1}, I_{n+1}, ..., I_N)$ is output.

\begin{figure}[t]
  \centering
  \includegraphics[width=0.99\textwidth]{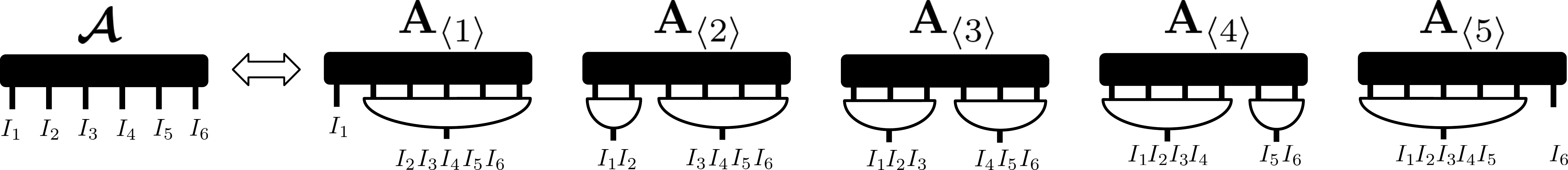}
  \caption{$k$-unfolding.}\label{fig:k_unfolding}
\end{figure}

\paragraph{$k$-unfolding}
{\em $k$-unfolding} is an operation of matricization of a tensor without using mode permutation.
$k$-unfolding of an $N$th-order tensor $\ten{A} \in \bbR{I_1 \times I_2 \times \cdots \times I_N}$ with mode-$k$ is written by
\begin{align}
  \mat{A}_{\ang{k}} \in \bbR{(I_1 \cdots I_k) \times (I_{k+1} \cdots I_N)}.
\end{align}
Each entry of $k$-unfolding is given by
\begin{align}
  {A}_{\ang{k}}(\overline{i_1 \cdots i_k}, \overline{i_{k+1} \cdots i_N}) = \mathcal{A}(i_1, i_2, ..., i_N).
\end{align}
Figure~\ref{fig:k_unfolding} shows $k$-unfolding of a tensor in diagrams.
$k$-unfolding is a key building block in the TT decomposition which is discussed in Section~\ref{sec:TT-decomposition}.

The following relationship holds between $k$-unfolding and mode-$n$ matricization: 
\begin{align}
  &\mat{A}_{(1)} = \mat{A}_{\ang{1}},  \\
  &\mat{A}_{(k)} \neq \mat{A}_{\ang{k}} \ \ (k = \{2, ..., N-2\}),\\
  &\mat{A}_{(N)} = \mat{A}_{\ang{N-1}}^\top.
\end{align}

\begin{figure}[t]
    \centering
    \includegraphics[width=0.8\textwidth]{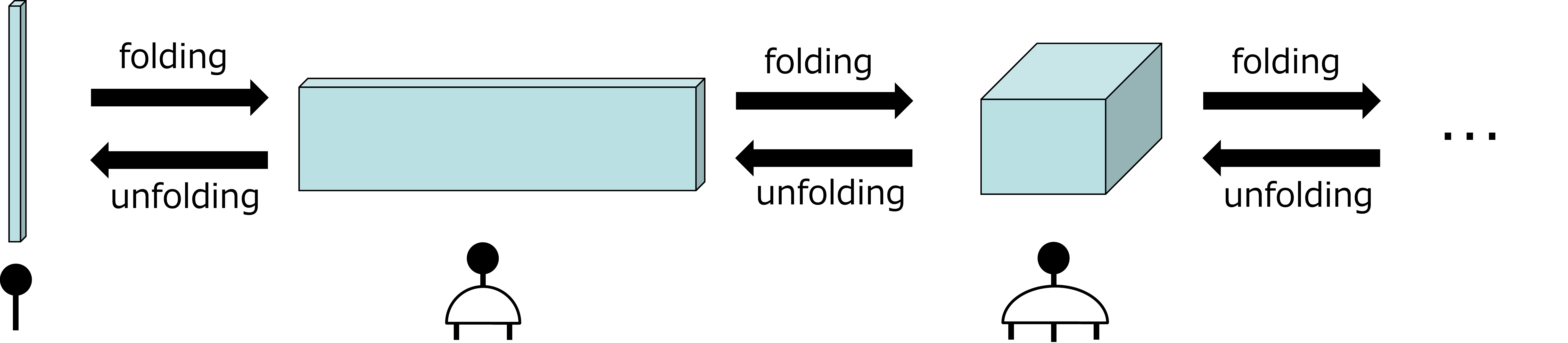}
    \caption{Folding and unfolding.}\label{fig:folding}
\end{figure}

\subsection{Folding (tensorization)}
Up to this point, we have introduced reshaping of tensors into vectors and matrices.
These operations are the transformation from higher-order tensors into lower-order tensors, and are called {\em unfolding} in general.

In this section, we will introduce the inverse transformation from lower-order tensors to higher-order tensors.
This type of operation is called {\em folding} or {\em tensorization}.
Figure~\ref{fig:folding} shows an image of folding and its diagrams.
In the diagram, we can represent a folding by connecting the vectorization operator in the ``reverse direction''.

\paragraph{Folding a vector to a matrix}
The simplest folding is a transformation from vector $\vect{a} \in \bbR{IJ}$ to matrix $\mat{A} \in \bbR{I \times J}$.
When vector $\vect{a} \in \bbR{IJ}$ can be expressed as a vector $\vect{a}_j \in \bbR{I}$ of length $I$ concatenated $J$ times
\begin{align}
\vect{a} = \begin{bmatrix} \vect{a}_1 \\ \vdots \\ \vect{a}_J \end{bmatrix} \in \bbR{IJ},
\end{align}
its folding is given as
\begin{align}
\mat{A} = \fold_{(I,J)}(\vect{a}) = [ \vect{a}_1, ..., \vect{a}_J] \in \bbR{I \times J}.
\end{align}
The folding can be said to be the inverse transformation (inverse mapping) of vectorization as 
\begin{align}
&\fold_{(I,J)}(\unvec(\mat{A})) = \mat{A}, \\
&\unvec(\fold_{(I,J)}(\vect{a})) = \vect{a}.
\end{align}

\begin{figure}[t]
    \centering
    \includegraphics[width=0.8\textwidth]{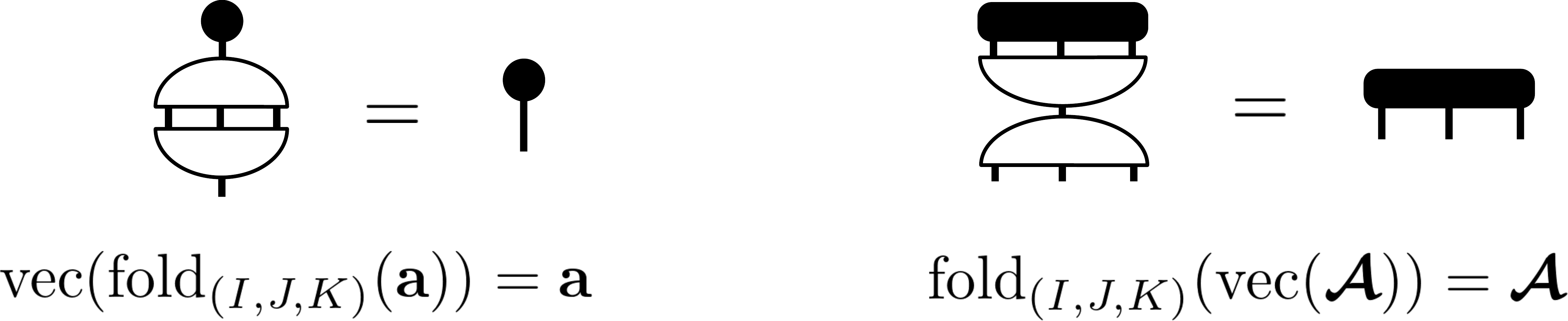}
    \caption{Inverse of folding and unfolding.}\label{fig:diagram_fold_unfold}
\end{figure}

\paragraph{Folding a vector to a tensor}
The same applies to third-order tensors.
The folding from vector $\vect{a} \in \bbR{IJK}$ to tensor $\ten{A} \in \bbR{I \times J \times K}$ can be defined\footnote{This means \texttt{A = reshape(a,[I J K])} in MATLAB.} as
\begin{align}
\ten{A} = \fold_{(I,J,K)}(\vect{a}),
\end{align}
and it is the inverse transformation of vectorization as
\begin{align}
    &\fold_{(I,J,K)}(\unvec(\ten{A})) = \ten{A}, \label{eq:ten_vec} \\
    &\unvec(\fold_{(I,J,K)}(\vect{a})) = \vect{a}. \label{eq:vec_ten}
\end{align}
Figure~\ref{fig:diagram_fold_unfold} shows the graphical notation of \eqref{eq:ten_vec} and \eqref{eq:vec_ten}.
From this diagram, we can get a more concrete idea that the operations of \eqref{eq:ten_vec} and \eqref{eq:vec_ten} are identity mappings.

\subsection{Folding/unfolding operators are tensors}
The diagram of folding or unfolding operator (the semicircular shapes) is node-with-edges and can be thought of as a tensor.
In addition, their operations can be understood as tensor products.

Let us consider the following simple example of matricization:
\begin{align}
\begin{bmatrix} a \\ b \\ c \\ d \end{bmatrix} \mathop{\longrightarrow}^{\text{fold}} \begin{bmatrix} a & c \\ b & d \end{bmatrix}. \label{eq:2by2matrix}
\end{align}
Note that \eqref{eq:2by2matrix} can be rewritten as a linear combination of all 2-by-2 matrix units as
\begin{align}
a\begin{bmatrix} 1 & 0 \\ 0 & 0 \end{bmatrix}
+b\begin{bmatrix} 0 & 0 \\ 1 & 0 \end{bmatrix}
+c\begin{bmatrix} 0 & 1 \\ 0 & 0 \end{bmatrix}
+d\begin{bmatrix} 0 & 0 \\ 0 & 1 \end{bmatrix}.
\end{align}
In this case, the folding operator is a third-order tensor of which all the 2-by-2 matrix units are arranged along the third mode.
Since individual matrix units are matricization of one-hot vectors, the third-order tensor can be given as the tensorization of 4-by-4 identity matrix\footnote{It can be written by \texttt{reshape(eye(4), [2,2,4])} in MATLAB.} as follow:
\begin{align}
\fold_{(2,2,4)}\left( \mat{I}_4 \right) \in \bbR{2 \times 2 \times 4}.
\end{align}
In more general, the folding operator from a vector of length $T=\prod_{n=1}^N I_n$ to $N$th-order tensor of size $(I_1, I_2, ..., I_N)$ can be represented by $(N+1)$th-order tensor as 
\begin{align}
\fold_{(I_1,I_2, ..., I_N,T)}\left( \mat{I}_T \right) \in \bbR{I_1 \times I_2 \times \cdots \times I_N \times T}.
\end{align}

\begin{figure}[t]
    \centering
    \includegraphics[width=0.99\textwidth]{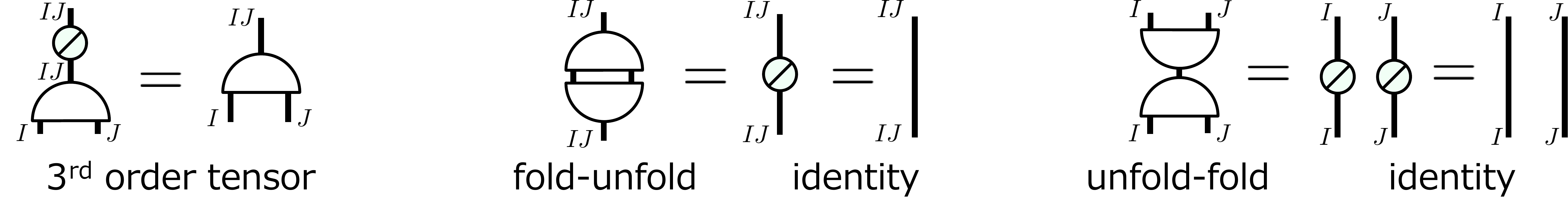}
    \caption{Tensor representations of folding and unfolding operators.}\label{fig:diagram_unfolding_2}
\end{figure}

Figure~\ref{fig:diagram_unfolding_2} shows examples of operators that perform matrix-vector unfolding and folding operations.
An operator that converts a vector of length $IJ$ into a matrix of size $(I,J)$ can be expressed as a third-order tensor of size $(I,J,IJ)$.
Since unfolding and folding are inverse mappings of each other, combining them results in an identity mapping.

\subsection{Linearity of reshaping operations}
Reshaping tensors ($\perm$, $\unvec$, $\fold$) changes the index of entry, not the value of the entry.
Therefore, for some operations, the result will be the same whether we reshape them before or after the operation.

For example, for any tensors $\ten{X}$, $\ten{Y}$ and scalar $\alpha$, the following linearity holds:
\begin{align}
&\unvec(\ten{X}+\ten{Y}) = \unvec(\ten{X}) + \unvec(\ten{Y}), \\
&\unvec(\alpha \ten{X}) = \alpha \unvec(\ten{X}).
\end{align}
The same can be applied to all the folding/unfolding operations and the mode permutations.

\newpage

\section{Calculations for Tensors}\label{sec:main_study}

\subsection{Addition, subtraction, and scaling}
\paragraph{Addition}
{\em Addition} of two $N$th-order tensors is written as 
\begin{align}
\ten{X} + \ten{Y},
\end{align}
where $\ten{X} \in \bbR{I_1 \times \cdots \times I_N}$ and $\ten{Y} \in \bbR{I_1 \times \cdots \times I_N}$ are same size.
Each entry of $\ten{X} + \ten{Y} \in \bbR{I_1 \times \cdots \times I_N}$ is given by
\begin{align}
[\mathcal{X}+\mathcal{Y}]_{i_1,...,i_N} = \mathcal{X}_{i_1,...,i_N}+ \mathcal{Y}_{i_1,...,i_N}.
\end{align}
In general, we have $\ten{X} + \ten{Y} = \ten{X} + \ten{Y}$.

\paragraph{Subtraction}
{\em Subtraction} of two $N$th-order tensors is written as 
\begin{align}
\ten{X} - \ten{Y}.
\end{align}
Each entry of $\ten{X} - \ten{Y} \in \bbR{I_1 \times \cdots \times I_N}$ is given by
\begin{align}
[\mathcal{X}-\mathcal{Y}]_{i_1,...,i_N} = \mathcal{X}_{i_1,...,i_N} - \mathcal{Y}_{i_1,...,i_N}.
\end{align}

Since the addition and subtraction of the tensors perform an independent operation for each entry, the results do not change even if the input and output tensors are unfolded.
Therefore, we have
\begin{align}
\unvec(\ten{X} \pm \ten{Y}) = \unvec(\ten{X}) \pm \unvec(\ten{Y}).
\end{align}
By performing a folding operation on both sides, we have
\begin{align}
\ten{X} \pm \ten{Y} = \fold_{(I_1,...,I_N)}(\unvec(\ten{X}) \pm \unvec(\ten{Y})).
\end{align}
Thus, the addition (or subtraction) of tensors can also be expressed by combining the reshaping of tensors and the addition (or subtraction) of vectors.

\paragraph{Scaling}
{\em Scaling} of an $N$th-order tensor $\ten{X} \in \bbR{I_1 \times \cdots \times I_N}$ with a scalar value $a$ is written as \begin{align}
a \ten{X}
\end{align}
which means that all entries of $\ten{X}$ are multiplied by $a$.
In other words, each entry of $a\ten{X} \in \bbR{I_1 \times \cdots \times I_N}$ is given by
\begin{align} 
[a\mathcal{X}]_{i_1,...,i_N} = a \mathcal{X}_{i_1,...,i_N}.
\end{align}
We can also unfold it into vectors, and we have
\begin{align}
&\unvec(a \ten{X}) = a \unvec(\ten{X}), \\
&a \ten{X} = \fold_{(I_1,...,I_N)}(a \unvec(\ten{X})).
\end{align}

\subsection{Entry-wise multiplication and division}
\paragraph{Hadamard product}
Entry-wise multiplication of two tensors is called the {\em Hadamard product}\footnote{Also known as the Schur product.}.
In this note, the Hadamard product is represented as $\boxdot$\footnote{Please note that in some literature the Hadamard product is represented by many symbols such as $\circ$, $\odot$, $*$ and $\circledast$, but these symbols have conflict to other operations. $\circ$ is often used for the outer product \cite{kolda2009tensor,cichocki2007nmf}, $\odot$ is often used for the Khatri-Rao product \cite{kolda2009tensor,cichocki2007nmf}, and $*$ and $\circledast$ are frequently used for the convolution \cite{jain1989fundamentals,strang2019linear}.}.

\begin{figure}[t]
    \centering
    \includegraphics[width=0.8\textwidth]{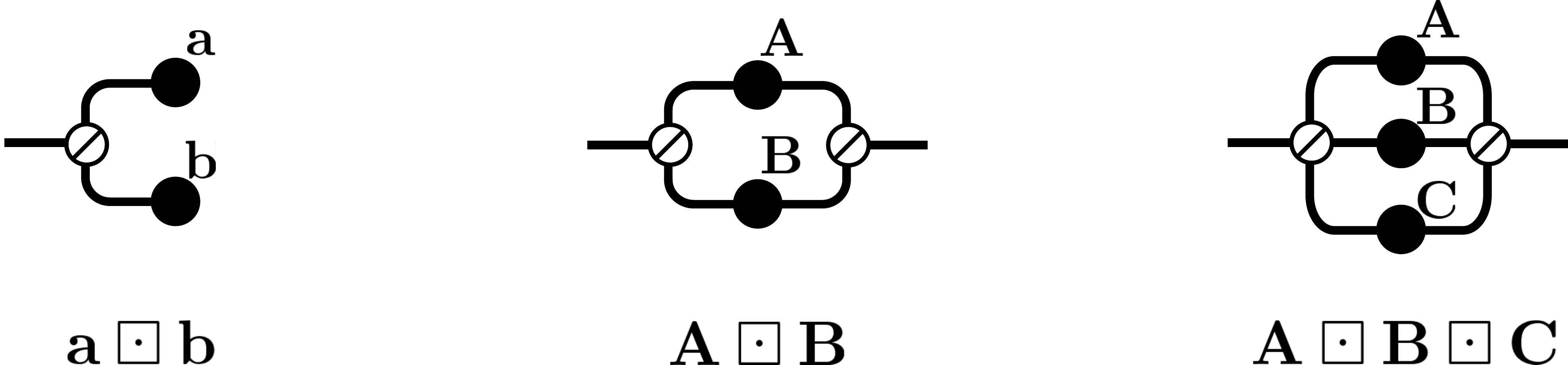}
    \caption{Hadamard products.}\label{fig:hadamard_product}
\end{figure}

First, let us consider the case of matrices.
The Hadamard product of two matrices $\mat{A} \in \bbR{I \times J}$ and $\mat{B} \in \bbR{I \times J}$ is given by
\begin{align}
\mat{A} \boxdot \mat{B} =
\begin{bmatrix}
A_{11}B_{11} & A_{12}B_{12} & \cdots & A_{1J}B_{1J} \\
A_{21}B_{21} & A_{22}B_{22} & \cdots & A_{2J}B_{2J} \\
\vdots & \vdots & \ddots & \vdots \\
A_{I1}B_{I1} & A_{I2}B_{I2} & \cdots & A_{IJ}B_{IJ}
\end{bmatrix} \in \bbR{I \times J}.
\end{align}
The each entry is $[A \boxdot B]_{ij} = A_{ij}B_{ij}$.
The Hadamard product is invariant with respect to order of matrices as follows:
\begin{align}
\mat{A} \boxdot \mat{B} &= \mat{B} \boxdot \mat{A}, \\
(\mat{A} \boxdot \mat{B}) \boxdot \mat{C}  & = \mat{A} \boxdot (\mat{B} \boxdot \mat{C}) = \mat{A} \boxdot \mat{B} \boxdot \mat{C}.
\end{align}

Figure~\ref{fig:hadamard_product} shows the graphical notation for the Hadamard product.
The diagonal line nodes are element-wise product operators\footnote{In fact, this operator is a super-diagonal tensor which all diagonal entries are 1.}. 
In Hadamard product of two vectors, two modes are input to the operator and the operator outputs one mode. The whole diagram can be seen as a vector.  
In the Hadamard product of two matrices, two matrices are sandwitched by two operators. The whole diagram can be seen as a matrix. 
Hadamard product of three matrices is illustrated in the same way.

The Hadamard product of two tensors of the same size, $\ten{X}$ and $\ten{Y} \in \bbR{I_1 \times \cdots \times I_N}$, is written as
\begin{align}
\ten{X} \boxdot \ten{Y} \in \bbR{I_1 \times \cdots \times I_N}.
\end{align}
Each entry is given by 
\begin{align}
[\mathcal{X} \boxdot \mathcal{Y}]_{i_1,...,i_N} = \mathcal{X}_{i_1,...,i_N} \mathcal{Y}_{i_1,...,i_N}.
\end{align}
In general, $\ten{X} \boxdot \ten{Y} = \ten{Y} \boxdot \ten{X}$.

\paragraph{Entry-wise division}
{\em Entry-wise division} is denoted by $\boxslash$.
The entry-wise division of two tensors of the same size, $\ten{X}$ and $\ten{Y} \in \bbR{I_1 \times \cdots \times I_N}$, is written by
\begin{align}
\ten{X} \boxslash \ten{Y} \in \bbR{I_1 \times \cdots \times I_N},
\end{align}
and the each entry is given by 
\begin{align}
[\mathcal{X} \boxslash \mathcal{Y}]_{i_1,...,i_N} = \frac{\mathcal{X}_{i_1,...,i_N}}{ \mathcal{Y}_{i_1,...,i_N}}.
\end{align}

\paragraph{Diagonal matrix representation of Hadamard product}
Here, let us consider the vectorization of the Hadamard product and entry-wise division.
For simple notation, we put
\begin{align}
   &T = \prod_{n=1}^N I_n, \\
   &\vect{x} = \unvec(\ten{X}) \in \bbR{ T }, \\
   &\vect{y} = \unvec(\ten{Y}) \in \bbR{ T}.
\end{align}
The point here is to regard the Hadamard product as a linear mapping $\bbR{T} \rightarrow \bbR{T}$.
Considering a diagonal matrix $\mat{D}_\vect{y} = \text{diag}(\vect{y}) \in \bbR{ T \times T }$, we have
\begin{align}
    &\vect{x} \boxdot \vect{y} = \mat{D}_\vect{y} \vect{x}. \\
    &\vect{x} \boxslash \vect{y} = \mat{D}_\vect{y}^{-1} \vect{x}.
\end{align}
Then, Hadamard product and entry-wise division can be written using folding/unfolding as
\begin{align}
    &\ten{X} \boxdot \ten{Y} = \fold_{(I_1,...,I_N)}(\mat{D}_{\unvec(\ten{Y})} \unvec(\ten{X}) ),\\
    &\ten{X} \boxslash \ten{Y} = \fold_{(I_1,...,I_N)}(\mat{D}_{\unvec(\ten{Y})}^{-1} \unvec(\ten{X}) ).
\end{align}
The understanding of Hadamard product as a diagonal matrix multiplication is important to discuss properties of the operation such as positive definiteness.

\subsection{Hadamard product operators are tensors}\label{sec:Hadamard_prod_superdiagonal}
The diagram of the Hadamard product operator (the diagonal line node) is a tensor and that operation can be understood as tensor product.

For a simple case, let us consider the following Hadamard product:
\begin{align}
\vect{a} \boxdot \vect{b}=\begin{bmatrix} a_1 \\ a_2 \\ a_3\end{bmatrix} \boxdot \begin{bmatrix} b_1 \\ b_2 \\ b_3\end{bmatrix} = 
\begin{bmatrix} a_1& 0 & 0 \\ 0& a_2 &0 \\ 0&0&a_3\end{bmatrix}\begin{bmatrix} b_1 \\ b_2 \\ b_3\end{bmatrix} = \begin{bmatrix} a_1b_1 \\ a_2b_2 \\ a_3b_3\end{bmatrix}.
\end{align}
The diagonal matrix can be represented as a linear combination of three matrix units as
\begin{align}
\begin{bmatrix} a_1& 0 & 0 \\ 0& a_2 &0 \\ 0&0&a_3\end{bmatrix}=a_1\begin{bmatrix} 1& 0 & 0 \\ 0& 0 &0 \\ 0&0&0\end{bmatrix}+a_2\begin{bmatrix} 0& 0 & 0 \\ 0& 1 &0 \\ 0&0&0\end{bmatrix}+a_3\begin{bmatrix} 0& 0 & 0 \\ 0& 0 &0 \\ 0&0&1\end{bmatrix}.
\end{align}
From the above, we can consider a third-order tensor in which three matrix units are arranged in the depth direction.
The size of this tensor $\ten{I}$ is $(3,3,3)$ and its entries are given by
\begin{align}
\mathcal{I}_{ijk} = \left\{\begin{array}{ll} 1 & (i=j=k) \\ 0 & \text{otherwise} \end{array}\right. .
\end{align}
The tensor $\ten{I}$ is called the {\em super-diagonal tensor}.
The tensor product of a third-order super-diagonal tensor and a vector is a diagonal matrix.
That is, the tensor product of two vectors with a super-diagonal tensor is the Hadamard product of the two vectors.

See Figure~\ref{fig:hadamard_product}, again.
You will understand the meaning of the following formula:
\begin{align}
 \vect{a} \boxdot \vect{b} = \ten{I} \times_2 \vect{a}^\top \times_3 \vect{b}^\top,
\end{align}
where $\times_n$ is a mode product, but we will not go into its definition here (see Section~\ref{sec:mode_product}).

\begin{figure}[t]
    \centering
    \includegraphics[width=0.95\textwidth]{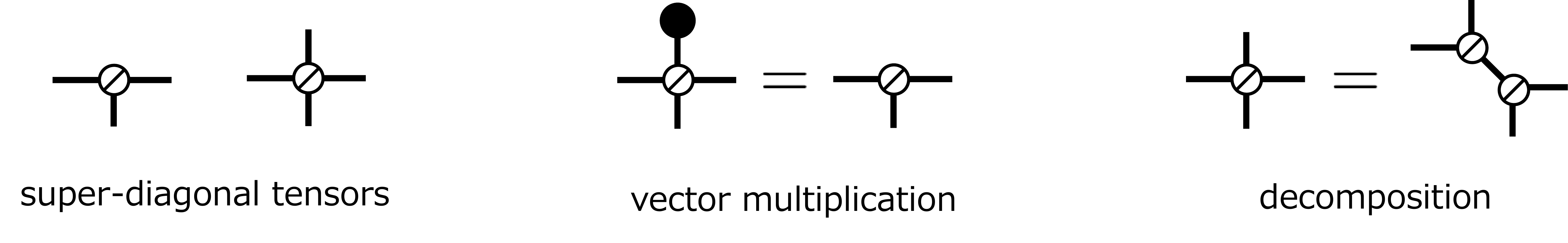}
    \caption{Super-diagonal tensors and their properties. Mode product of a super-diagonal tensor with a vector is still super-diagonal. A higher-order super-diagonal tensor can be decomposed into two lower-order super-diagonal tensors.}\label{fig:superdiagonal}
\end{figure}

\paragraph{Properties of super-diagonal tensors}
Here, we discuss super-diagonal tensors.
Figure~\ref{fig:superdiagonal} succinctly illustrates the properties of super-diagonal tensors.
If we write the $N$th-order super-diagonal tensor whose diagonal entries are 1 as $\ten{I}^{(N)}$, the following property holds for each entry:
\begin{align}
\mathcal{I}^{(3)}_{i_1 i_2 i_3} &= \mathcal{I}^{(2)}_{i_1 i_2}\mathcal{I}^{(2)}_{i_2 i_3}, \\
\mathcal{I}^{(4)}_{i_1 i_2 i_3 i_4} &= \mathcal{I}^{(3)}_{i_1 i_2 i_3}\mathcal{I}^{(2)}_{i_3 i_4}, \\
\mathcal{I}^{(5)}_{i_1 i_2 i_3 i_4 i_5} &= \mathcal{I}^{(4)}_{i_1 i_2 i_3 i_4}\mathcal{I}^{(2)}_{i_4 i_5}, \\
& \ \vdots \notag \\
\mathcal{I}^{(N)}_{i_1 i_2 \cdots i_{N-1} i_N} &= \mathcal{I}^{(N-1)}_{i_1 i_2 \cdots i_{N-1}}\mathcal{I}^{(2)}_{i_{N-1} i_N}.
\end{align}
That is, we can construct a higher-order super-diagonal tensor by multiplying lower-order super-diagonal tensors.

In contrast, summation along a mode in a higher-rank super-diagonal tensor reduces to a lower-rank super-diagonal tensor as
\begin{align}
\sum_{i_3} \mathcal{I}^{(3)}_{i_1 i_2 i_3} &= \mathcal{I}^{(2)}_{i_1 i_2}, \\
\sum_{i_4}\mathcal{I}^{(4)}_{i_1 i_2 i_3 i_4} &= \mathcal{I}^{(3)}_{i_1 i_2 i_3}, \\
\sum_{i_5}\mathcal{I}^{(5)}_{i_1 i_2 i_3 i_4 i_5} &= \mathcal{I}^{(4)}_{i_1 i_2 i_3 i_4}, \\
& \ \vdots \notag \\
 \sum_{i_N} \mathcal{I}^{(N)}_{i_1 i_2 \cdots i_{N-1} i_N} &= \mathcal{I}^{(N-1)}_{i_1 i_2 \cdots i_{N-1}}.
\end{align}
From above equations, we can derive the following identity
\begin{align}
&\mathcal{I}^{(4)}_{i_1 i_2 i_4 i_5} = \sum_{i_3} \mathcal{I}^{(3)}_{i_1 i_2 i_3}\mathcal{I}^{(3)}_{i_3 i_4 i_5} \\
&\Leftrightarrow \ten{I}^{(4)} = \ten{I}^{(3)} \mathbin{_3\times^1} \ten{I}^{(3)},
\end{align}
where $\mathbin{_n\times^m}$ is a tensor product, but we will not go into its definition here (see Section~\ref{sec:tensor_product}).
In more general, we have
\begin{align}
 \ten{I}^{(5)} &= \ten{I}^{(4)} \mathbin{_4\times^1} \ten{I}^{(3)}, \\
 \ten{I}^{(6)} &= \ten{I}^{(5)} \mathbin{_5\times^1} \ten{I}^{(3)}, \\
 & \ \vdots \notag \\
 \ten{I}^{(N)} &= \ten{I}^{(N-1)} \mathbin{_{N-1}\times^1} \ten{I}^{(3)}.
\end{align}

\subsection{Broadcasting option for entry-wise calculation}
{\em Broadcasting} is an operation to duplicate a tensor to construct a higher-order tensor.
For example, broadcasting of a column vector to a matrix is as follow:
\begin{align}
    \begin{bmatrix}
        1 \\ 2 \\ 3 
    \end{bmatrix}
    \mathop{\longrightarrow}^{\text{broadcasting}}
    \begin{bmatrix}
        1 & 1 & \cdots & 1\\ 2 & 2 & \cdots & 2 \\ 3 & 3 & \cdots & 3 
    \end{bmatrix}.
\end{align}

There are many cases in which some entry-wise calculation of two different size tensors $\ten{X}$ and $\ten{Y}$ is performed by allowing broadcasting of the alignment of the tensor sizes.
We denote the broadcasted addition, subtraction, multiplication, and division by
\begin{align}
  &\ten{X} \boxplus  \ten{Y},\\
  &\ten{X} \boxminus \ten{Y},\\
  &\ten{X} \boxdot   \ten{Y}, \\
  &\ten{X} \boxslash \ten{Y},
\end{align}
respectively.
Since the Hadamard product and entry-wise division are special cases of broadcasted operations, we reuse $\boxdot$ and $\boxslash$ for more general ones.
These operations can be defined only for two alignable tensors with broadcasting.
The alignability of two tensors is called the {\em broadcast condition}\cite{matsui2024broadcast}.
In addition, the broadcasting option is implemented in many computational software such as MATLAB\footnote{\url{https://www.mathworks.com/help/matlab/matlab_prog/compatible-array-sizes-for-basic-operations.html}}, Python\footnote{\url{https://numpy.org/doc/stable/user/basics.broadcasting.html}} and Julia\footnote{\url{https://docs.julialang.org/en/v1/manual/arrays/\#Broadcasting}}.

\paragraph{Centralization and standardization}
A typical example of the broadcasted operation would be {\em centralization}:
\begin{align}
  \tilde{\vect{x}}_j = \vect{x}_j - \bm\mu,
\end{align}
where $\{\vect{x}_1, ..., \vect{x}_J \}$ are data, and $\bm\mu = \frac{1}{J}\sum_{j=1}^J \vect{x}_j$ is a mean vector.
For matrix representation of data $\mat{X} = [\vect{x}_1, ..., \vect{x}_J] $, its centralization $\widetilde{\mat{X}} = [\tilde{\vect{x}}_1, ..., \tilde{\vect{x}}_J] $ can be written by
\begin{align}
  \widetilde{\mat{X}} = \mat{X} \boxminus \bm\mu = \mat{X} - [\bm\mu, ..., \bm \mu],
\end{align}
where $\bm\mu$ is broadcasted along row direction.
Furthermore, let $\bm\sigma$ be a column vector arranging the standard deviations of each dimension, its {\em standardization} $\widehat{\mat{X}}$ can be written as
\begin{align}
\widehat{\mat{X}} = (\mat{X} \boxminus \mu) \boxslash \bm\sigma.
\end{align}
The entry-wise representation of this calculation can be written by
\begin{align}
  \widehat{X}_{ij} = \frac{ X_{ij} - \mu_i }{\sigma_i}.
\end{align}

\subsection{Inner product}
{\em Inner product} of two vectors $\vect{a}$ and $\vect{b} \in \bbR{I} $ is defined by
\begin{align}
    \langle \vect{a}, \vect{b} \rangle = \sum_{i=1}^I a_i b_i  = \vect{a}^\top \vect{b}.
\end{align}
It calculates the sum of all the entries of the Hadamard product.

In more general, inner product of two tensors $\ten{X}$ and $\ten{Y} \in \bbR{I_1 \times \cdots \times I_N} $ is defined by 
\begin{align}
    \langle \ten{X}, \ten{Y} \rangle &= \sum_{i_1, ..., i_N} \mathcal{X}_{i_1,...,i_N}\mathcal{Y}_{i_1,...,i_N}\\  &= \unvec(\ten{X})^\top \unvec(\ten{Y}) = \langle \unvec(\ten{X}), \unvec(\ten{Y}) \rangle. \label{eq:unvec_inner_prod}
\end{align}
Inner product of two tensors is equivalent to inner product of their vectorizations.
The same applies to any unfolding and folding.

\begin{figure}[t]
    \centering
    \includegraphics[width=0.7\textwidth]{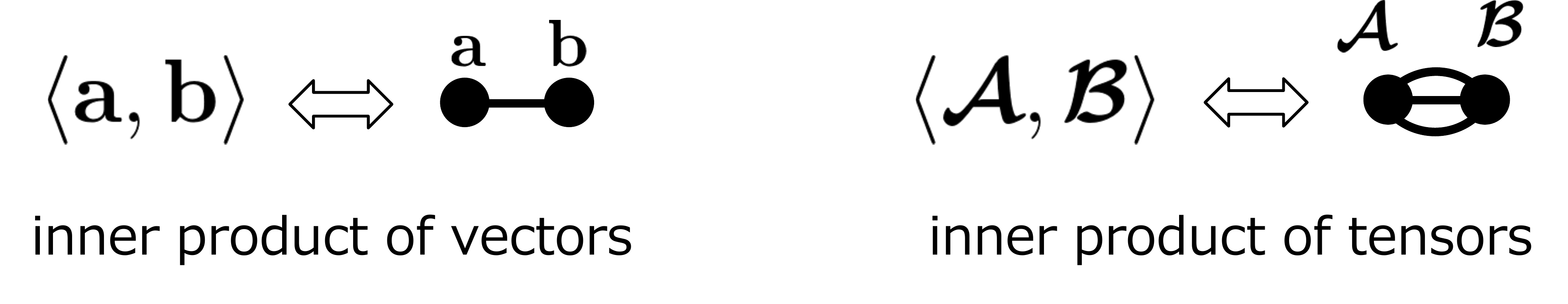}
    \caption{Inner products.}\label{fig:inner_product}
\end{figure}

Figure~\ref{fig:inner_product} shows the graphical notations for the inner products.
In the tensor network diagram, the inner products are represented as edge connections.
Computation of the inner product which is represented by the connected edge results a scalar value which is represented as a node without edges in the diagram.
Such operations (computations) are called {\em tensor network contractions}.
Contraction is usually used in the context of operations that actually spend the computational costs to eliminate edges, rather than as a conceptual operation that connects edges.

\subsection{Trace}
{\em Trace} of a square matrix $\mat{S} \in \bbR{I \times I}$ is defined by
\begin{align}
  \tr(\mat{S}) = \sum_{i=1}^I S_{ii}.
\end{align}
An important property of the trace is $\tr(\mat{A}\mat{B}) = \tr(\mat{B}\mat{A})$.
Since the trace $\tr(\mat{S})$ is a sum of all the diagonal entries, it can be considered as an inner product between a matrix $\mat{S}$ and an identity matrix $\mat{I}$ as
\begin{align}
  \tr(\mat{S}) = \inp{\mat{S}}{\mat{I}}.
\end{align}

\begin{figure}[t]
    \centering
    \includegraphics[width=0.25\textwidth]{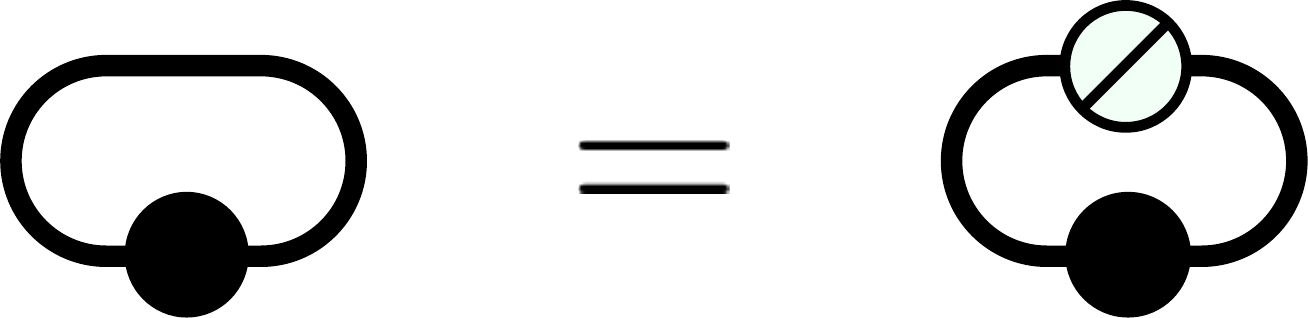}
    \caption{Trace.}\label{fig:trace}
\end{figure}

Figure~\ref{fig:trace} shows the diagram of the matrix trace.
This can be easily understood with the interpretation by the inner product.

\subsection{Outer product}
{\em Outer product} of two vectors $\vect{a} \in \bbR{I}$ and $\vect{b} \in \bbR{J} $ is denoted by
\begin{align}
    \vect{a} \circ \vect{b}.
\end{align}
The outer product of two vectors results in a matrix $\mat{X} = \vect{a} \circ \vect{b} \in \bbR{I \times J}$, and each entry is given by
\begin{align}
    X_{ij} = a_i b_j.
\end{align}
We have $\vect{a} \circ \vect{b} = \vect{a} \vect{b}^\top = \vect{a} \boxdot \vect{b}^\top$.

In the same way, the outer product of three vectors $\vect{a} \in \bbR{I}$, $\vect{b} \in \bbR{J} $ and $\vect{c} \in \bbR{K}$ is denoted by
\begin{align}
    \vect{a} \circ \vect{b} \circ \vect{c}.
\end{align}
The outer product of three vectors results in a third-order tensor $\ten{X} = \vect{a} \circ \vect{b} \circ \vect{c} \in \bbR{I \times J \times K}$, and each entry is given by
\begin{align}
    \mathcal{X}_{ijk} = a_i b_j c_k.
\end{align}

In general, the outer product of $N$ vectors $\vect{a}_1 \in \bbR{I_1}$, $\cdots$, $\vect{a}_N \in \bbR{I_N}$ is denoted by 
\begin{align}
    \vect{a}_1 \circ \cdots \circ \vect{a}_N .
\end{align}
The outer product of $N$ vectors results in an $N$th-order tensor $\ten{X} = \vect{a}_1 \circ \cdots \circ \vect{a}_N \in \bbR{I_1 \times \cdots \times I_N}$, and each entrie is given by
\begin{align}
    \mathcal{X}(i_1,...,i_N) = a_1(i_1) \cdots a_N(i_N).
\end{align}

\begin{figure}[t]
    \centering
    \includegraphics[width=0.7\textwidth]{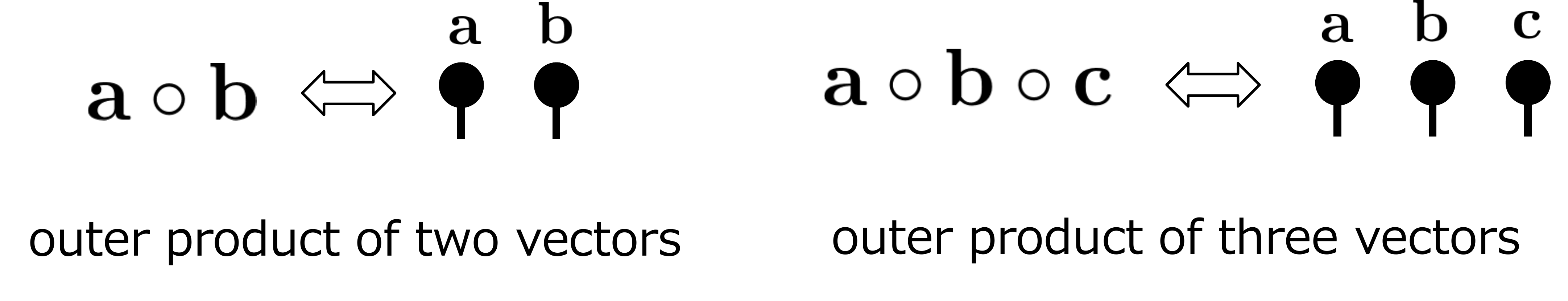}
    \caption{Outer products.}\label{fig:outer_product}
\end{figure}

Figure~\ref{fig:outer_product} shows the graphical notations for the outer products.
In the graphical notations, the outer product is represented by arranging nodes without connecting their edges.
By counting the number of unconnected edges, we can understand them as matrices or tensors.

\paragraph{Sum of all entries}
Let us consider the sum of all entries of a tensor:
\begin{align}
\sum_{i_1, ..., i_N} \mathcal{A}_{i_1, ...,i_N}.
\end{align}
Note that it can be written as an inner product between the tensor $\ten{A} \in \bbR{I_1 \times \cdots \times I_N}$ and the all-ones tensor $\mat{1}_{I_1 \times \cdots \times I_N} \in \bbR{I_1 \times \cdots \times I_N}$, and the all-ones tensor can be represented as an outer product of $N$ all-ones vectors $\vect{1}_{I_1} \circ \cdots \circ \vect{1}_{I_N}$.
Thus, we have
\begin{align}
 \sum_{i_1, ..., i_N} \mathcal{A}_{i_1, ...,i_N} = \inp{\ten{A}}{\mat{1}_{I_1 \times \cdots I_N}} = \inp{\ten{A}}{\vect{1}_{I_1} \circ \cdots \circ \vect{1}_{I_N} }.
\end{align}

\begin{figure}[t]
    \centering
    \includegraphics[width=0.5\textwidth]{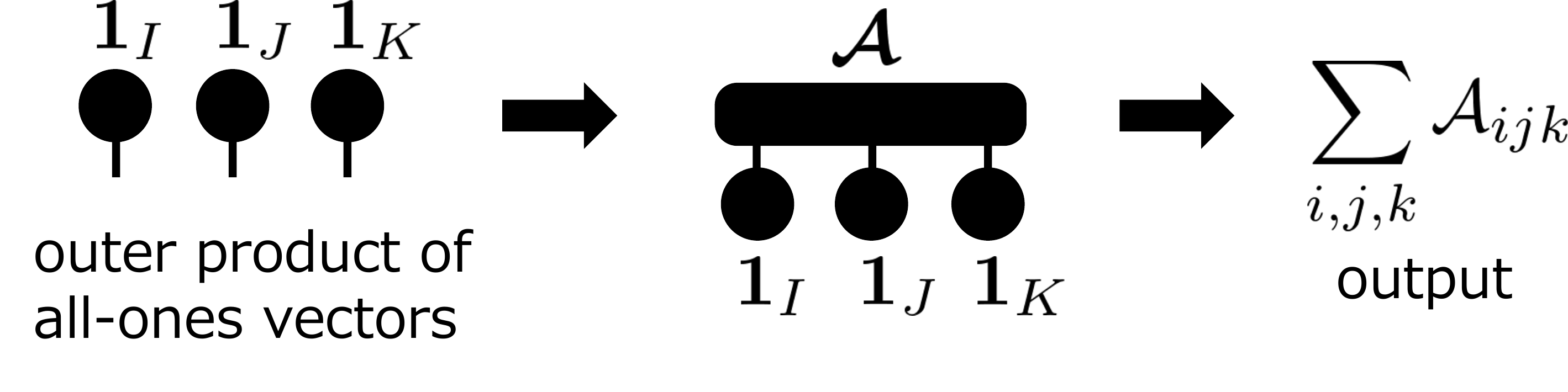}
    \caption{Sum of all entries.}\label{fig:sum_all}
\end{figure}

Figure~\ref{fig:sum_all} shows the graphical notation for the sum of all entries in a tensor.
The sum of all entries can be seen as an operation that connects the edges of the all-ones vectors to all the edges of the tensor.

\subsection{Matrix product}\label{sec:matrix_product}
\begin{figure}[t]
    \centering
    \includegraphics[width=0.4\textwidth]{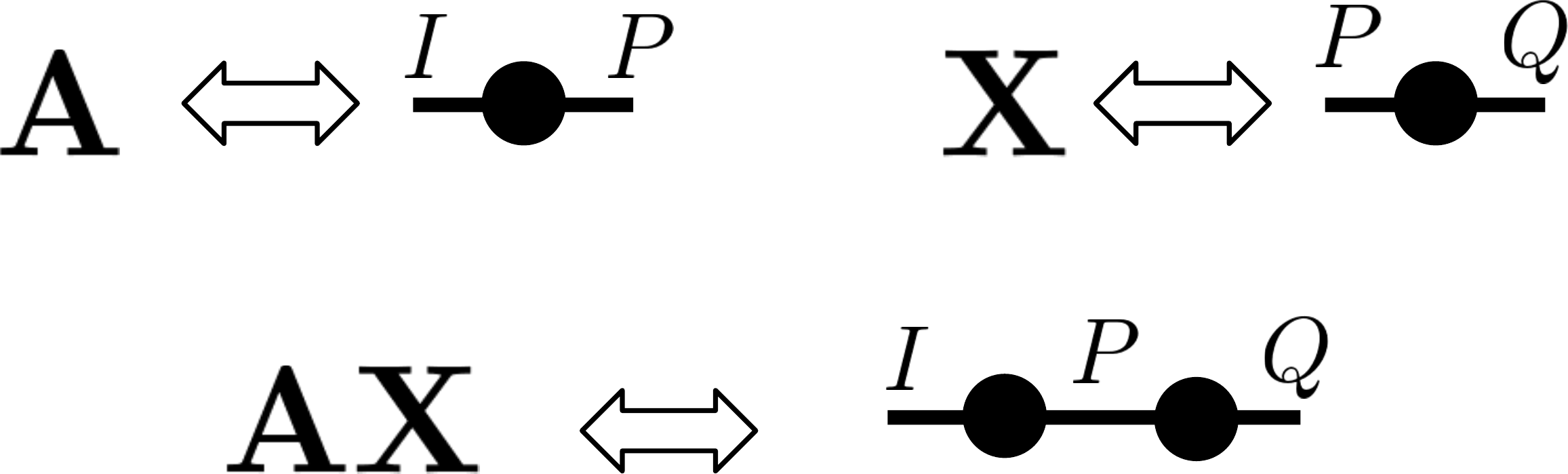}
    \caption{Matrix product.}\label{fig:mat_product}
\end{figure}

{\em Matrix product} is defined between two matrices with modes of the same length.
Let us consider two matrices
\begin{align}
   &\mat{A} = \begin{bmatrix} 
    | &  & | \\
    \vect{a}_1& \cdots &\vect{a}_P \\
    | &  & |
   \end{bmatrix} = 
   \begin{bmatrix} 
    \text{---} & \tilde{\vect{a}}_1^\top & \text{---} \\
    & \vdots & \\
    \text{---} & \tilde{\vect{a}}_I^\top & \text{---}
   \end{bmatrix} \in \bbR{I \times P},\\
   &\mat{X} = \begin{bmatrix} 
    | &  & | \\
    \vect{x}_1& \cdots &\vect{x}_Q \\
    | &  & |
   \end{bmatrix} = 
   \begin{bmatrix} 
    \text{---} & \tilde{\vect{x}}_1^\top & \text{---} \\
    & \vdots & \\
    \text{---} & \tilde{\vect{x}}_P^\top & \text{---}
   \end{bmatrix} \in \bbR{P \times Q},
\end{align}
and its product $\mat{Y} = \mat{A} \mat{X} \in \bbR{I \times Q}$.
A matrix of size $(I,P)$ and a matrix of size $(P,Q)$ have a common mode of the same length $P$, and the product results in a matrix of size $(I,Q)$ removing the mode of length $P$.
Figure~\ref{fig:mat_product} shows the diagram of the matrix product.
The matrix product is represented by an operation that connects the edges of length $P$.

There are two interpretations for the matrix product.
One of them is as follow:
\begin{align}
    \mat{Y} &= \mat{A} \mat{X} \notag \\
            &= \begin{bmatrix} 
                \text{---} & \tilde{\vect{a}}_1^\top & \text{---} \\
                & \vdots & \\
                \text{---} & \tilde{\vect{a}}_I^\top & \text{---}
               \end{bmatrix}
               \begin{bmatrix} 
                | &  & | \\
                \vect{x}_1& \cdots &\vect{x}_Q \\
                | &  & |
               \end{bmatrix} \notag \\
            &= \begin{bmatrix} 
                \langle \tilde{\vect{a}}_1,\vect{x}_1 \rangle &  \cdots & \langle \tilde{\vect{a}}_1,\vect{x}_Q \rangle\\
                \vdots& \ddots &\vdots \\
                \langle \tilde{\vect{a}}_I,\vect{x}_1 \rangle &  \cdots & \langle \tilde{\vect{a}}_I,\vect{x}_Q \rangle
               \end{bmatrix}.
\end{align}
Each entry of $\mat{Y}$ is an inner product.
The another one is as follow:
\begin{align}
    \mat{Y} &= \mat{A} \mat{X} \notag \\
            &= \begin{bmatrix} 
                | &  & | \\
                \vect{a}_1& \cdots &\vect{a}_P \\
                | &  & |
               \end{bmatrix}
               \begin{bmatrix} 
                \text{---} & \tilde{\vect{x}}_1^\top & \text{---} \\
                & \vdots & \\
                \text{---} & \tilde{\vect{x}}_P^\top & \text{---}
               \end{bmatrix} \notag \\
            &= \vect{a}_1 \circ \tilde{\vect{x}}_1 + \cdots + \vect{a}_P \circ \tilde{\vect{x}}_P.
\end{align}
$\mat{Y}$ is given in the form of a sum of outer products.
The operation of taking the sum of multiplications is an inner product, and the operation of arranging the combinations of multiplications is an outer product.
Matrix product can be thought of as an operation that computes an inner product for modes of length $P$, and an outer product for modes of sizes $I$ and $Q$.

\subsection{Frobenius norm}
{\em Euclidean norm} or {\em $\ell_2$-norm} of a vector $\vect{a} \in \bbR{I}$ is given as follows:
\begin{align}
|| \vect{a} ||_2 = \sqrt{ \langle \vect{a}, \vect{a} \rangle } = \sqrt{ \sum_{i=1}^I a_i^2 }.
\end{align}
It is the square root of the inner product with itself (the sum of the squares of all entries).

\begin{figure}[t]
    \centering
    \includegraphics[width=0.6\textwidth]{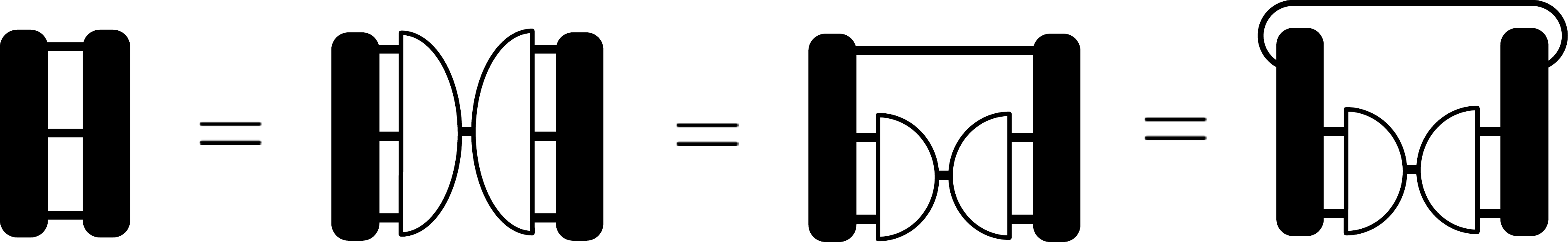}
    \caption{Square of Frobenius norm.}\label{fig:fro_norm}
\end{figure}

{\em Frobenius norm} is its extension to tensors.
Frobenius norm of an $N$th-order tensor $\ten{X} \in \bbR{I_1 \times \cdots \times I_N}$ is given by:
\begin{align}
    || \ten{X} ||_F = \sqrt{ \langle \ten{X}, \ten{X} \rangle } 
                     = \sqrt{ \sum_{i_1, ..., i_N} \mathcal{X}_{i_1,...,i_N}^2 }.
\end{align}
We have
\begin{align}
|| \ten{X} ||_F^2 = || \unvec(\ten{X}) ||_2^2 = || \mat{X}_{(n)} ||_F^2 = \tr(\mat{X}_{(n)} \mat{X}_{(n)}^\top),
\end{align}
and it can be easily understood with graphical notations shown in Figure~\ref{fig:fro_norm}.

\subsection{Kronecker product}\label{sec:kron}
{\em Kronecker product} is defined for two matrices $\mat{A} \in \bbR{I \times J}$ and $\mat{B} \in \bbR{P \times Q}$ and denoted as
\begin{align}
    \mat{A} \otimes \mat{B}.
\end{align}
When the size of $\mat{A}$ is $(I,J)$ and the size of $\mat{B}$ is $(P,Q)$, then the size of their Kronecker product $\mat{A} \otimes \mat{B}$ is $(PI,QJ)$.
It can be written as a block matrix as
\begin{align}
    \mat{A} \otimes \mat{B}= 
    \begin{bmatrix}
        A_{11}\mat{B} & A_{12}\mat{B} & \cdots & A_{1J}\mat{B} \\
        A_{21}\mat{B} & A_{22}\mat{B} & \cdots & A_{2J}\mat{B} \\
        \vdots & \vdots & \ddots & \vdots \\
        A_{I1}\mat{B} & A_{I2}\mat{B} & \cdots & A_{IJ}\mat{B}
    \end{bmatrix} \in \bbR{PI \times QJ}. \label{eq:kron}
\end{align}
Matrix $\mat{B}$ is arranged on a grid of $(I,J)$ and each $(i,j)$-th block is scaled by $A_{ij}$.
Each entry is given by
\begin{align}
    [A \otimes B](\overline{pi}, \overline{qj}) = A_{ij}B_{pq}.\label{eq:kron_entry}
\end{align}
We can see that all combinations of $A_{ij}B_{pq}$ are calculated.

\paragraph{Relationship between Kronecker product and outer product}
Kronecker product of two vectors $\vect{a} \in \bbR{I}$ and $\vect{b} \in \bbR{P}$ is given by
\begin{align}
    [a \otimes b](\overline{pi}) = a_{i}b_{p} = [b \circ a](p,i).\label{eq:kron_entry_vec}
\end{align}
From \eqref{eq:kron_entry_vec}, we have 
\begin{align}
  \vect{a} \otimes \vect{b} &= \unvec(\vect{b} \circ \vect{a}). \label{eq:kron_vec}
\end{align}
Note that the order of $\vect{a}$ and $\vect{b}$ is reversed in the outer product $\circ$ and the Kronecker product $\otimes$.

\begin{figure}[t]
    \centering
    \includegraphics[width=0.8\textwidth]{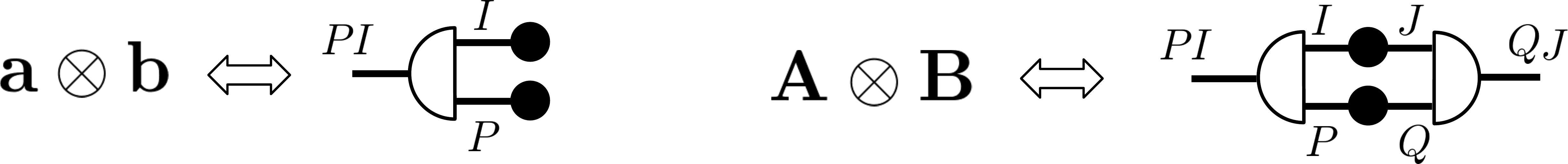}
    \caption{Kronecker product.}\label{fig:kron}
\end{figure}

Figure~\ref{fig:kron} shows the graphical notation of the Kronecker product.
It can be easily convinced by \eqref{eq:kron_vec} which is a combination of the outer product and vectorization.

It's a little complicated, but we have
\begin{align}
&\ten{Y} = \perm_{[2,4,1,3]}(\ten{X}), \\
&\ten{X} = \fold_{(P,I,Q,J)}(\mat{A} \otimes \mat{B}) \in \bbR{P \times I \times Q \times J}, \\
&\ten{Y} = \fold_{(I,J,P,Q)}(\unvec(\mat{A}) \circ \unvec(\mat{B})) \in \bbR{I \times J \times P \times Q}.
\end{align}
This suggests a relationship between the outer product and the Kronecker product.

\begin{figure}[t]
    \centering
    \includegraphics[width=0.7\textwidth]{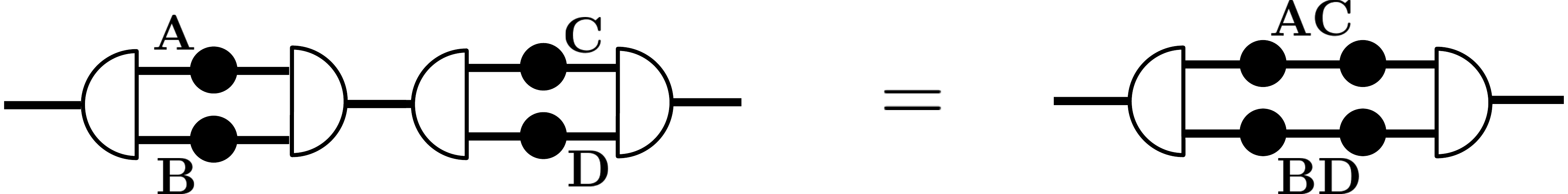}
    \caption{Proof of $(\mat{A} \otimes \mat{B})(\mat{C} \otimes \mat{D}) =  \mat{A}\mat{C} \otimes \mat{B}\mat{D}$.}\label{fig:proof_kron_identity}
\end{figure}

\paragraph{Properties of Kronecker product}
Typical identities for the Kronecker product are shown as follows:
\begin{align}
    \mat{A} \otimes \mat{B} & \neq  \mat{B} \otimes \mat{A}, \\
    c(\mat{A} \otimes \mat{B}) & =  (c\mat{A}) \otimes \mat{B} = \mat{A} \otimes (c\mat{B}), \\
\mat{A} \otimes (\mat{B} \otimes \mat{C}) & = (\mat{A} \otimes \mat{B}) \otimes \mat{C},\\
\mat{A} \otimes (\mat{B}+\mat{C}) & =  (\mat{A} \otimes \mat{B}) + (\mat{A} \otimes \mat{C}),\\
(\mat{A}+\mat{B}) \otimes \mat{C} & =  (\mat{A} \otimes \mat{C}) + (\mat{B} \otimes \mat{C}),\\
(\mat{A} \otimes \mat{B})(\mat{C} \otimes \mat{D}) & =  \mat{A}\mat{C} \otimes \mat{B}\mat{D}, \label{eq:kron_identity} \\
(\mat{A} \otimes \mat{B})^\top & =  \mat{A}^\top \otimes \mat{B}^\top, \\
(\mat{A} \otimes \mat{B})^{-1} & =  \mat{A}^{-1} \otimes \mat{B}^{-1}, \label{eq:kron_inv}\\
(\mat{A} \otimes \mat{B})^\dagger & =  \mat{A}^\dagger \otimes \mat{B}^\dagger,\label{eq:kron_pinv}
\end{align}
where the matrix inverse and the Moore-Penrose pseudo-inverse are denoted by $\cdot^{-1}$ and $\cdot^\dagger$, respectively.
Many of the above are easy to understand from the linearity of multiplication and definition of the Kronecker product, but \eqref{eq:kron_identity} may not be straightforward.
Nevertheless, the graphical notation makes \eqref{eq:kron_identity} very easy to prove as in Figure~\ref{fig:proof_kron_identity}.

\eqref{eq:kron_inv} and \eqref{eq:kron_pinv} are proved using \eqref{eq:kron_identity}.
For an example, \eqref{eq:kron_inv} can be verified as follow:
\begin{align}
(\mat{A}^{-1} \otimes \mat{B}^{-1}) (\mat{A} \otimes \mat{B}) &= \mat{A}^{-1} \mat{A} \otimes \mat{B}^{-1} \mat{B} \notag \\
&= \mat{I} \otimes \mat{I} = \mat{I}.
\end{align}
Multiplying $(\mat{A}^{-1} \otimes \mat{B}^{-1})$ on the right also results in an identity matrix
\begin{align}
 (\mat{A} \otimes \mat{B})(\mat{A}^{-1} \otimes \mat{B}^{-1}) &= \mat{A} \mat{A}^{-1} \otimes \mat{B} \mat{B}^{-1} \notag \\
&= \mat{I} \otimes \mat{I} = \mat{I},
\end{align}
then $(\mat{A}^{-1} \otimes \mat{B}^{-1})$ is the inverse matrix of $(\mat{A} \otimes \mat{B})$.
The case of pseudo-inverse can be proved in a similar way.

These identities can be generalized to cases where the Kronecker product appears multiple times, such as
\begin{align}
(\mat{A}_1 \otimes \cdots \otimes \mat{A}_{N})(\mat{B}_1 \otimes \cdots \otimes \mat{B}_{N}) &= (\mat{A}_1\mat{B}_1 \otimes \cdots \otimes \mat{A}_{N}\mat{B}_N), \\
(\mat{A}_1 \otimes \cdots \otimes \mat{A}_{N})^{-1} & =  \mat{A}_1^{-1} \otimes \cdots \otimes \mat{A}_N^{-1}.
\end{align}

\begin{figure}[t]
    \centering
    \includegraphics[width=0.7\textwidth]{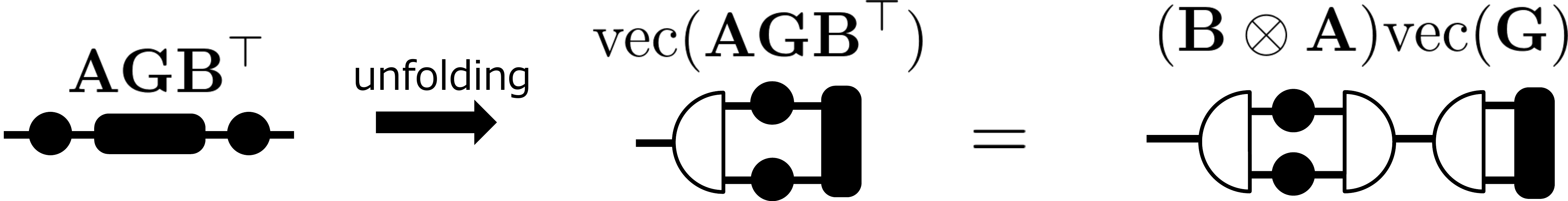}
    \caption{Proof of $\unvec(\mat{A}\mat{G}\mat{B}^\top) = (\mat{B} \otimes \mat{A}) \unvec(\mat{G})$.}\label{fig:kron2}
\end{figure}

Finally, we introduce the following identity:
\begin{align}
   \unvec(\mat{A}\mat{G}\mat{B}^\top) = (\mat{B} \otimes \mat{A}) \unvec(\mat{G}).\label{eq:kron_matrix_decomp}
\end{align}
This can be easily proved by using graphical notation as shown in Figure~\ref{fig:kron2}.
It is also possible to prove it using the formula.
Let us put $\mat{A} = [\vect{a}_1, ..., \vect{a}_J] \in \bbR{I \times J}$，$\mat{B} = [\vect{b}_1, ..., \vect{b}_Q] \in \bbR{P \times Q}$，and $\mat{G} \in \bbR{J \times Q}$, then we have
\begin{align}
   \mat{A}\mat{G}\mat{B}^\top = \sum_{j=1}^J \sum_{q=1}^Q G_{jq} (\vect{a}_j \circ \vect{b}_q) \in \bbR{I \times P}.
\end{align}
Its vectorization can be written by
\begin{align}
   \unvec(\mat{A}\mat{G}\mat{B}^\top) &= \sum_{j=1}^J \sum_{q=1}^Q G_{jq} \unvec(\vect{a}_j \circ \vect{b}_q) \notag\\ 
   &= \sum_{j=1}^J \sum_{q=1}^Q G_{jq} (\vect{b}_q \otimes \vect{a}_j) \notag\\ 
   &= (\mat{B} \otimes \mat{A}) \unvec(\mat{G}).
\end{align}

\begin{figure}[t]
    \centering
    \includegraphics[width=0.5\textwidth]{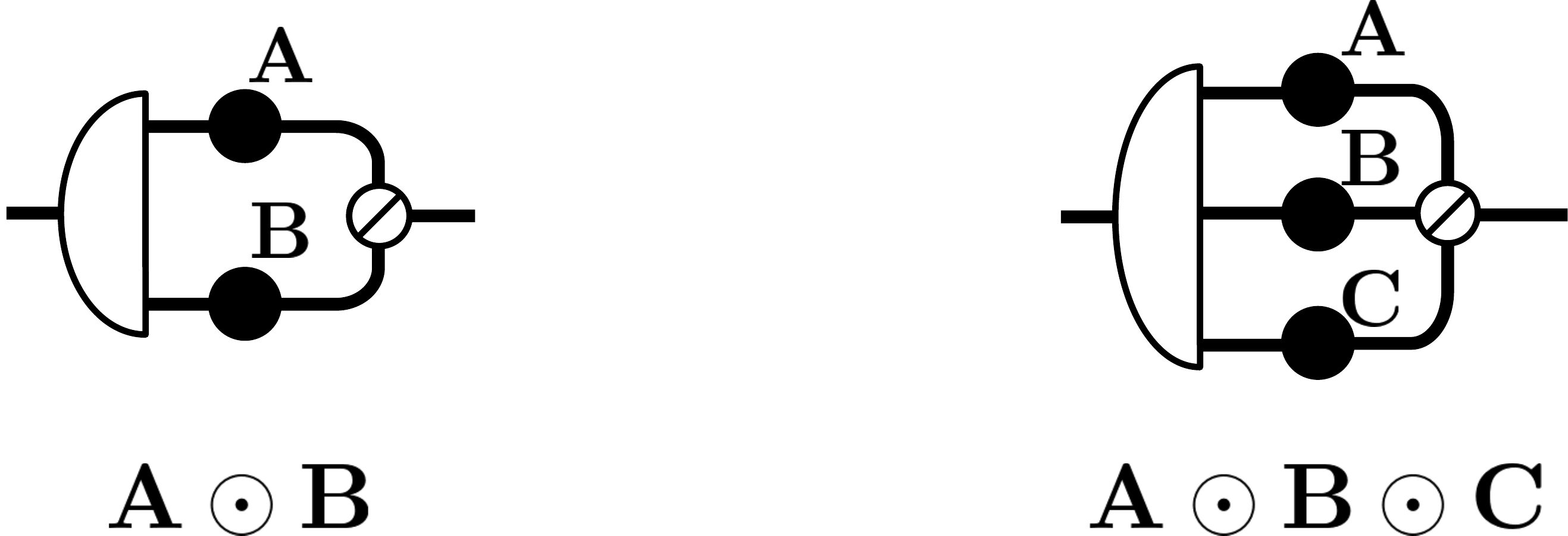}
    \caption{Khatri-Rao product.}\label{fig:KRprod}
\end{figure}

\subsection{Khatri-Rao product}\label{sec:khatri_rao}
{\em Khatri-Rao product} is defined for two matrices of which the same row length $\mat{A}=[\vect{a}_1, ..., \vect{a}_R] \in \bbR{I \times R}$，$\mat{B}=[\vect{b}_1, ..., \vect{b}_R] \in \bbR{J \times R}$ and given by
\begin{align}
    \mat{A} \odot \mat{B} = [\vect{a}_1 \otimes \vect{b}_1, ..., \vect{a}_R \otimes \vect{b}_R] \in \bbR{JI \times R}.
\end{align}
The Khatri-Rao product is an operation that the Kronecker product is performed for each column.
Figure~\ref{fig:KRprod} shows the diagrams of the Khatri-Rao product, where matrices are sandwiched between an unfolding operator and a Hadamard product operator.

\begin{figure}[t]
    \centering
    \includegraphics[width=0.95\textwidth]{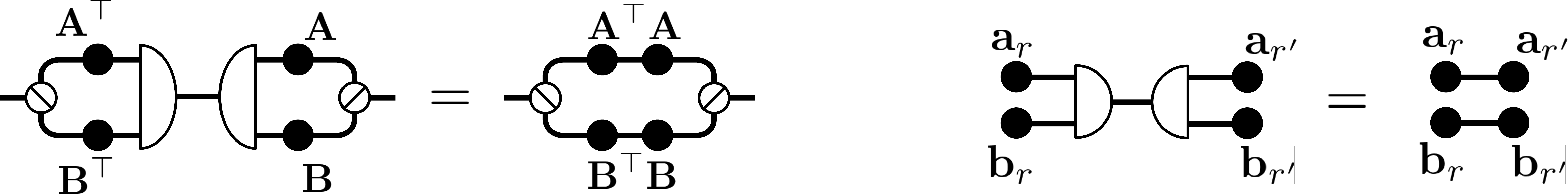}
    \caption{Proof of $(\mat{A} \odot \mat{B})^\top (\mat{A} \odot \mat{B}) = \mat{A}^\top\mat{A} \boxdot \mat{B}^\top\mat{B}$ and $\langle \vect{a}_r \otimes \vect{b}_r , \vect{a}_{r'} \otimes \vect{b}_{r'} \rangle 
   = \langle \vect a_r , \vect a_{r'} \rangle \langle \vect b_r , \vect b_{r'} \rangle.$}\label{fig:khatri}
\end{figure}

\paragraph{Properties of Khatri-Rao product}
Typical identities for the Khatri-Rao product are shown as follows:
\begin{align}
    \mat{A} \odot \mat{B} & \neq  \mat{B} \odot \mat{A}, \\
    c(\mat{A} \odot \mat{B}) & =  (c\mat{A}) \odot \mat{B} = \mat{A} \odot (c\mat{B}), \\
\mat{A} \odot (\mat{B} \odot \mat{C}) & = (\mat{A} \odot \mat{B}) \odot \mat{C},\\
\mat{A} \odot (\mat{B}+\mat{C}) & =  (\mat{A} \odot \mat{B}) + (\mat{A} \odot \mat{C}),\\
(\mat{A}+\mat{B}) \odot \mat{C} & =  (\mat{A} \odot \mat{C}) + (\mat{B} \odot \mat{C}).
\end{align}
In addition, we have
\begin{align}
(\mat{A} \odot \mat{B})^\top (\mat{A} \odot \mat{B}) & =  \mat{A}^\top \mat{A} \boxdot \mat{B}^\top \mat{B}. \label{eq:khatri_adamard}
\end{align}
Although \eqref{eq:khatri_adamard} is not straightforward, it can by verified by focusing on $(r,r')$-th entry as follow:
\begin{align}
\langle \vect{a}_r \otimes \vect{b}_r , \vect{a}_{r'} \otimes \vect{b}_{r'} \rangle 
   &= \sum_{i=1}^I \sum_{j=1}^J a_r(i) b_r(j) a_{r'}(i) b_{r'}(j) \notag \\
   &= \sum_{i=1}^I \sum_{j=1}^J a_r(i) a_{r'}(i) b_r(j) b_{r'}(j) \notag \\
   &= \left(\sum_{i=1}^I a_r(i) a_{r'}(i)\right) \left(\sum_{j=1}^J  b_r(j) b_{r'}(j)\right) \notag \\
   &= \langle \vect a_r , \vect a_{r'} \rangle \langle \vect b_r , \vect b_{r'} \rangle. \label{eq:khatri_adamard_2}
\end{align}
This can be easily understood by graphical notation as shown in Figure~\ref{fig:khatri}.
In more general, we have
\begin{align}
   \hspace{-2mm} (\mat{A}_1 \odot \cdots \odot \mat{A}_N)^\top (\mat{A}_1 \odot \cdots \odot \mat{A}_N) 
    =  \mat{A}_1^\top \mat{A}_1 \boxdot \cdots \boxdot \mat{A}_N^\top \mat{A}_N.
\end{align}

\subsection{Mode product}\label{sec:mode_product}
\begin{figure}[t]
    \centering
    \includegraphics[width=0.6\textwidth]{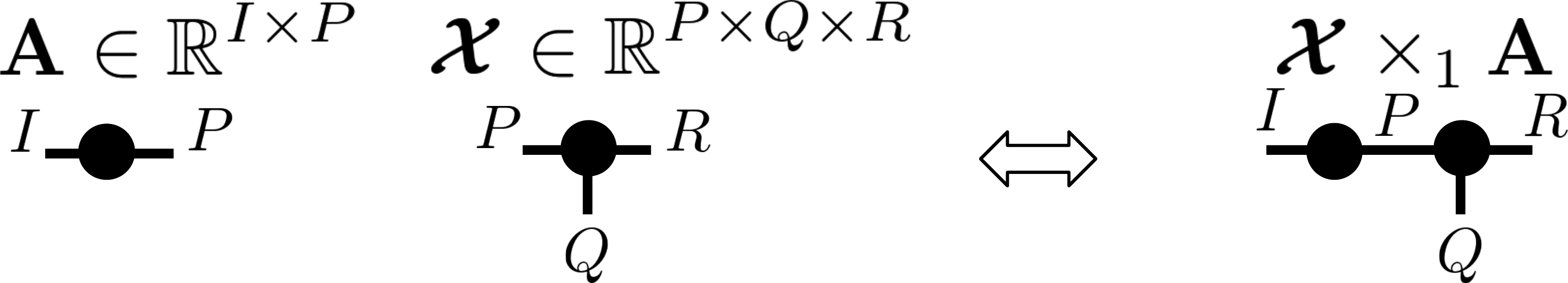}
    \caption{Mode product $\ten{X} \times_1 \mat{A}$.}\label{fig:mode_prod}
\end{figure}

{\em Mode product ($n$-mode or mode-$n$ product)} of a tensor $\ten{X}$ with a matrix $\mat{A}$ is denoted as
\begin{align}
\ten{X} \times_n \mat{A}. 
\end{align}
This is the operation of multiplying the $n$-th mode of a tensor $\ten{X}$ by a matrix $\mat{A}$.
An example of $\ten{X} \times_1 \mat{A}$ is shown in Figure~\ref{fig:mode_prod}.

Here, we consider the case of the third-order tensor for simplicity.
For a tensor $\ten{X} \in \bbR{P \times Q \times R}$ and a matrix $\mat{A} \in \bbR{I \times P}$, mode-1 product can be written as
\begin{align}
    \ten{Y} = \ten{X} \times_1 \mat{A} \in \bbR{I \times Q \times R}.
\end{align}
The mode of size $P$ common to the $(I,P)$-matrix and the $(P,Q,R)$-tensor is contracted, resulting in the $(I,Q,R)$-tensor.
The each entry of the resulting third-order tensor $\ten{Y} \in \bbR{I \times Q \times R}$ is given by
\begin{align}
\mathcal{Y}_{iqr} = \sum_{p=1}^P \mathcal{X}_{pqr} A_{ip}.
\end{align}
We calculate the inner product with the mode of index $p$, and the outer product with the modes of indices $i, q, r$.
Since the mode of index $p$ is the first mode for the tensor $\ten{X}$, we call it the mode-1 product and use the notation $\times_1$.
Similarly, let us consider matrices $\mat{B} \in \bbR{J \times Q}$ and $\mat{C} \in \bbR{K \times R}$, mode-2 product and mode-3 product are denoted as follows:
\begin{align}
&\ten{X} \times_2 \mat{B} \in \bbR{P \times J \times R}, \\
&\ten{X} \times_3 \mat{C} \in \bbR{P \times Q \times K}.
\end{align}

\begin{figure}[t]
    \centering
    \includegraphics[width=0.13\textwidth]{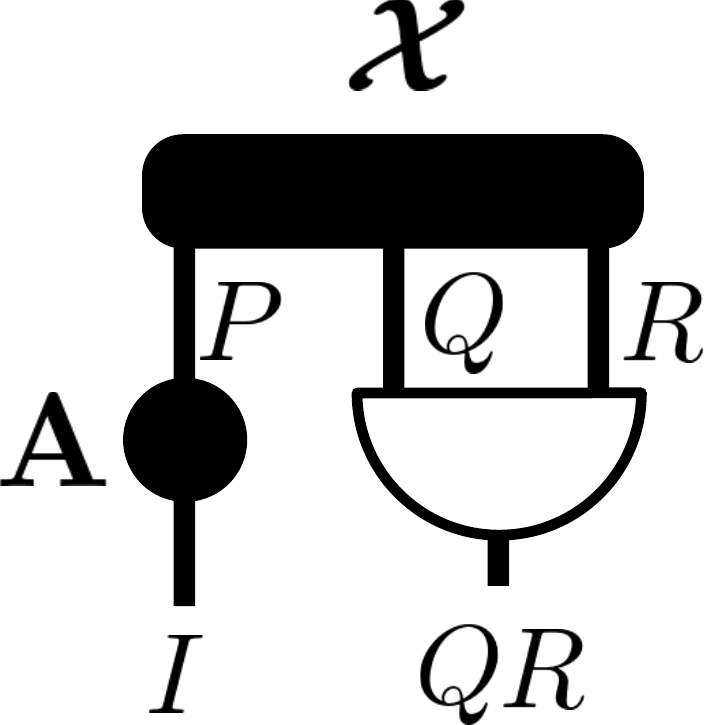}
    \caption{Proof of $\left[ \ten{X} \times_1 \mat{A} \right]_{(1)} = \mat{A} \mat{X}_{(1)}$.}\label{fig:mode_prod_mat}
\end{figure}

\paragraph{Properties of mode product}
Now, we introduce several important identities for the mode product.
First for the matricization of mode product, we have 
\begin{align}
    \left[ \ten{X} \times_1 \mat{A} \right]_{(1)} &= \mat{A} \mat{X}_{(1)} \in \bbR{I \times QR}, \\
    \left[ \ten{X} \times_2 \mat{B} \right]_{(2)} &= \mat{B} \mat{X}_{(2)} \in \bbR{J \times PR}, \\
    \left[ \ten{X} \times_3 \mat{C} \right]_{(3)} &= \mat{C} \mat{X}_{(3)} \in \bbR{K \times PQ}.
\end{align}
We can see this clearly in the diagram as shown in Figure~\ref{fig:mode_prod_mat}.

\begin{figure}[t]
    \centering
    \includegraphics[width=0.45\textwidth]{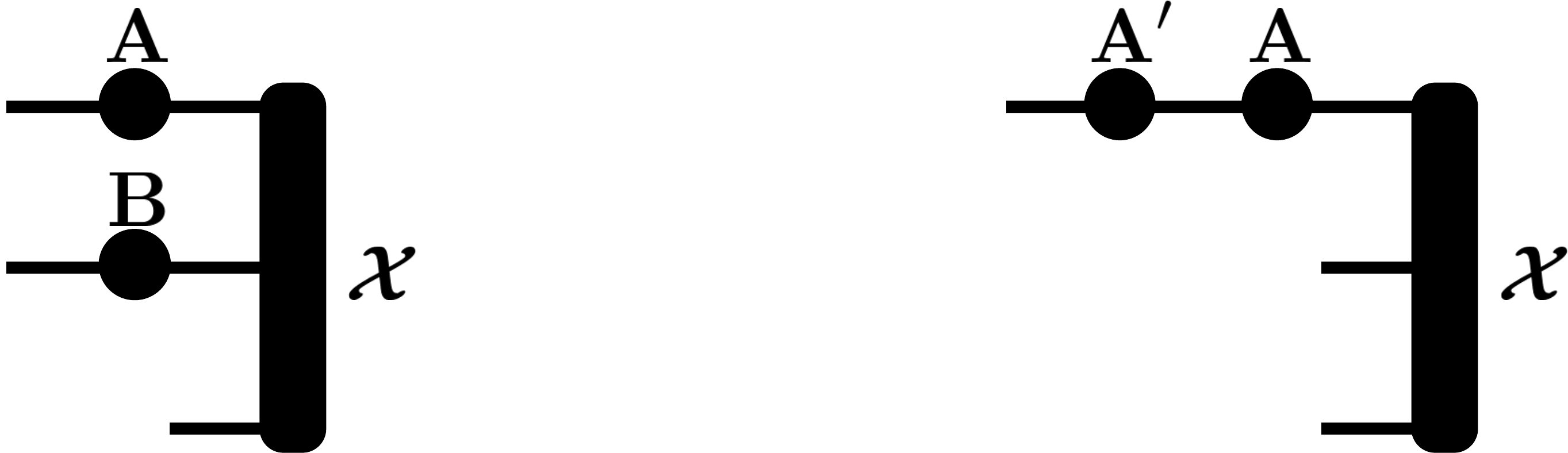}
    \caption{Diagrams of $\ten{X} \times_1 \mat{A} \times_2 \mat{B}$ and $(\ten{X} \times_1 \mat{A}) \times_1 \mat{A}'$.}\label{fig:mode_prod_iden2}
\end{figure}

In addition, we have
\begin{align}
   (\ten{X} \times_1 \mat{A}) \times_2 \mat{B} & = (\ten{X} \times_2 \mat{B}) \times_1 \mat{A} = \ten{X} \times_1 \mat{A} \times_2 \mat{B},\label{eq:mode_prod_rule1}\\
   (\ten{X} \times_1 \mat{A}) \times_1 \mat{A}' & = \ten{X} \times_1 \mat{A}'\mat{A}. \label{eq:mode_prod_rule2}
\end{align}
Regarding \eqref{eq:mode_prod_rule1}, this is similar to the fact that for the matrix product of three matrices, $\mat{A}\mat{G}\mat{B} = \mat{A}(\mat{G}\mat{B}) = (\mat{A}\mat{G})\mat{B}$ holds.
The order does not matter when the mode products are performed from different modes.
However, the order is important when the mode products are performed from a single mode sequentially.
\eqref{eq:mode_prod_rule2} can be understood in the form of its matricization $\mat{A}'\mat{A}\mat{X}_{(1)} \neq \mat{A}\mat{A}'\mat{X}_{(1)}$.
Figure~\ref{fig:mode_prod_iden2} also helps to understand.

\paragraph{All-mode product}
The mode product of a third-order tensor $\ten{X}$ by three matrices $\mat{A}$, $\mat{B}$, and $\mat{C}$ from all modes is called the {\em all-mode product}.
The all-mode product can be expressed simply by using mode products $\times_n$ or by using $\llbracket \rrbracket$:
\begin{align}
 \ten{X} \times_1 \mat{A} \times_2 \mat{B} \times_3 \mat{C} =
\llbracket{\ten{X}}; \mat{A}, \mat{B}, \mat{C} \rrbracket \in \bbR{I \times J \times K}.
\end{align}
Each entry of all-mode product is given by
\begin{align}
[ \mathcal{X} \times_1 A \times_2 B \times_3 C ]_{ijk} = \sum_{p = 1}^P \sum_{q=1}^Q \sum_{r=1}^R \mathcal{X}_{pqr} A_{ip} B_{jq} C_{kr}.
\end{align}
Let be $\mat{A} = [\vect{a}_1, ..., \vect{a}_P]$, $\mat{B} = [\vect{b}_1, ..., \vect{b}_Q]$, and $\mat{C} = [\vect{c}_1, ..., \vect{c}_R]$, then all-mode product can be also represented as a linear combinations of outer products:
\begin{align}
\ten{X} \times_1 \mat{A} \times_2 \mat{B} \times_3 \mat{C} = \sum_{p = 1}^P \sum_{q=1}^Q \sum_{r=1}^R \mathcal{X}_{pqr} \vect{a}_{p} \circ \vect{b}_{q} \circ \vect{c}_{r}.
\end{align}
Vectorizing the above representation gives us an extension of \eqref{eq:kron_matrix_decomp} as follow:
\begin{align}
\unvec(\ten{X} \times_1 \mat{A} \times_2 \mat{B} \times_3 \mat{C}) = (\mat{C} \otimes \mat{B} \otimes \mat{A}) \unvec(\ten{X}).
\end{align}

Note that all-mode product with identity matrices does not change the tensor, then we have
\begin{align}
\ten{X} = \ten{X} \times_1 \mat{I} = \ten{X} \times_1 \mat{I} \times_2 \mat{I} = \ten{X} \times_1 \mat{I} \times_2 \mat{I} \times_3 \mat{I}.
\end{align}
Hence, the vectorization of single mode product can be represented by
\begin{align}
   \unvec(\ten{X} \times_1 \mat{A}) &= (\mat{I} \otimes \mat{I} \otimes \mat{A}) \unvec(\ten{X}), \\
   \unvec(\ten{X} \times_2 \mat{B}) &= (\mat{I} \otimes \mat{B} \otimes \mat{I}) \unvec(\ten{X}), \\
   \unvec(\ten{X} \times_3 \mat{C}) &= (\mat{C} \otimes \mat{I} \otimes \mat{I}) \unvec(\ten{X}).
\end{align}

\begin{figure}[t]
    \centering
    \includegraphics[width=0.8\textwidth]{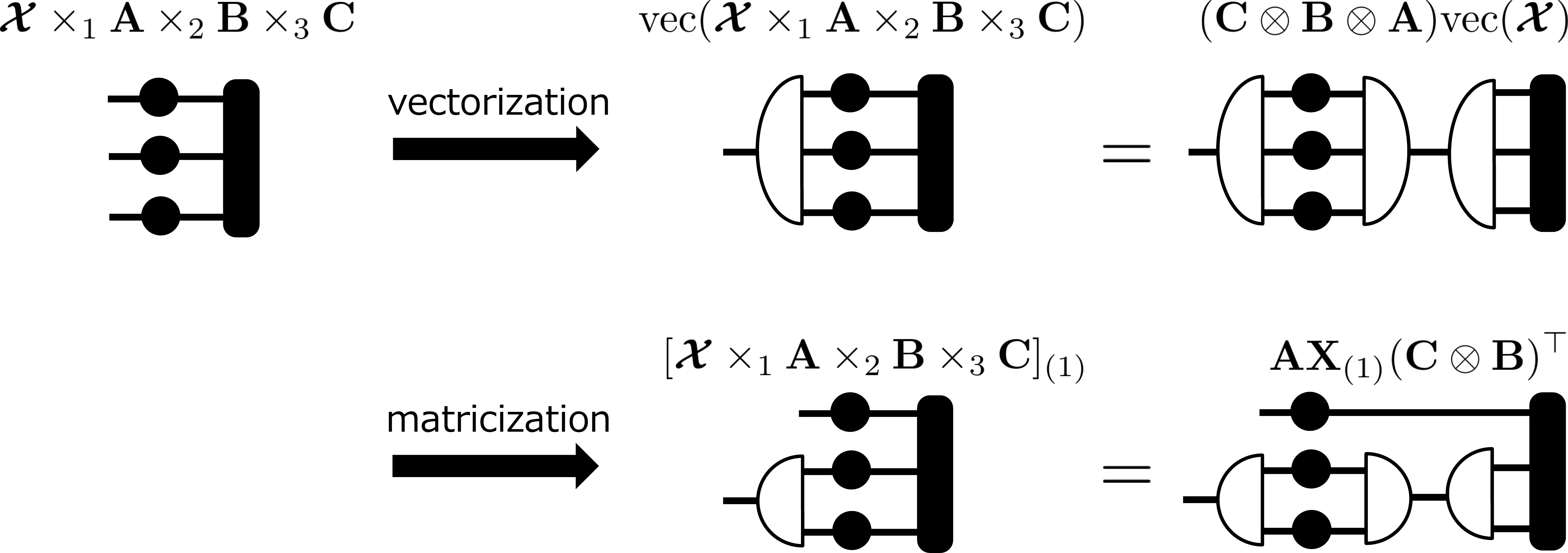}
    \caption{Vectorization and matricization of all-mode product}\label{fig:all_mode_unfold}
\end{figure}

In addition, the matricization of all-mode product can be representad by
\begin{align} 
    \left[ \ten{X} \times_1 \mat{A} \times_2 \mat{B} \times_3 \mat{C} \right]_{(1)} &= \mat{A} \mat{X}_{(1)}(\mat{C} \otimes \mat{B})^\top, \\
    \left[ \ten{X} \times_1 \mat{A} \times_2 \mat{B} \times_3 \mat{C} \right]_{(2)} &= \mat{B} \mat{X}_{(2)}(\mat{C} \otimes \mat{A})^\top, \\ 
    \left[ \ten{X} \times_1 \mat{A} \times_2 \mat{B} \times_3 \mat{C} \right]_{(3)} &= \mat{C} \mat{X}_{(3)}(\mat{B} \otimes \mat{A})^\top.  
\end{align}
Figure~\ref{fig:all_mode_unfold} shows the diagrams for vectorization and matricization of the all-mode product.

\paragraph{Mode product for $N$th-order tensors}
Finally, we introduce the mode product for the $N$th-order tensor $\ten{G} \in \bbR{R_1 \times R_2 \times \cdots \times R_N}$ with a matrix $\mat{A}^{(n)} \in \bbR{I_n \times R_n}$.
It is denoted by 
\begin{align}
    \ten{G} \times_n \mat{A}^{(n)} \in \bbR{R_1 \times \cdots \times R_{n-1} \times I_n \times R_{n+1} \times \cdots \times R_{N}},
\end{align}
and its matricization is given by
\begin{align}
  \left[ \ten{G} \times_n \mat{A}^{(n)} \right]_{(n)} = \mat{A}^{(n)} \mat{G}_{(n)} \in \bbR{I_n \times \prod_{k \neq n}R_k}.
\end{align}

The all-mode product with $N$ matrices $\mat{A}^{(1)} \in \bbR{I_1 \times R_1}$，$\mat{A}^{(2)} \in \bbR{I_2 \times R_2}$，...，$\mat{A}^{(N)} \in \bbR{I_N \times R_N}$ can be written in three different notations as
\begin{align}
    &\ten{G} \times_1 \mat{A}^{(1)} \times_2 \mat{A}^{(2)} \cdots \times_N \mat{A}^{(N)} \notag \\
     &=\llbracket \ten{G} ; \mat{A}^{(1)}, \mat{A}^{(2)}, ..., \mat{A}^{(N)} \rrbracket \\
    &=\ten{G} \times \left\{ \mat{A} \right\}. \notag 
\end{align}
The notation $\ten{G} \times \left\{ \mat{A} \right\}$ is concise and is often used when considering a general case of $N$th-order tensors.

Vectorization and matricization of the all-mode product for $N$th-order tensor can by given by
\begin{align}
&\ten{Y} = \ten{G} \times_1 \mat{A}^{(1)} \times_2 \mat{A}^{(2)} \cdots \times_N \mat{A}^{(N)} \notag \\
 &\Leftrightarrow \unvec(\ten{Y}) = (\mat{A}^{(N)} \otimes \cdots \otimes \mat{A}^{(2)} \otimes \mat{A}^{(1)}) \unvec(\ten{G}) \\
 &\Leftrightarrow \mat{Y}_{(n)} = \mat{A}^{(n)} \mat{G}_{(n)} (\mat{A}^{(N)} \otimes \cdots \otimes \mat{A}^{(n+1)} \otimes \mat{A}^{(n-1)} \otimes \cdots \otimes \mat{A}^{(1)})^\top. \notag
\end{align}

The all-mode product except the $n$-th mode is also often considered.
It is denoted by using $\times_{-n}$ as
\begin{align}
\ten{G} \times_{-n} \left\{ \mat{A} \right\} = \ten{G}  \times_1 \mat{A}^{(1)} \cdots \times_{n-1} \mat{A}^{(n-1)} \times_{n+1} \mat{A}^{(n+1)} \cdots \times_N \mat{A}^{(N)}.
\end{align}
Let us put $\ten{Y} = \ten{G} \times \left\{ \mat{A} \right\}$ and $\ten{X} = \ten{G} \times_{-n} \left\{ \mat{A} \right\}$, we have
\begin{align}
    \mat{Y}_{(n)} = \mat{A}^{(n)} \mat{X}_{(n)}.
\end{align}
This matrix formulation is important for considering the optimization problem for the tensor decompositions.

\subsection{Tensor product}\label{sec:tensor_product}
{\em Tensor product} is a generalization of the mode product.
Whereas the mode product considers the multiplication of a tensor by a matrix, the tensor product considers the multiplication of a tensor by a tensor.

\begin{figure}[t]
    \centering
    \includegraphics[width=0.95\textwidth]{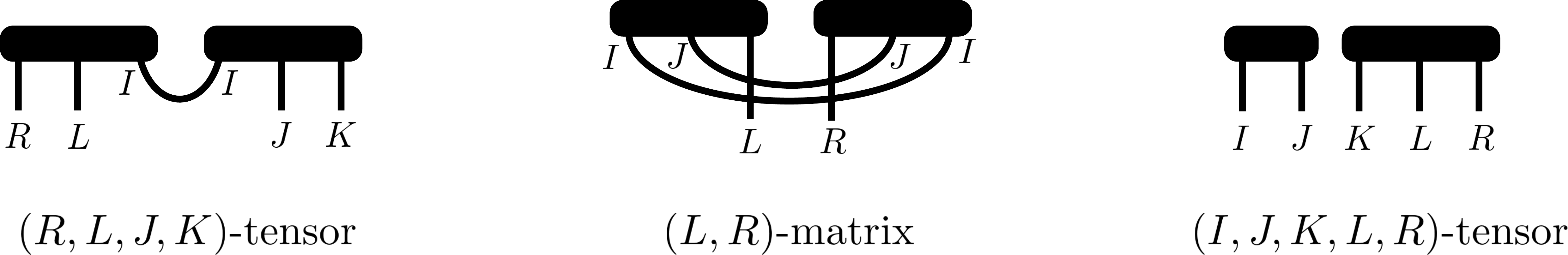}
    \caption{Examples of tensor prduct.}\label{fig:tensor_prod_ex}
\end{figure}

For an example, we consider the following mode product of a third-order tensor $\ten{X} \in \bbR{P \times Q \times R}$ with a matrix $\mat{A} \in \bbR{I \times P}$:
\begin{align}
    \ten{X} \times_1 \mat{A} \in \bbR{I \times Q \times R}.
\end{align}
In this case, the inner product is performed with respect to the first mode of $\ten{X}$ and the second mode of $\mat{A}$.
However, the notation $\times_1$ only conveys the information about the first mode of $\ten{X}$, not the information about the second mode of $\mat{A}$.
A clearer, unabbreviated notation would be:
\begin{align}
    \mat{A} \mathbin{_2\times^1} \ten{X} = \ten{X} \times_1 \mat{A} \in \bbR{I \times Q \times R},
\end{align}
where subscript `2' of $\mathbin{_2\times^1}$ stands for the second mode of the left operand tensor $\mat{A} \in \bbR{I \times P}$ and superscript `1' of $\mathbin{_2\times^1}$ stands for the first mode of the right operand tensor $\ten{X} \in \bbR{P \times Q \times R}$.
This notation can be understood more intuitively as 
\begin{align}
(I,P)\text{-matrix} \mathbin{_2\times^1} (P,Q,R)\text{-tensor} = (I,Q,R)\text{-tensor}.
\end{align}
More examples are as follows:
\begin{align}
&(R,L,I)\text{-tensor} \mathbin{_3\times^1} (I,J,K)\text{-tensor} = (R,L,J,K)\text{-tensor}, \label{eq:tensorprod_ex1} \\
&(I,J,L)\text{-tensor} \mathbin{_{1,2}\times^{3,2}} (R,J,I)\text{-tensor} = (L,R)\text{-matrix}, \label{eq:tensorprod_ex2}\\
&(I,J)\text{-matrix} \mathbin{\times} (K,L,R)\text{-tensor} = (I,J,K,L,R)\text{-tensor}.\label{eq:tensorprod_ex3}
\end{align}
In \eqref{eq:tensorprod_ex1}, the inner product is performed with the mode of length $I$ and a fourth-order tensor is constructed with the remaining modes. 
In \eqref{eq:tensorprod_ex2}, the inner product is performed with two modes of length $I$ and length $J$ and a matrix is constructed with the remaining modes.
In \eqref{eq:tensorprod_ex3}, no inner product is performed, but a fifth-order tensor is constructed as the result of the outer product.
Figure~\ref{fig:tensor_prod_ex} shows above three examples of tensor product in graphical notation.

\paragraph{General concept of tensor product}
The tensor product can be thought of as an operation of taking the inner product (contraction) with respect to pairs of modes of tensors.
Let us consider an $N$th-order tensor $\ten{A} \in \bbR{I_1 \times \cdots \times I_N}$ and a $M$th-order tensor $\ten{B} \in \bbR{J_1 \times \cdots \times J_M}$, and there are $K$ pairs of modes $(n_1,m_1)$, ..., $(n_K,m_K)$ which satisfy
\begin{align}
    &1 \leq n_k \leq N, \ \ k \in \{1, ..., K\},\\
    &1 \leq m_k \leq M, \ \ k \in \{1, ..., K\},\\   
    &n_1 \neq \cdots \neq n_K,\\
    &m_1 \neq \cdots \neq m_K,\\
    &I_{n_k} = J_{m_k}, \ \ k \in \{1, ..., K\},
\end{align}
then the following tensor product 
\begin{align}
    \ten{C} = \ten{A} \mathbin{_{n_1, ..., n_K}\times^{m_1, ..., m_K}} \ten{B}
\end{align}
can be defined.
In this case, $\ten{C}$ is the $(N+M-2K)$th-order tensor.

\paragraph{Tensor-train (TT) product}
{\em Tensor-train (TT) product} is an operation that connects the final mode of the first operand tensor $\ten{X}$ with the first mode of the second operand tensor $\ten{Y}$ and it is denoted as
\begin{align}
  \llangle \ten{X}, \ten{Y} \rrangle.
\end{align}
In case of that the TT product of a $(M+1)$th-order tensor $ \ten{X} \in \bbR{I_1 \times \cdots \times I_M \times R}$ with $(N+1)$th-order tensor $\ten{Y} \in \bbR{R \times J_1 \times \cdots \times J_N}$ can be written as
\begin{align}
   \llangle \ten{X}, \ten{Y} \rrangle = \ten{X} \mathbin{_{M+1}\times^1} \ten{Y} \in \bbR{I_1 \times \cdots \times I_M \times J_1 \times \cdots \times J_N}
\end{align}
using the symbol of tensor product.

\begin{figure}[t]
  \centering
  \includegraphics[width=0.9\textwidth]{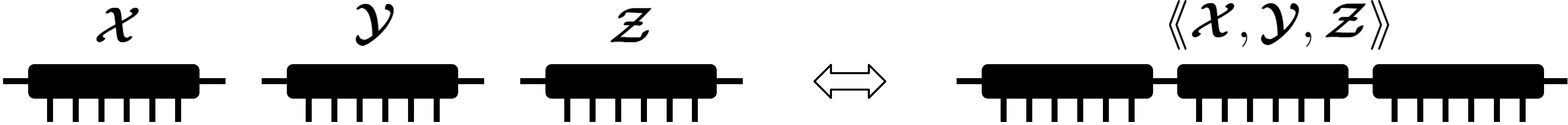}
  \caption{TT product.}\label{fig:TTprod}
\end{figure}

When $\llangle \ten{Y}, \ten{Z} \rrangle$ is computable, then
\begin{align}
  \llangle \ten{X}, \ten{Y}, \ten{Z} \rrangle = \llangle \ten{X}, \llangle \ten{Y}, \ten{Z} \rrangle \rrangle = \llangle \llangle \ten{X}, \ten{Y} \rrangle , \ten{Z} \rrangle
\end{align}
is also computable.
We write this by $\llangle \ten{X}, \ten{Y}, \ten{Z} \rrangle$ since the result of the tensor product does not depend on the order of calculation.
Figure~\ref{fig:TTprod} shows the TT product in diagrams.
TT product is a key building block in the TT decomposition which is discussed in Section~\ref{sec:TT-decomposition}.

\subsection{Tensor networks}
\begin{figure}[t]
    \centering
    \includegraphics[width=0.55\textwidth]{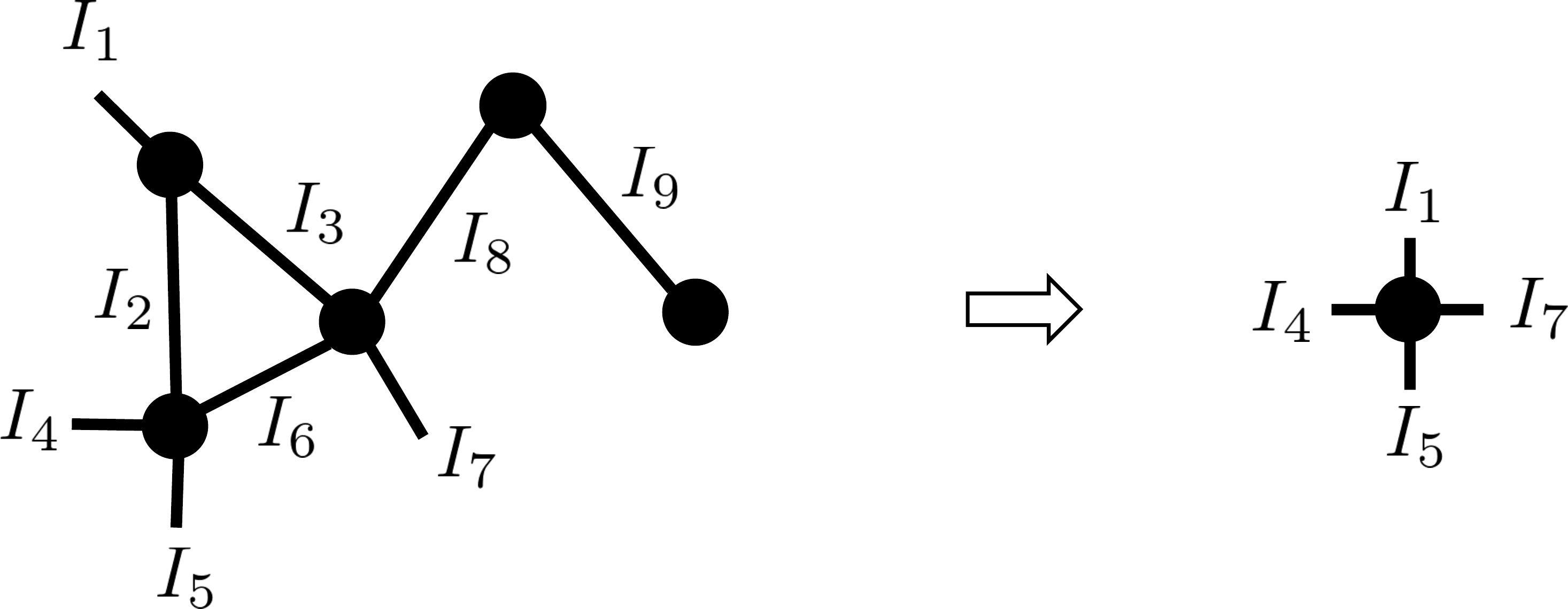}
    \caption{A tensor network and its evaluation.}\label{fig:TN}
\end{figure}

\begin{figure}[t]
    \centering
    \includegraphics[width=0.95\textwidth]{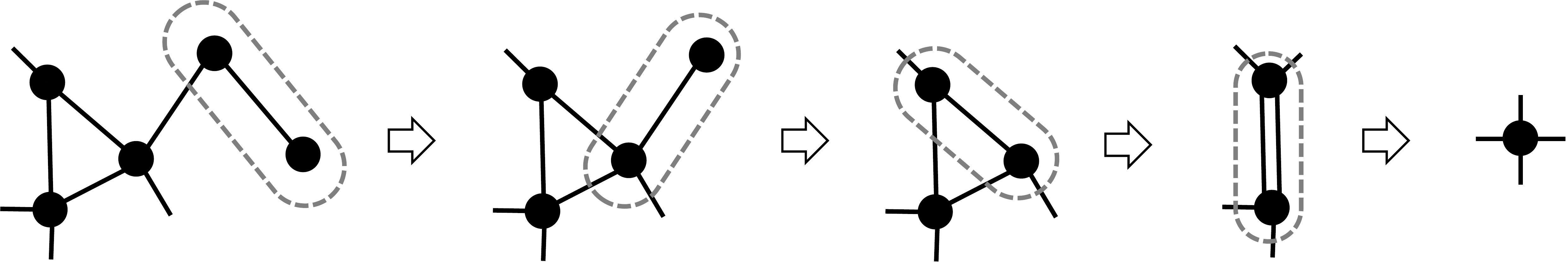}
    \caption{A step-by-step procedure for evaluating a tensor network. The two tensors that are contracted by a tensor product are enclosed in a dotted line.}\label{fig:TN2}
\end{figure}

{\em Tensor networks} (TN) are combinations of tensor products of two or more tensors.
In graphical notation, tensor products are represented as edge connections, so they can be thought of as networks.
{\em Evaluation} of TN is to perform a calculation of the tensor product and to obtain the value of a single tensor as a result.
Figure~\ref{fig:TN} shows an example of a tensor network and the tensor obtained after evaluation.
As a result, the coupled modes disappear, and uncoupled (free) modes remain.
Figure~\ref{fig:TN2} shows an example of the step-by-step procedure to evaluate this TN.
The tensor product (contraction) of two tensors, which contracted them into a single tensor, is performed repeatedly.


Two important properties of evaluating tensor networks are:
\begin{enumerate}
\item {\bf the order of contractions does not change the resulting tensor},
\item {\bf the order of contractions changes the computational cost}.
\end{enumerate}
A simple example of property 1 and 2 is
\begin{align}
  \mat A \mat B \vect v,
\end{align}
where $\mat{A} \in \bbR{I \times I}$, $\mat{B} \in \bbR{I \times I}$, and $\vect{v} \in \bbR{I}$.
Clearly, we have $(\mat A \mat B) \vect v = \mat A (\mat B \vect v)$ and the calculation procedure does not change the results.
However, the computational complexities are $\mathcal{O}(I^3)$ for $(\mat A \mat B) \vect v$, and $\mathcal{O}(I^2)$ for $\mat A (\mat B \vect v)$.

\begin{figure}[t]
    \centering
    \includegraphics[width=0.8\textwidth]{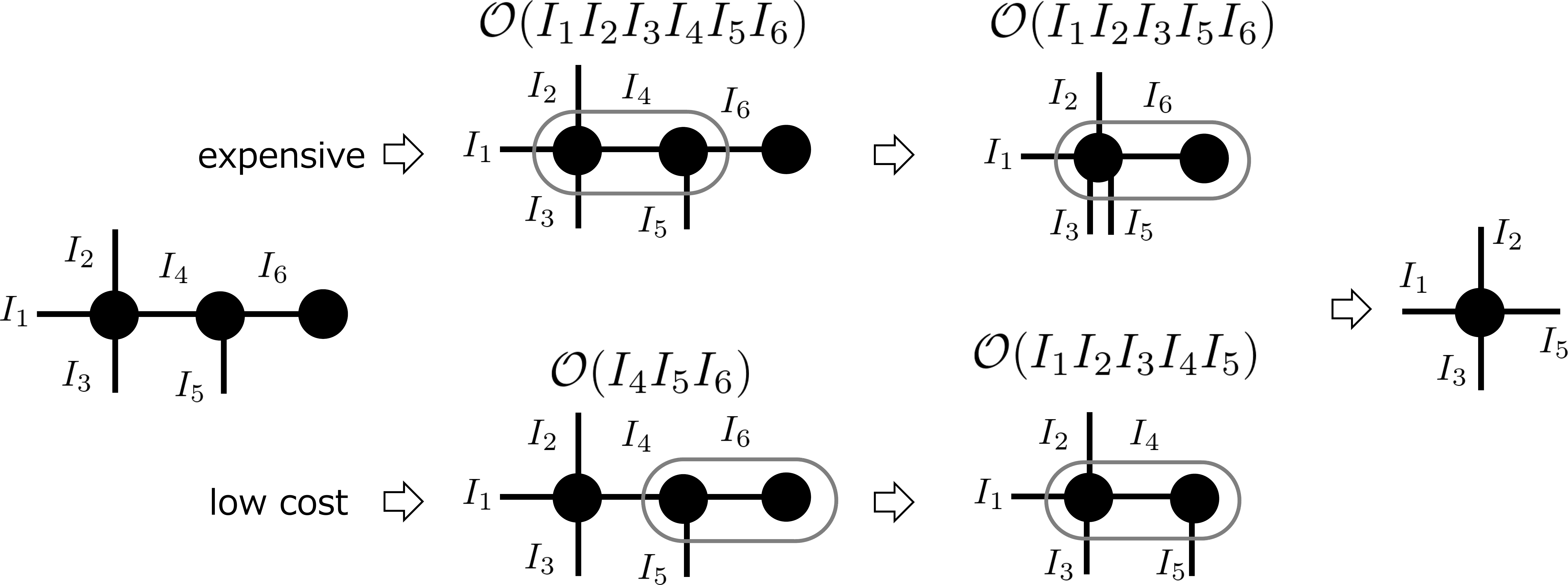}
    \caption{Computational cost for evaluating a TN.}\label{fig:TN3}
\end{figure}

Let us consider a slightly more complicated case as shown in Figure~\ref{fig:TN3}.
In this figure, we are trying to evaluate a TN consisting of three tensors, and in the top procedure they are contracted from the left, and in the bottom procedure they are contracted from the right.
First, the basic fact is that {\bf the computational complexity of a contraction of two tensors is given by the product of the lengths of all the modes involved}.
Following this rule, we can obtain the computational complexity of each procedure.
The maximum computational complexity of the top procedure is $\mathcal{O}(I_1 I_2 I_3 I_4 I_5 I_6)$, whereas in the bottom procedure it is $\mathcal{O}(I_1 I_2 I_3 I_4 I_5)$.
In this case, the computational cost can be saved by using right-side contraction.

\clearpage
\newpage

\section{Tensor decompositions}
{\em Tensor decomposition} is the operation of decomposing a single tensor into a form of tensor product of multiple tensors (i.e., tensor network).
This can be thought of as the inverse operation of evaluating TNs.
Although TN evaluation involves only calculations, tensor decomposition usually involves optimization.
Let us consider to find some tensor network $\text{TN}(\ten{G}_1, ..., \ten{G}_M)$ consisting of $M$ tensors to approximate a given tensor $\ten{Y}$.
The least-squares problem for tensor decomposition can be given as
\begin{align}
\minimize_{\ten{G}_1, ..., \ten{G}_M} || \ten{Y} - \text{TN}(\ten{G}_1, ..., \ten{G}_M) ||_F^2.
\end{align}
{\em Alternating least squares} (ALS) is the simplest algorithm to solve the above problem.
In the ALS, the following sub-optimization problems
\begin{align}
\ten{G}_n \leftarrow \argmin_{\ten{G}_n} || \ten{Y} - \text{TN}(\ten{G}_1, ..., \ten{G}_M) ||_F^2,
\end{align}
for $n \in \{1, 2, ..., N\}$ are repeatedly solved until convergence.
Here, we will not go into details of optimization but will introduce some typical decompositions and their properties using diagrams.

\subsection{QR decomposition}
\begin{figure}[t]
    \centering
    \includegraphics[width=0.8\textwidth]{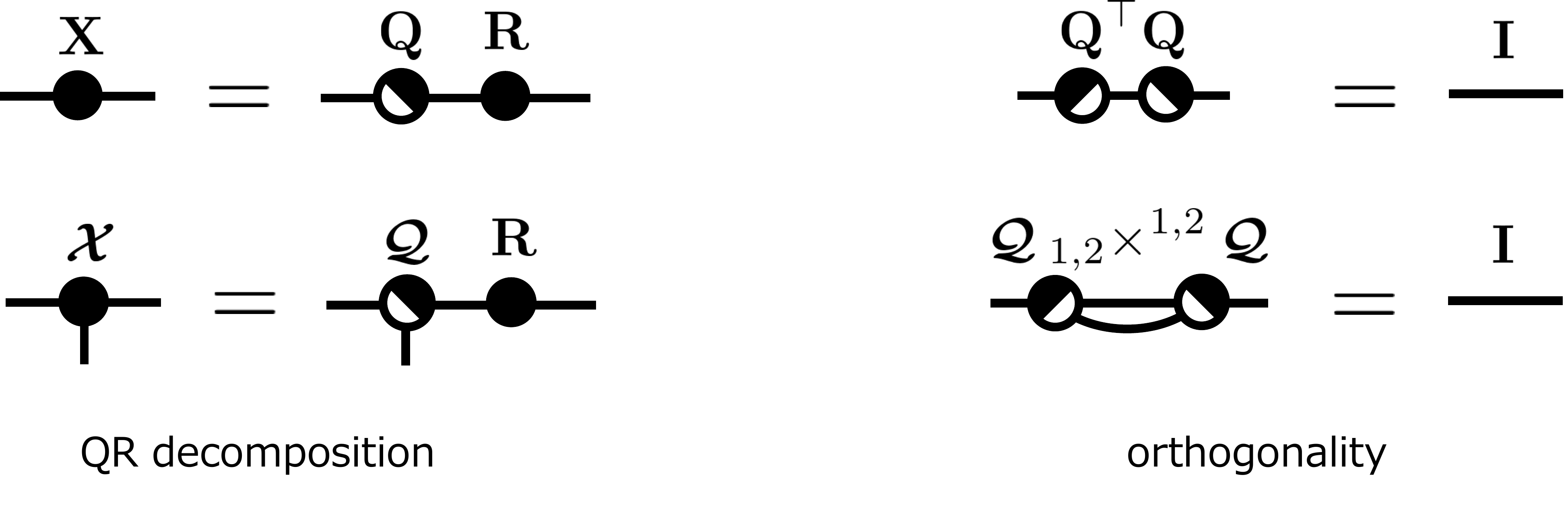}
    \caption{QR decompositions of a matrix and a tensor.}\label{fig:QR}
\end{figure}

{\em QR decomposition} is the operation of decomposing a tall (or square) matrix $\mat{X} \in \bbR{I \times J}$ into a column orthogonal matrix $\mat{Q} \in \bbR{I \times J}$ and an upper triangular matrix $\mat{R} \in \bbR{J \times J}$ as
\begin{align}
\mat{X} = \mat{Q} \mat{R},
\end{align}
where $I \geq J$ and $\mat{Q}^\top\mat{Q} = \mat{I}$.
There are several algorithms for performing QR decomposition, such as the Gram-Schmidt process and the Householder transform, but we will not go into detail here and instead focus on the fact that {\bf QR decomposition is possible for any tall matrix}\footnote{QR decomposition is unique when we assume $\rank(\bm X) = J$ and diagonal entries of $\mat{R}$ are positive \cite{horn2012matrix}.}.
The upper row of Figure~\ref{fig:QR} shows the diagram of QR decomposition for the matrix.
The column orthogonal matrix is represented by black and white nodes, and when the edges of the white side are contracted, they become an identity matrix due to orthogonality.

\paragraph{QR decompositions for tensors}
In addition, let us consider an $N$th-order tensor $\ten{X} \in \bbR{I_1 \times \cdots \times I_N}$ and its QR decomposition.
Although QR decomposition is an operation on matrices, it can also be applied to tensors through matricization.
In the form of mode-$n$ matricization, QR decomposition of a tensor $\ten{X}$ can be written by
\begin{align}
&\mat{X}_{(n)}^\top = \mat{Q}_{(n)}^\top \mat{R} \\
&\Leftrightarrow \ten{X} = \ten{Q} \mathbin{_{n}\times^{1}} \mat{R},
\end{align}
where $\ten{Q} \in \bbR{I_1 \times \cdots \times I_N}$, $\mat{R} \in \bbR{I_n \times I_n}$, and $\mat{Q}_{(n)}\mat{Q}_{(n)}^\top = \mat{I}$.
In general, we have $|| \ten{X} ||_F = || \mat{R} ||_F$.
The bottom row of Figure~\ref{fig:QR} shows the diagram of QR decomposition for the third-order tensor.
The orthogonal tensor is represented by a black and white node, and when the edges of the white side are contracted, they become an identity matrix due to its orthogonality.

\subsection{Singular value decomposition}
{\em Singular value decomposition} (SVD)\footnote{Strictly speaking, there is a distinction between {\em full SVD} and {\em economy SVD}. Note that this text introduces the ``economy SVD'' 
 as SVD for its compact representation.} is the operation that decomposes a matrix $\mat{X} \in \bbR{I \times J}$ into a form of matrix product of a column orthogonal matrix $\mat{U} \in \bbR{I \times R}$, a diagonal matrix $\mat{\Sigma} \in \bbR{R \times R}$, and a row orthogonal matrix $\mat{V}^\top \in \bbR{R \times J}$ as
\begin{align}
\mat{X} = \mat{U} \mat{\Sigma} \mat{V}^\top,
\end{align}
where $R = \min(I,J)$, $\mat{U}^\top \mat{U} = \mat{I}$, and $\mat{V}^\top \mat{V} = \mat{I}$.
Note that {\bf SVD is possible for any matrix}.
The diagonal entries $\Sigma_{rr} = \sigma_r$, called {\em singular values}, are non-negative and arranged in descending order as:
\begin{align}
  \sigma_1 \geq \sigma_2 \geq \cdots \geq \sigma_R \geq 0,
\end{align}
and these are uniquely determined by $\mat{X}$.

\paragraph{Rank of a matrix}
Let $\mat{U} = [\vect{u}_1, ..., \vect{v}_R]$ and $\mat{V} = [\vect{v}_1, ..., \vect{v}_R]$, then the SVD can be written as
\begin{align}
\mat{X} = \mat{U} \mat{\Sigma} \mat{V}^\top = \sum_{r=1}^R \sigma_r \vect{u}_r \circ \vect{v}_r. \label{eq:SVD_sum_of_outer_product}
\end{align}
That is, $\mat{X}$ is represented as the sum of outer products.
A matrix that can be expressed as the outer product of non-zero vectors, such as $\vect{u}_r \circ \vect{v}_r$, is called a {\em rank-1 matrix}.
Additionally, {\em rank} of a matrix refers to the minimum number of rank-1 matrices needed to represent the matrix as a sum of rank-1 matrices.
That is, the number of nonzero (positive) singular values of $\mat{X}$ is equivalent to the rank of $\mat{X}$.
Therefore, we have
\begin{align}
  \rank(\mat{X}) = \sum_{r=1}^R \delta(\sigma_r),
\end{align}
where
\begin{align}
\delta(z) = \left\{\begin{array}{ll} 1 & (z \neq 0)\\ 0 & (z = 0) \end{array}\right. .
\end{align}

\begin{figure}[t]
    \centering
    \includegraphics[width=0.9\textwidth]{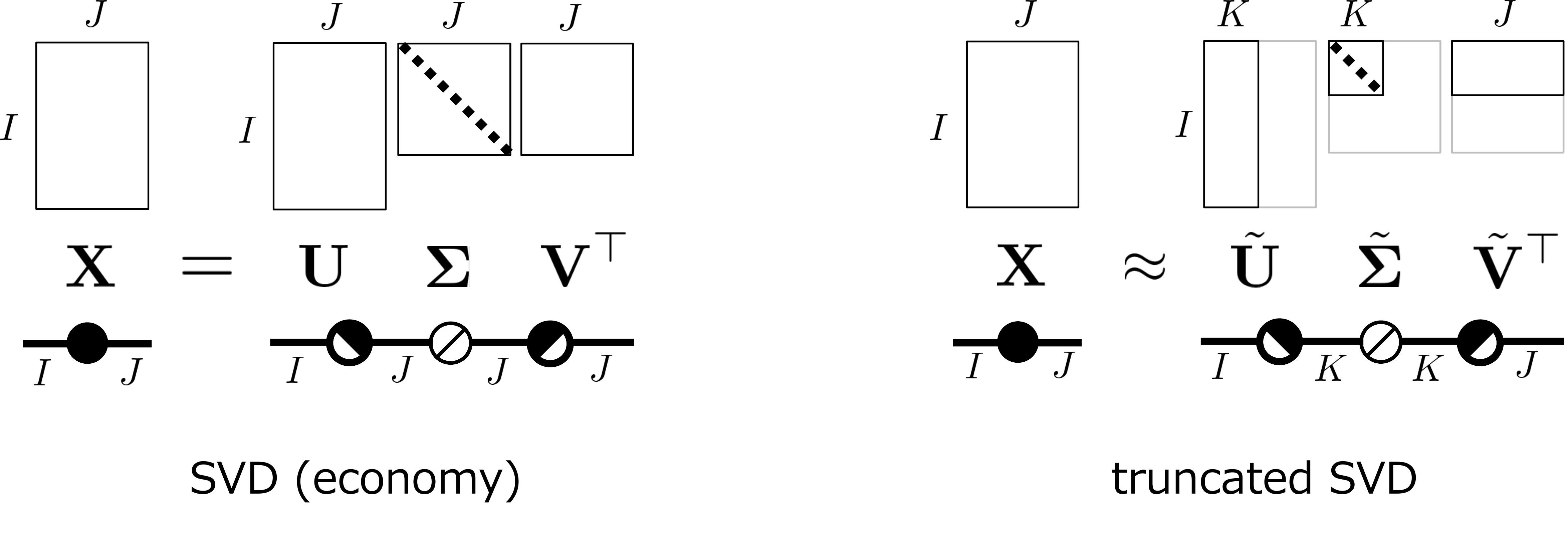}
    \caption{SVD and truncated SVD of a tall matrix ($I > J$).}\label{fig:SVD}
\end{figure}

\paragraph{Best rank-$K$ approximation}
{\em Best rank-$K$ approximation problem} of a matrix $\mat{X}$ is given by
\begin{align}
\minimize_{\mat{Y}} || \mat{X} - \mat{Y} ||_F^2, \text{ s.t. } \rank(\mat{Y}) = K.
\end{align}
This problem is to find the matrix $\widehat{\mat{Y}}$ that is closest to $\mat{X}$ in a set of rank-$K$ matrices.
The solution of the best rank-$K$ approximation is given by
\begin{align}
\widehat{\mat{Y}} = \sum_{r=1}^K \sigma_r \vect{u}_r \circ \vect{v}_r,
\end{align}
and this fact is known as the {\em Eckart-Young theorem} \cite{eckart1936approximation}.
The discarded components by this approximation are given as
\begin{align}
\mat{X} - \widehat{\mat{Y}} = \sum_{r'=K+1}^R \sigma_{r'} \vect{u}_{r'} \circ \vect{v}_{r'}.
\end{align}
Let $\tilde{\mat{U}} = [\vect{u}_1, ..., \vect{v}_K]$, $\tilde{\mat{V}} = [\vect{v}_1, ..., \vect{v}_K]$, and $\tilde{\mat{\Sigma}} = \diag(\sigma_1, ..., \sigma_K)$, then the approximation is written as
\begin{align}
\mat{X} \approx \mat{Y} = \tilde{\mat{U}}\tilde{\mat{\Sigma}}\tilde{\mat{V}}^\top.
\end{align}
This is called {\em truncated SVD} of $\mat{X}$.
Figure~\ref{fig:SVD} shows illustrations of SVD and truncated SVD.

\begin{figure}[t]
    \centering
    \includegraphics[width=0.9\textwidth]{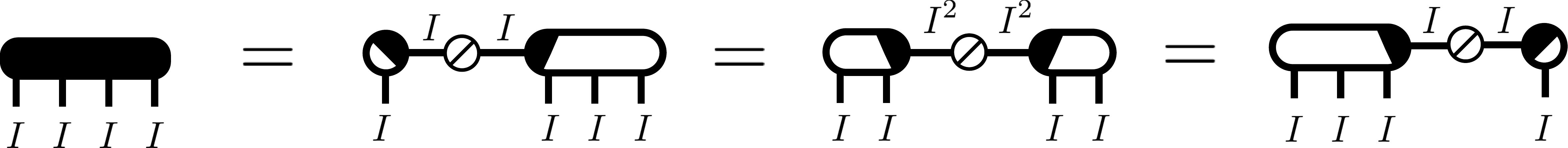}
    \caption{Various forms of SVD for a tensor.}\label{fig:SVD_tensor}
\end{figure}

\paragraph{SVD for tensors}
In a way similar to QR decomposition, SVD can also be applied to tensors through matricization.
In the form of mode-$n$ matricization, SVD of a tensor $\ten{X} \in \bbR{I_1 \times \cdots \times I_N} $ can be written by
\begin{align}
&\mat{X}_{(n)}^\top = \mat{U}_{(n)}^\top \mat{\Sigma} \mat{V}^\top \\
&\Leftrightarrow \ten{X} = \ten{U} \mathbin{_{n}\times^{1}} \mat{\Sigma} \mat{V}^\top,
\end{align}
where $\ten{U} \in \bbR{I_1 \times \cdots \times I_N}$, $\mat{\Sigma} \in \bbR{I_n \times I_n}$, and $\mat{V} \in \bbR{I_n \times I_n}$.
The left and right matrices have orthogonality $\mat{U}_{(n)}\mat{U}_{(n)}^\top = \mat{I}$ and $\mat{V}^\top \mat{V} = \mat{I}$.
In general, we have $|| \ten{X} ||_F = || \mat{\Sigma} ||_F$.
Since it is arbitrary whether each of the $N$ modes is assigned to the left or right matrix, various forms of SVD can be considered (see Figure~\ref{fig:SVD_tensor}).

\subsection{CP decomposition}\label{sec:CPD}
{\em Canonical polyadic (CP) decomposition} \cite{hitchcock1927expression,carroll1970analysis,harshman1970foundations} for a third-order tensor $\ten{X} \in \bbR{I \times J \times K}$ is given by
\begin{align}
  \ten{X} = \sum_{r=1}^R \lambda_r \vect a_r \circ \vect b_r \circ \vect c_r.\label{eq:CPD_sum_of_outer_product}
\end{align} 
Let $\ten{D} \in \bbR{R \times R \times R}$ be a super-diagonal tensor with diagonal entries $\lambda_1$, ..., $\lambda_R$ as
\begin{align}
\mathcal{D}_{pqr} = \left\{\begin{array}{ll} \lambda_r & (p=q=r) \\ 0 & \text{otherwise} \end{array}\right.,
\end{align}
and $\mat{A} = [\vect{a}_1, ..., \vect{a}_R] \in \bbR{I \times R}$, $\mat{B} = [\vect{b}_1, ..., \vect{b}_R] \in \bbR{J \times R}$, and $\mat{C} = [\vect{c}_1, ..., \vect{c}_R] \in \bbR{K \times R}$,
then the CP decomposition can be written by using all-mode product as
\begin{align}
\ten{X} &= \ten{D} \times_1 \mat A \times_2 \mat B \times_3 \mat C \notag \\
 &= \llbracket \ten{D} ; \mat A, \mat B, \mat C \rrbracket.
\end{align}
In addition, let $\ten{I} \in \bbR{R \times R \times R \times R}$ be a super-diagonal tensor which all diagonal entries are 1 and $\bm\lambda = [\lambda_1, ..., \lambda_R]^\top \in \bbR{R}$, then we have $\ten{D} = \ten{I} \times_4 \bm\lambda^\top$ and
\begin{align}
\ten{X} = \ten{I} \times_1 \mat A \times_2 \mat B \times_3 \mat C \times_4 \bm\lambda^\top.
\end{align}

\paragraph{CP-rank of a tensor}
Based on the form of SVD in \eqref{eq:SVD_sum_of_outer_product} and the form of CP decomposition in \eqref{eq:CPD_sum_of_outer_product}, the CP decomposition can be understood as a straightforward extension of ``sum of rank-1 matrices'' to ``sum of rank-1 tensors''.
In a way similar to the rank of a matrix, {\em CP-rank} of a tensor is refers to
the minimum number of rank-1 tensors needed to represent the tensor as a sum of rank-1 tensors.

\begin{figure}[t]
    \centering
    \includegraphics[width=0.8\textwidth]{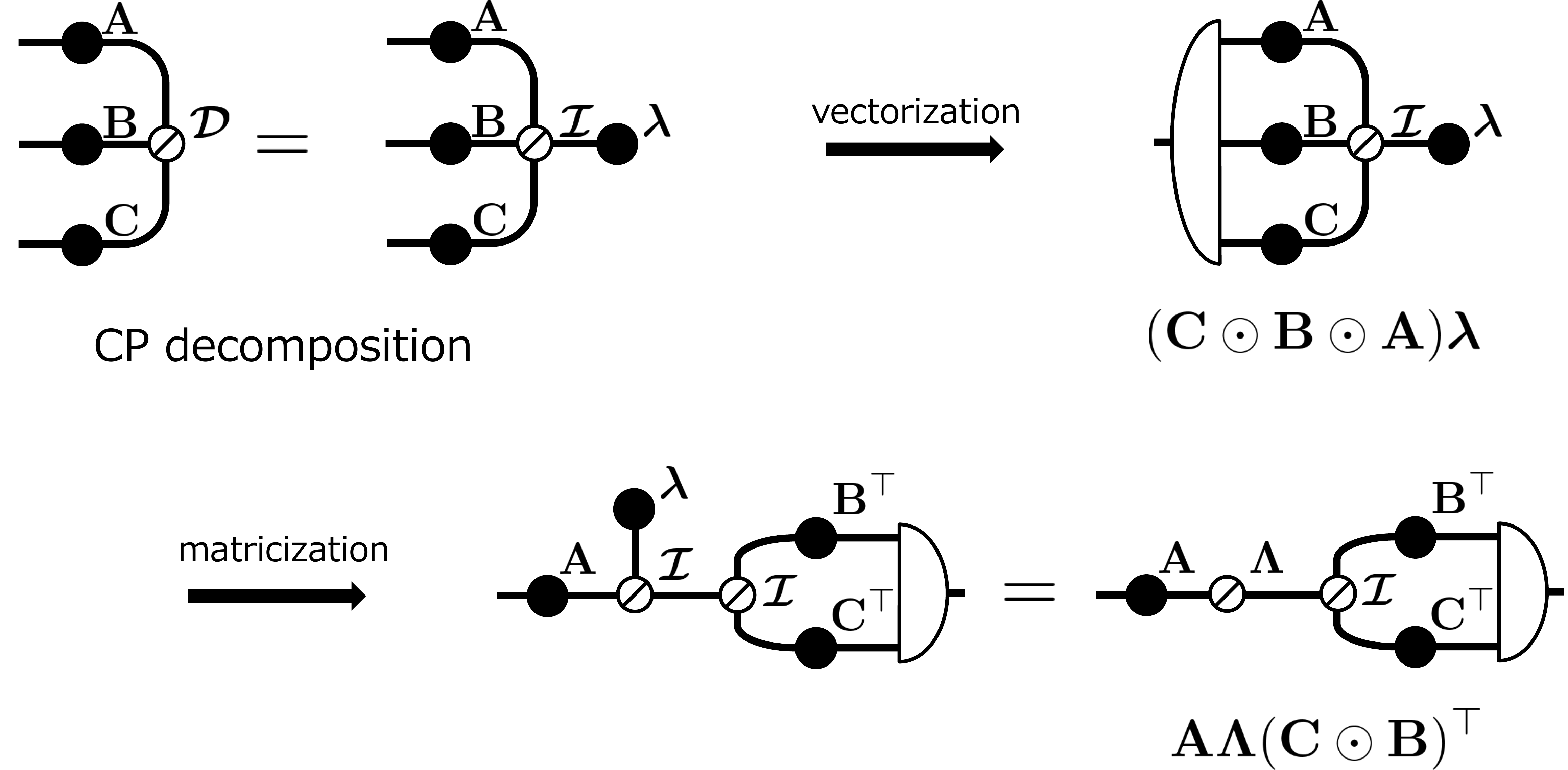}
    \caption{CP decomposition and its unfolding.}\label{fig:CPD_various}
\end{figure}

\paragraph{Unfolding of CP decomposition}
Forms of vectorization and matricization are given by
\begin{align}
  \unvec(\ten{X}) &= (\mat{C} \odot \mat{B} \odot \mat{A}) \bm\lambda, \\
  \mat{X}_{(1)} &= \mat{A} \bm\Lambda (\mat{C} \odot \mat{B})^\top, \\
  \mat{X}_{(2)} &= \mat{B} \bm\Lambda (\mat{C} \odot \mat{A})^\top, \\
  \mat{X}_{(3)} &= \mat{C} \bm\Lambda (\mat{B} \odot \mat{A})^\top,
\end{align}
where $\bm\Lambda = \diag(\bm\lambda) \in \bbR{R \times R}$.
Figure~\ref{fig:CPD_various} shows diagrams in various forms of CP decomposition.
This diagram should make sense on the basis of discussions about super-diagonal tensors in Section~\ref{sec:Hadamard_prod_superdiagonal}.
It also shows that the Khatri-Rao product can be represented using a super-diagonal tensor in graphical notation.

\subsection{Tucker decomposition}
{\em Tucker decomposition} (TKD) \cite{tucker1963implications,tucker1964extension,tucker1966some} for a third-order tensor $\ten{X} \in \bbR{I\times J \times K}$ is given by
\begin{align}
  \ten{X} &= \sum_{p = 1}^P \sum_{q = 1}^Q \sum_{r = 1}^R \mathcal{G}_{pqr} \vect a_p \circ \vect b_q \circ \vect c_r \notag \\ 
  &= \ten{G} \times_1 \mat{A} \times_2 \mat{B} \times_3 \mat{C} \label{eq:TKD} \\
  &= \llbracket \ten{G}; \mat{A}, \mat{B}, \mat{C} \rrbracket,\notag
\end{align}
where $\ten{G} \in \bbR{P \times Q \times R}$ is called a {\em core tensor}, and $\mat{A} = [\vect{a}_1, ..., \vect{a}_P] \in \bbR{I \times P}$, $\mat{B}= [\vect{b}_1, ..., \vect{b}_Q] \in \bbR{J \times Q}$ and $\mat{C} = [\vect{c}_1, ..., \vect{c}_K]\in \bbR{K \times R}$ are called {\em factor matrices}.
From \eqref{eq:TKD}, the TKD is represented as a linear combination of rank-1 tensors $\vect a_p \circ \vect b_q \circ \vect c_r $.
We say that {\em Tucker-rank} is $(P,Q,R)$ in \eqref{eq:TKD} when the minimum size of core tensor is $P \times Q \times R$.
When the core tensor $\ten{G}$ is super-diagonal, Tucker decomposition is equivalent to CP decomposition.

\begin{figure}[t]
  \centering
  \includegraphics[width=0.90\textwidth]{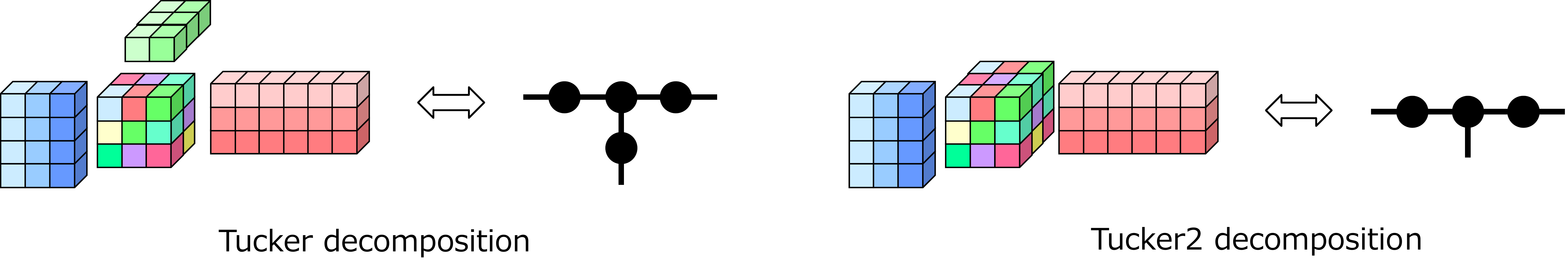}
  \caption{Tucker decompositions.}\label{fig:TKD}
\end{figure}

\paragraph{Tucker2 decomposition}
When we assume $N=3$ and only two factor matrices as
\begin{align}
  \ten{X} = \ten{G} \times_1 \mat{A} \times_2 \mat{B},
\end{align}
it called {\em Tucker2 decomposition}.
Tucker decomposition with $\mat C = \mat I$ is equivalent to Tucker2 decomposition.
Each slice of the core tensor in Tucker2 $\ten{G}(:,:,k)$ can be regarded as a low-dimensional code of the original slice $\ten{X}(:,:,k)$.
Figure~\ref{fig:TKD} shows the diagram for Tucker decompositions.

\paragraph{Unfolding of TKD}
Since Tucker decomposition is mathematically equivalent to the all-mode product, its unfolding is also the same as shown in Figure~\ref{fig:all_mode_unfold}.
Vectorization and matricization of Tucker decomposition can be given by
\begin{align}
  \unvec(\ten{X}) &= (\mat{C} \otimes \mat{B} \otimes \mat{A}) \unvec(\ten{G}), \\
  \mat X_{(1)} &= \mat{A} \mat{G}_{(1)} (\mat{C} \otimes \mat{B})^\top, \\
  \mat X_{(2)} &= \mat{B} \mat{G}_{(2)} (\mat{C} \otimes \mat{A})^\top, \\
  \mat X_{(3)} &= \mat{C} \mat{G}_{(3)} (\mat{B} \otimes \mat{A})^\top. 
\end{align}

\paragraph{TKD for $N$th-order tensors}
Tucker decomposition for the $N$th-order tensor $\ten{Y} \in \bbR{I_1 \times I_2 \times \cdots \times I_N}$ is given by
\begin{align}
  \ten{Y}  = \ten{G} \times \{ \mat{A} \} = \ten{G} \times_1 \mat{A}^{(1)} \times_2 \mat{A}^{(2)}\cdots \times_N \mat{A}^{(N)}, \label{eq:TKD_N}
\end{align}
where $\ten{G} \in \bbR{R_1 \times R_2 \times \cdots \times R_N}$ is a core tensor and $\mat{A}^{(n)} \in \bbR{I_n \times R_n}$ are factor matrices of which the sizes are $R_n \leq I_n$.

\paragraph{Tucker rank of a tensor}
The minimum size of the core tensor $\ten{G}$ that satisfies \eqref{eq:TKD_N} is called the {\em Tucker rank} or {\em multilinear tensor rank}.

The CP rank is an integer scalar, while the Tucker rank is an integer $N$-dimensional vector.
The $n$-th entry of the Tucker rank is called the {\em $n$-rank} or {\em mode rank}.
The $n$-rank of a tensor $\ten{Y}$ is equivalent to the matrix rank of its mode-$n$ matricization as
\begin{align}
\rank_{n}(\ten{Y}) = \rank(\mat{Y}_{(n)}) \leq I_n.
\end{align}
This means that the $n$-rank is the dimension of the space spanned by all fibers along the $n$-th mode of the tensor $\ten{Y}$.
At the same time, we can say that each factor matrix $\mat{A}^{(n)}$ in the Tucker decomposition is a basis matrix for representing all fibers along the $n$-th mode.

\paragraph{Orthogonalization of TKD}\label{sec:uniqueness_TKD}
Tucker decomposition is not unique in general.
Different Tucker decompositions can be obtained using any regular matrices $\mat{U} \in \bbR{P \times P}$, $\mat{V} \in \bbR{Q \times Q}$, and $\mat{W} \in \bbR{R \times R}$ as follows:
\begin{align}
\llbracket \ten{G}; \mat{A}, \mat{B}, \mat{C} \rrbracket = \llbracket \ten{G} \times_1 \mat{U} \times_2 \mat{V} \times_3 \mat{W}; \mat{A} \mat{U}^{-1}, \mat{B} \mat{V}^{-1}, \mat{C} \mat{W}^{-1} \rrbracket
\end{align}
Since $\mat{A} \neq \mat{A} \mat{U}^{-1}$, $\mat{B} \neq \mat{B} \mat{V}^{-1}$, and $\mat{C} \neq \mat{C} \mat{W}^{-1}$, which can be said to be a different Tucker decomposition.

\begin{figure}[t]
  \centering
  \includegraphics[width=0.90\textwidth]{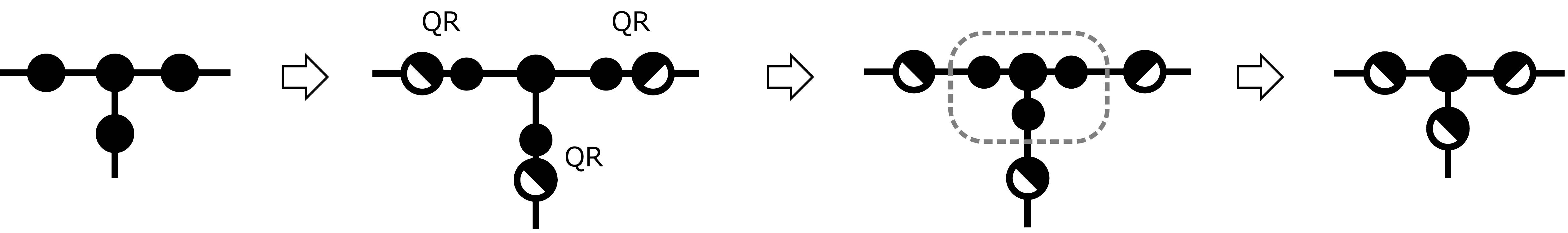}
  \caption{Orthogonalization of TKD.}\label{fig:TKD_orth}
\end{figure}

Any Tucker decomposition can be transformed into a Tucker decomposition in which each factor matrix is a column orthogonal matrix as follows:
\begin{align}
  \mat A^\top \mat A = \mat I, \ \mat B^\top \mat B = \mat I, \ \mat C^\top \mat C = \mat I.
\end{align}
This is obtained by QR decomposition of each factor matrix and integrating the upper triangular matrix into the core tensor, as shown in Figure~\ref{fig:TKD_orth}.
This operation is called {\em orthogonalization of TKD}.

\begin{figure}[t]
  \centering
  \includegraphics[width=0.70\textwidth]{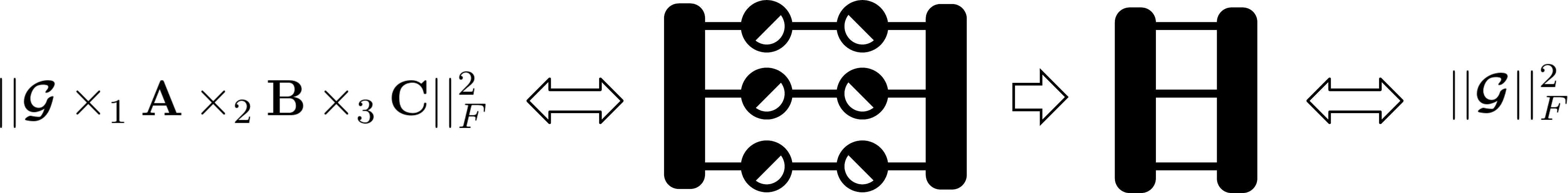}
  \caption{Frobenius norm of orthogonalized TKD.}\label{fig:TKD_norm}
\end{figure}

In addition, Frobenius norm of the orthogonalized TKD is given by only its core tensor as
\begin{align}
  || \ten{G} \times_1 \mat{A} \times_2 \mat{B} \times_3 \mat{C} ||_F^2 &= || (\mat C \otimes \mat B \otimes \mat A) \unvec(\ten{G}) ||_2^2  \notag \\
  &= \unvec(\ten{G})^\top (\mat C \otimes \mat B \otimes \mat A)^\top (\mat C \otimes \mat B \otimes \mat A) \unvec(\ten{G}) \notag \\ 
  &= || \unvec(\ten{G}) ||_2^2 = ||\ten{G}||_F^2. 
\end{align}
Figure~\ref{fig:TKD_norm} shows it in diagrams.

\begin{figure}[t]
  \centering
  \includegraphics[width=0.85\textwidth]{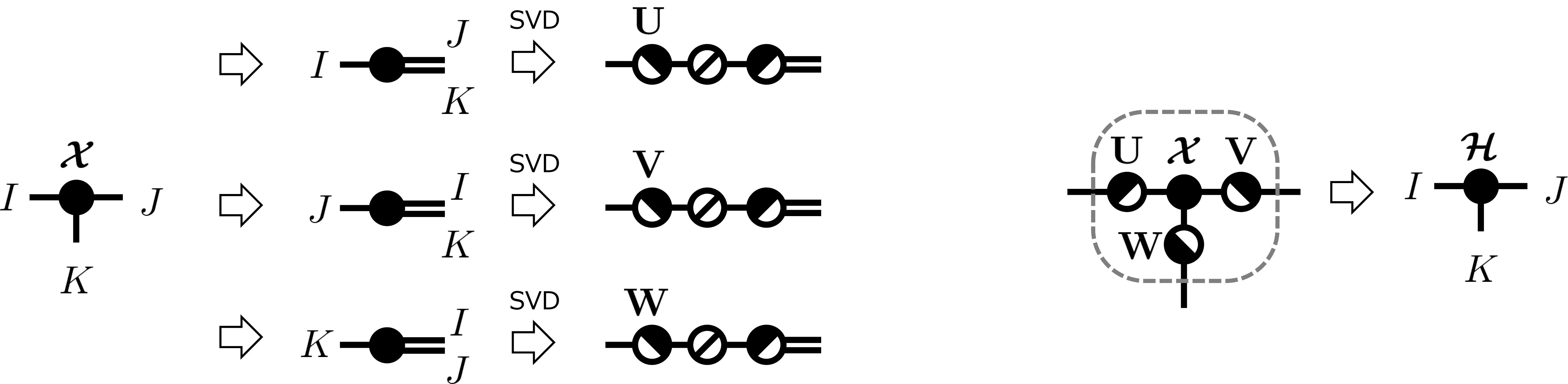}
  \caption{Process for obtaining HOSVD.}\label{fig:HOSVD}
\end{figure}

\paragraph{Higher-order SVD (HOSVD)}
{\em Higher-order singular value decomposition (HOSVD)}\cite{de2000multilinear} is an extension of SVD for tensors.
HOSVD of a tensor $\ten{X} \in \bbR{I \times J \times K}$ is given by
\begin{align}
  &\ten{X} = \ten{H} \times_1 \mat{U} \times_2 \mat{V} \times_3 \mat{W}, \label{eq:hosvd}
\end{align}
where
\begin{align}
  &\mat{U} \in \bbR{I \times I}: \text{left singular vectors of }\mat{X}_{(1)}, \\
  &\mat{V} \in \bbR{J \times J}: \text{left singular vectors of }\mat{X}_{(2)}, \\
  &\mat{W} \in \bbR{K \times K}: \text{left singular vectors of }\mat{X}_{(3)}.
\end{align}
From \eqref{eq:hosvd}, we can see this is a type of Tucker decomposition.
Since the factor matrices are the orthogonal matrices obtained by SVD, we have
\begin{align}
  \mat{U}^\top \mat{U} &=  \mat{U} \mat{U}^\top = \mat{I} \in \bbR{I \times I}, \\
  \mat{V}^\top \mat{V} &=  \mat{V} \mat{V}^\top = \mat{I} \in \bbR{J \times J}, \\
  \mat{W}^\top \mat{W} &=  \mat{W} \mat{W}^\top = \mat{I} \in \bbR{K \times K}.
\end{align}
Then, core tensor $\ten{H}$ can be given by
\begin{align}
  \ten{H} = \ten{X} \times_1 \mat{U}^\top \times_2 \mat{V}^\top \times_3 \mat{W}^\top \in \bbR{I \times J \times K}.
\end{align}
Figure~\ref{fig:HOSVD} shows a process for obtaining HOSVD in diagrams.

\paragraph{All orthogonality}
The core tensor of HOSVD has an interesting property called {\em all orghogonality}.
A tensor $\ten{H}$ is said to be {\em all orthogonal} if the following holds:
\begin{align}
  &\mat{H}_{(1)} \mat{H}_{(1)}^\top = \diag(\bm\lambda_1) \in \bbR{I \times I}, \\
  &\mat{H}_{(2)} \mat{H}_{(2)}^\top = \diag(\bm\lambda_2) \in \bbR{J \times J}, \\
  &\mat{H}_{(3)} \mat{H}_{(3)}^\top = \diag(\bm\lambda_3) \in \bbR{K \times K}.
\end{align}
That is, the autocorrelation matrix of its matricization is diagonal for all modes.
It can be verified by
\begin{align}
  \mat{H}_{(1)} \mat{H}_{(1)}^\top &= \mat{U}^\top \mat{X}_{(1)} (\mat{W} \otimes \mat{V})(\mat{W} \otimes \mat{V})^\top \mat{X}_{(1)}^\top \mat{U} \notag \\
  &= \mat{U}^\top \mat{X}_{(1)} \mat{X}_{(1)}^\top \mat{U} \notag \\
  &= \diag(\bm\lambda_1),
\end{align}
where entries of $\bm\lambda_1$ are eigenvalues of $\mat{X}_{(1)} \mat{X}_{(1)}^\top$.

\paragraph{Truncated HOSVD}
Since the eigenvalues are arranged in descending order, we can expect that the absolute value of each element of the core tensor (the magnitude of the contribution of the $(i,j,k)$ rank 1 component) decreases from $\mathcal{H}_{111}$ to $\mathcal{H}_{IJK}$.
From this, {\em truncated HOSVD} can be considered for approximating the tensor by
\begin{align}
\ten{X} \approx \ten{G} \times_1 \mat{A} \times_2 \mat{B} \times_3 \mat{C},
\end{align}
where $P < I$, $Q < J$, and $R < K$, and 
\begin{align}
  \mat{A} &= \mat{U}(:,1:P) = \mat{U}_{:,1:P} \in \bbR{I \times P},\\
  \mat{B} &= \mat{V}(:,1:Q) = \mat{V}_{:,1:Q} \in \bbR{J \times Q},\\
  \mat{C} &= \mat{W}(:,1:R) = \mat{W}_{:,1:R} \in \bbR{K \times R},\\
  \ten{G} &= \ten{H}(1:P,1:Q,1:R) = \ten{H}_{1:P,1:Q,1:R} \in \bbR{P \times Q \times R}.
\end{align}

\begin{figure}[t]
  \centering
  \includegraphics[width=0.8\textwidth]{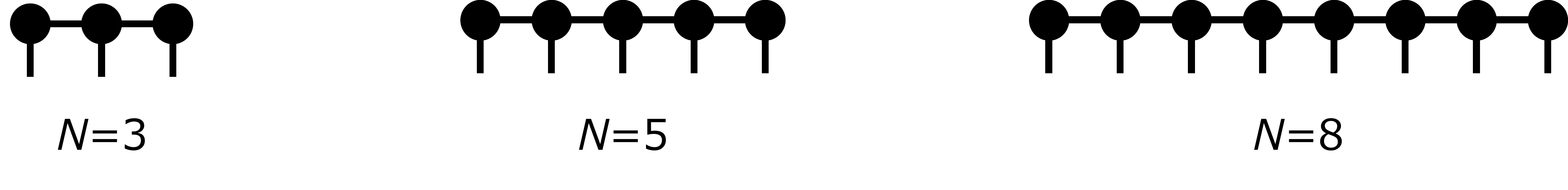}
  \caption{TT decomposition.}\label{fig:TTD}
\end{figure}

\subsection{Tensor-train decomposition}\label{sec:TT-decomposition}
{\em Tensor-train (TT) decomposition} \cite{oseledets2011tensor} is a tensor decomposition model that represents higher-order tensors by chaining second- or third-order tensors.
In tensor networks used in physics, it is also known as {\em matrix product states (MPS)} \cite{ran2020tensor,xiang2023density}.
Figure~\ref{fig:TTD} shows the decompositions of TT in diagrams.

TT decomposition of $N$th-order tensor $\ten{X} \in \bbR{I_1 \times I_2 \times \cdots \times I_N}$ is written by
\begin{align}
  \ten{X} = \llangle \ten{G}_1, \ten{G}_2, ..., \ten{G}_N \rrangle,
\end{align}
where $\{\ten{G}_1, \ten{G}_2, ..., \ten{G}_N \}$ are called {\em TT-cores} and their sizes are 
\begin{align}
  &\ten{G}_1 \in \bbR{R_0 \times I_1 \times R_1}, \\
  &\ten{G}_2 \in \bbR{R_1 \times I_2 \times R_2}, \\
  &\ \ \ \ \vdots \notag \\
  &\ten{G}_N \in \bbR{R_{N-1} \times I_N \times R_N}. 
\end{align}
The sizes $(R_0, ..., R_N)$ are called {\em TT rank} or {\em bond dimension}.
Both ends of the TT rank are usually $R_0=R_N=1$ although we write them as variables for convenience.
In other words, both ends of the TT core are matrices (second-order tensors).

Each entry of the TT decomposition is given by matrix product as
\begin{align}
  \mathcal{X}(i_1, i_2, ..., i_N) 
  &= \sum_{r_1, ..., r_{N-1}}^{R_1, ..., R_{N-1}} \mathcal{G}_1(1, i_1, r_1) \mathcal{G}_2(r_1, i_2, r_2) \cdots \mathcal{G}_N(r_{N-1}, i_N, 1) \notag \\
  &= \ten{G}_1(1,i_1,:)^\top \ten{G}_2(:, i_2, :) \cdots \ten{G}_N(:,i_N,1).
\end{align}
The sizes of each matrix from left to right are $(1,R_1)$, $(R_1,R_2)$, ..., $(R_{N-1},1)$.
When $N=2$, it is equivalent to matrix decomposition, and when $N=3$, it is equivalent to Tucker2 decomposition.
When $R_k$ are all 1, it is a rank 1 decomposition.

\paragraph{Orthogonalization of TT decomposition}
TT decomposition is not unique in general.
For any regular matrix $\mat{H} \in \bbR{R_1 \times R_1}$, we put two TT-cores as follow:
\begin{align}
  \ten{G}'_1 &= \llangle \ten{G}_1, \mat{H} \rrangle,  \\
  \ten{G}'_2 &= \llangle \mat{H}^{-1}, \ten{G}_2 \rrangle.
\end{align}
Then the following TT decompositions are different:
\begin{align}
  \llangle \ten{G}'_1, \ten{G}'_2, ..., \ten{G}_N \rrangle = \llangle \ten{G}_1, \ten{G}_2, ..., \ten{G}_N \rrangle,
\end{align}
where $\ten{G}'_1 \neq \ten{G}_1$ and $\ten{G}'_2 \neq \ten{G}_2$.

\begin{figure}[t]
  \centering
  \includegraphics[width=0.6\textwidth]{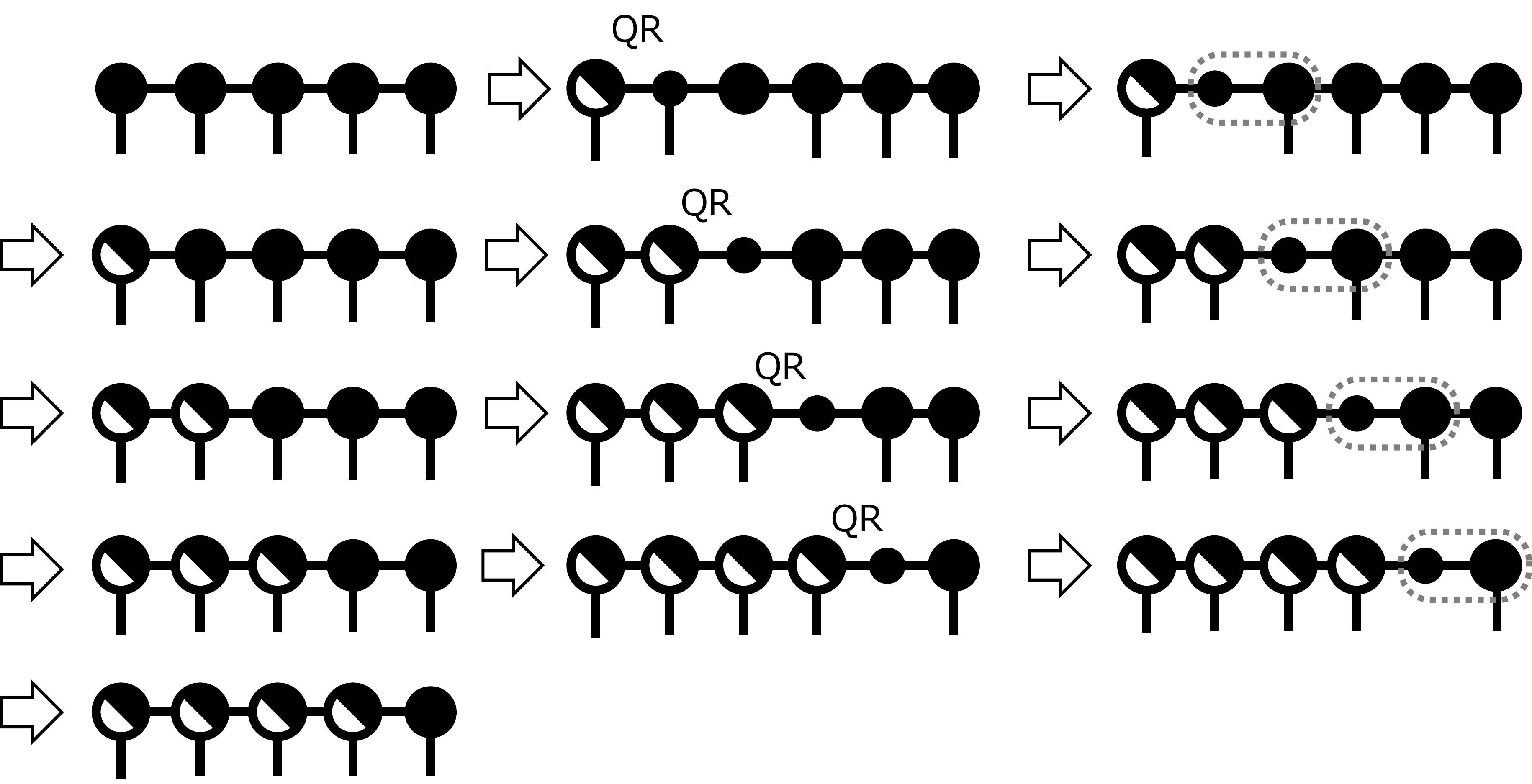}
  \caption{Orthogonalization of TT decomposition using QR decomposition.}\label{fig:TTorth}
\end{figure}

For any TT decomposition, all but one of the TT cores can be orthogonalized.
First any third-order tensor $\ten{G} \in \bbR{P \times Q \times R}$ can be decomposed as follows:
\begin{align}
\ten{G} = \llangle \ten{U}, \mat{R} \rrangle
\end{align}
using QR decomposition.
Then, the orthogonalized TT decomposition can be obtained by 
\begin{align}
  \ten{X} &= \llangle \ten{G}_1, \ten{G}_2, \ten{G}_3, ..., \ten{G}_N \rrangle \notag \\
    &= \llangle \ten{U}_1, \mat{R}_1, \ten{G}_2, \ten{G}_3, ..., \ten{G}_N \rrangle \notag \\
    &= \llangle \ten{U}_1, \llangle \mat{R}_1, \ten{G}_2 \rrangle, \ten{G}_3, ..., \ten{G}_N \rrangle \notag \\
    &= \llangle \ten{U}_1, \ten{G}_2', \ten{G}_3, ..., \ten{G}_N \rrangle \notag \\
    &= \llangle \ten{U}_1, \ten{U}_2, \mat{R}_2, \ten{G}_3, ..., \ten{G}_N \rrangle \notag \\
    &\ \ \ \vdots \notag \\
    &= \llangle \ten{U}_1, ..., \ten{U}_{N-1}, \ten{G}_N' \rrangle. 
\end{align}
Figure~\ref{fig:TTorth} shows the algorithm of this orthogonalization.

\begin{figure}[t]
  \centering
  \vspace{5mm}
  \includegraphics[width=0.6\textwidth]{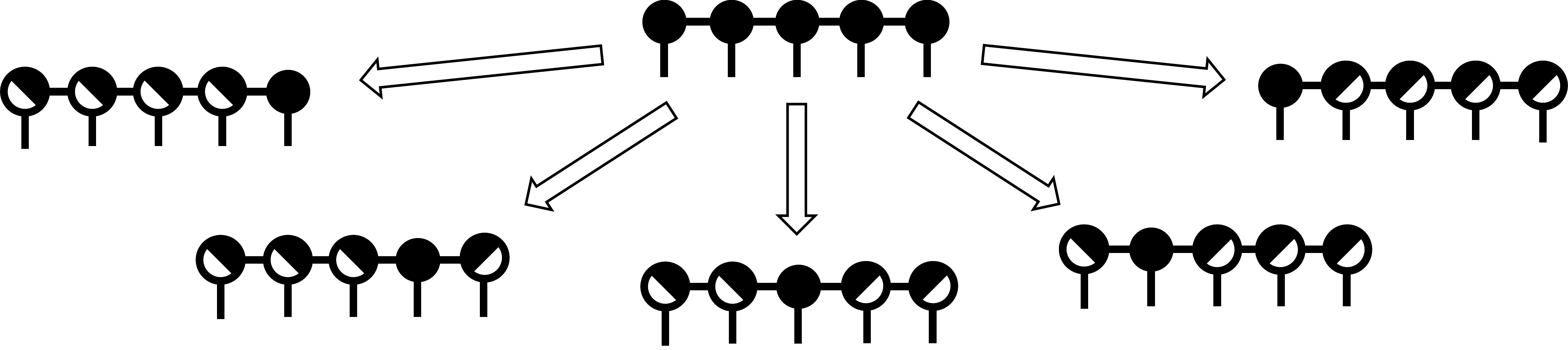}
  \caption{Multiple ways of orthogonalizing TT decompositions.}\label{fig:TTorthvar}
\end{figure}

In general, various ways of orthogonalization can be considered, such as when the right side is orthogonal, or when both the left and right sides are orthogonal, as shown in Figure~\ref{fig:TTorthvar}.

\begin{figure}[t]
  \centering
  \includegraphics[width=0.99\textwidth]{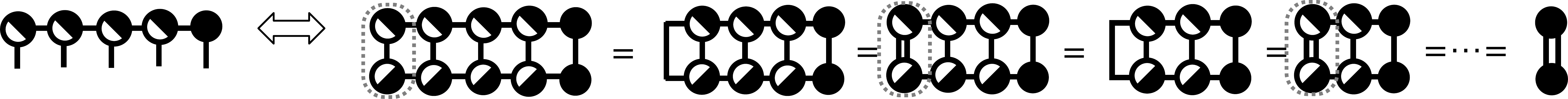}
  \caption{Frobenius norm of orthogonalized TT decomposition.}\label{fig:right_orth_TT}
\end{figure}

\paragraph{Frobenius norm of orthogonalized TT decomposition}
Let us consider the following orthogonalized TT decomposition:
\begin{align}
  \ten{X} = \llangle \ten{U}_1, ..., \ten{U}_{N-1}, \ten{G}_N \rrangle.
\end{align}
Its Frobenius norm depends only on $\ten{G}_N$ as
\begin{align}
  || \ten{X} ||_F^2 &= \langle \llangle \ten{U}_1, ..., \ten{U}_{N-1}, \ten{G}_N \rrangle, \llangle \ten{U}_1, ..., \ten{U}_{N-1}, \ten{G}_N \rrangle \rangle \notag \\
  &= \langle \llangle \ten{U}_2, ...,  \ten{U}_{N-1}, \ten{G}_{N} \rrangle, \llangle \ten{U}_2, ..., \ten{U}_{N-1}, \ten{G}_{N} \rrangle \rangle \notag \\
  &\ \ \ \vdots \notag \\
  &= \langle \llangle \ten{U}_{N-1}, \ten{G}_N \rrangle, \llangle \ten{U}_{N-1}, \ten{G}_N \rrangle \rangle \notag \\
&= \langle \ten{G}_N, \ten{G}_N \rangle = || \ten{G}_N ||_F^2.
\end{align}
The above formulation can be easily understood by a diagram as shown in Figure~\ref{fig:right_orth_TT}.

\begin{figure}[t]
  \centering
  \includegraphics[width=0.7\textwidth]{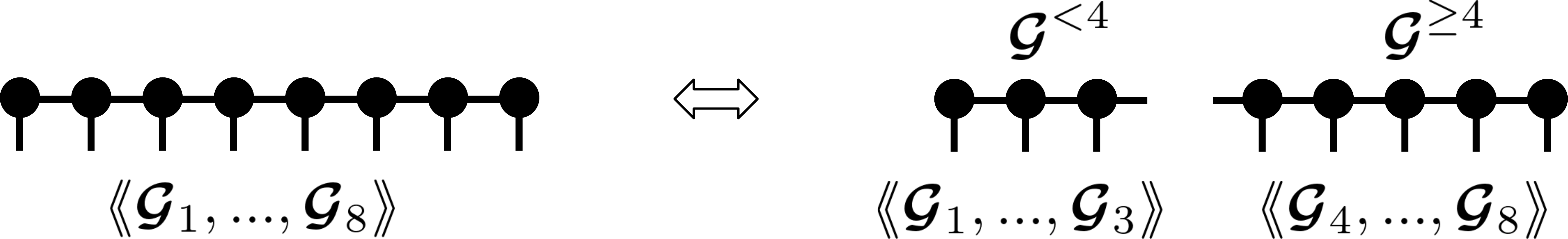}
  \caption{Sub-train.}\label{fig:sub_train}
\end{figure}

\paragraph{Sub-train}
TT decomposition $\llangle \ten{G}_1, ..., \ten{G}_N \rrangle $ can be separated into 
\begin{align}
  \ten{G}^{<k} &= \llangle \ten{G}_1, ..., \ten{G}_{k-1} \rrangle \in \bbR{R_0 \times I_1 \times \cdots \times I_{k-1} \times R_{k-1}}, \\
  \ten{G}^{\geq k} &= \llangle \ten{G}_k, ..., \ten{G}_{N} \rrangle \in \bbR{R_{k-1} \times I_k \times \cdots \times I_{N} \times R_{N}},
\end{align}
where $1 < k < N$.
These are called {\em sub-trains} of TT decomposition.
Figure~\ref{fig:sub_train} shows sub-trains in diagrams.
Cleary, we have
\begin{align}
  \llangle \ten{G}^{<k}, \ten{G}^{\geq k} \rrangle = \llangle \ten{G}_1, ..., \ten{G}_{k-1}, \ten{G}_{k}, ..., \ten{G}_N \rrangle=\llangle \ten{G}_1, ..., \ten{G}_N \rrangle.
\end{align}

\paragraph{TT rank}
For $\ten{A} = \llangle \ten{G}_1, \ten{G}_2, ..., \ten{G}_N \rrangle$, the sizes $(R_0, ..., R_N)$ of the TT core $\ten{G} \in \bbR{R_{n-1} \times I_n \times R_n}$ are called the {\em TT rank}.
If each TT core and sub-train is not rank-deficient, its TT-rank satisfies
\begin{align}
  \rank(\mat{A}_{\ang{k}}) = R_k, \label{eq:TTrank}
\end{align}
where $\mat{A}_{\ang{k}}$ is $k$-unfolding of $\ten{A}$.
This property is important when obtaining the TT decomposition from a tensor using SVD.

Considering sub-trains would be helpful to understand the property of TT rank \eqref{eq:TTrank}.
Let us consider sub-trains as
\begin{align}
  &\ten{U} = \llangle \ten{G}_1, ..., \ten{G}_k \rrangle \in \bbR{I_1 \times \cdots \times I_k \times R_k}, \\
  &\ten{V} = \llangle \ten{G}_{k+1}, ..., \ten{G}_N \rrangle \in \bbR{R_k \times I_{k+1} \times \cdots \times I_N}, 
\end{align}
then we have
\begin{align}
  &\mathcal{A}(i_1,...,i_N) = \sum_{r=1}^{R_k} \mathcal{U}(i_1, ..., i_k, r) \mathcal{V}(r, i_{k+1}, ..., i_N) \\
  &\Leftrightarrow \mat{A}_{\ang{k}} = \mat{U}_{\ang{k}} \mat{V}_{\ang{1}}.
\end{align}
From the size of matrices $\mat{U}_{\ang{k}} \in \bbR{(I_1 \cdots I_k) \times R_k}$ and $\mat{V}_{\ang{1}} \in \bbR{R_k \times (I_{k+1} \cdots I_N)}$, clearly $\rank(\mat{A}_{\ang{k}}) \leq R_k$.

\paragraph{TT-SVD}
{\em Tensor-train singular value decomposition (TT-SVD)}\cite{oseledets2011tensor} is an algorithm to obtain TT decomposition of a tensor $\ten{A} \in \bbR{I_1 \times I_2 \times \cdots \times I_N}$ as
\begin{align}
\ten{A} = \llangle \ten{G}_1, \ten{G}_2, ..., \ten{G}_N \rrangle. \label{eq:exact_TT}
\end{align}
To achieve \eqref{eq:exact_TT}, we set the TT rank by
\begin{align}
  R_k \leftarrow \rank(\mat{A}_{\ang{k}} ).
\end{align}

\begin{figure}[t]
  \centering
  \includegraphics[width=0.99\textwidth]{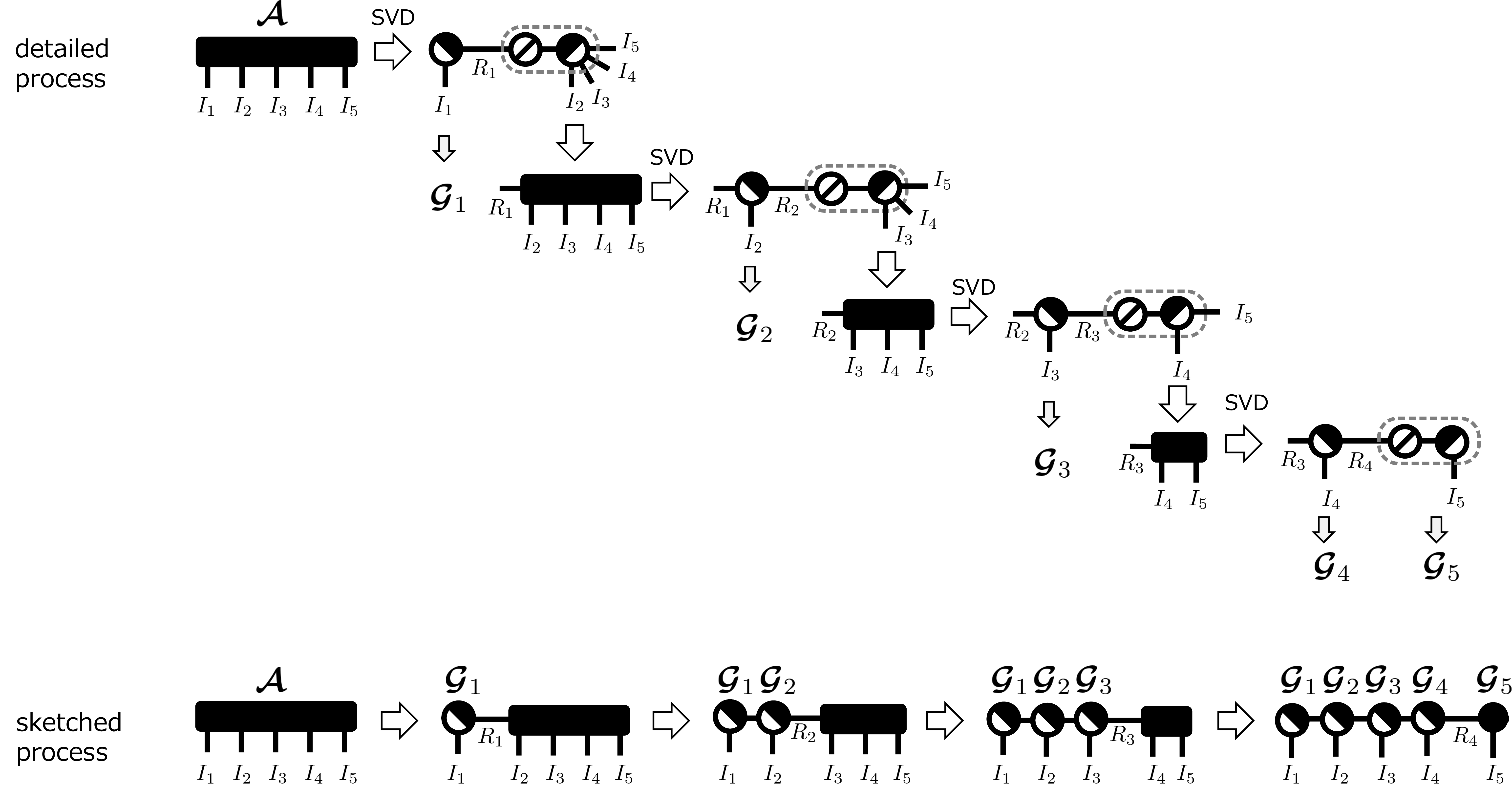}
  \caption{Process of TT-SVD.}\label{fig:TTSVD}
\end{figure}

Figure~\ref{fig:TTSVD} shows a process of TT-SVD that repeats to apply SVD to the remaining tensor.
First, we consider SVD of $\mat{A}_{\ang{1}}$ as
\begin{align}
  &\mat{A}_{\ang{1}} \mathop{\longrightarrow}^{\text{{\tiny SVD}}} \mat{U} \mat{\Sigma} \mat{V}_{\ang{1}} = \mat{U} {\mat{B}}_{\ang{1}} \\
  &\Leftrightarrow \ten{A} \mathop{\longrightarrow}^{\text{{\tiny SVD}}} \llangle \mat{U}, \mat{\Sigma}, \ten{V} \rrangle =\llangle \mat{U}, \ten{B} \rrangle, 
\end{align}
where $\mat{B}_{\ang{1}} =  \mat{\Sigma} \mat{V}_{\ang{1}}$ and $\ten{B} = \llangle \mat{\Sigma}, \ten{V} \rrangle$.
Then, the first TT core is obtained by
\begin{align}
\ten{G}_1 \leftarrow \mat{U},
\end{align}
and $\ten{B}$ is obtained as the remaining tensor.
Second, we consider SVD of $\mat{B}_{\ang{2}}$ as
\begin{align}
  &\mat{B}_{\ang{2}} \mathop{\longrightarrow}^{\text{{\tiny SVD}}} \mat{U}_{\ang{2}} \mat{\Sigma} \mat{V}_{\ang{1}} = \mat{U}_{\ang{2}} {\mat{C}}_{\ang{1}} \\
  &\Leftrightarrow \ten{B} \mathop{\longrightarrow}^{\text{{\tiny SVD}}} \llangle \ten{U}, \mat{\Sigma}, \ten{V} \rrangle =\llangle \ten{U}, \ten{C} \rrangle, 
\end{align}
where $\mat{C}_{\ang{1}} =  \mat{\Sigma} \mat{V}_{\ang{1}}$ and $\ten{C} = \llangle \mat{\Sigma}, \ten{V} \rrangle$.
Then, the second TT core is obtained by
\begin{align}
\ten{G}_2 \leftarrow \ten{U},
\end{align}
and $\ten{C}$ is obtained as the remaining tensor.
In a similar way, $\ten{G}_3$, ..., $\ten{G}_N$ can be sequentially obtained by using SVD.

When $R_k$ is set to less than $rank(\mat{A}_{\ang{k}})$ and replace SVD with truncated SVD with rank $R_k$, an error will occur and the result is called {\em truncated TT-SVD}.

\begin{figure}[t]
  \centering
  \includegraphics[width=0.85\textwidth]{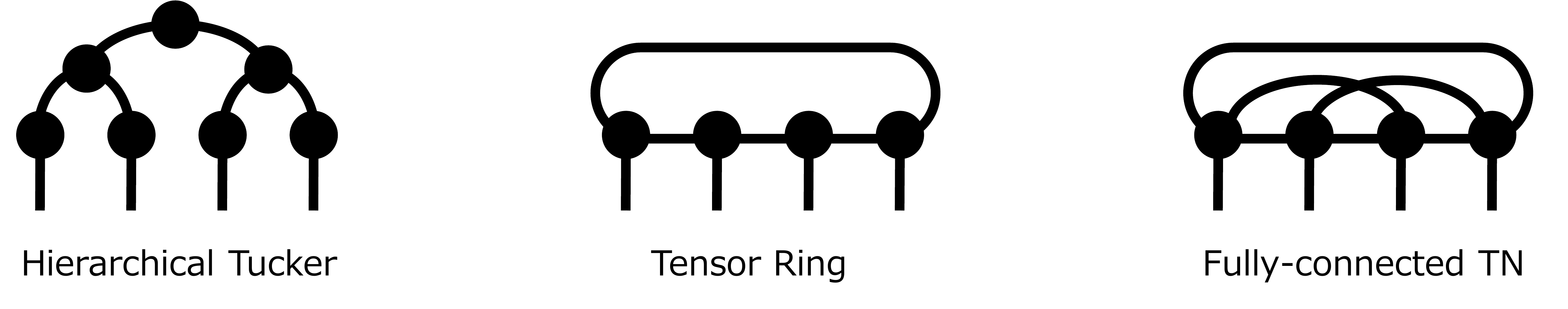}
  \caption{Tensor networks.}\label{fig:TNDs}
\end{figure}

\subsection{Other variants}
Here we briefly introduce other tensor decompositions: {\em hierarchical Tucker decomposition}, {\em tensor-ring decomposition}, and {\em fully-connected tensor network decomposition}.
Figure~\ref{fig:TNDs} shows diagrams for these tensor decompositions.

\paragraph{Hierarchical Tucker decomposition}
{\em Hierarchical Tucker decomposition} \cite{grasedyck2010hierarchical} is a decomposition model that represents a higher-order tensor with a tree-structured tensor network (usually binary tree).
For example, 4 modes of a tensor $\{1,2,3,4\}$ are separated into $\{1,2\}$ and $\{3,4\}$ at the first level.
Next, $\{1,2\}$ are separated into $\{1\}$ and $\{2\}$, and $\{3,4\}$ are separated into $\{3\}$ and $\{4\}$ at the second level.
This mode separation $\{ \{ \{1\}, \{2\} \}, \{ \{3\}, \{4\} \} \}$ can be represented as a binary tree, which corresponds to the tensor network shown as the diagram on the left of Figure~\ref{fig:TNDs}.
In other words, there are as many variations of hierarchical Tucker decompositions as there are binary trees.
TT decomposition is a special case of hierarchical Tucker decomposition with mode separation $\{ 1, \{2, \{ 3, \{ 4 \} \} \} \}$.

\paragraph{Tensor-ring decomposition}
{\em Tensor-ring (TR) decomposition} \cite{zhao2016tensor} is a tensor network in which both ends of the TT decomposition have connections and are composed of one closed loop.
Unlike the TT decomposition, all core tensors are third-order tensors $\{ \ten{G}_n \in \bbR{R_{n-1} \times I_n \times R_n} \}_{n=1}^N$, and $R_0=R_N$.
Each entry of TR decomposition is given by matrix product with trace as
\begin{align}
  \tr (\ten{G}_1(:,i_1,:) \ten{G}_2(:,i_2,:) \cdots \ten{G}_N(:,i_N,:)).
\end{align}
Since the trace is invariant under circular shifts, it has some symmetry with respect to the ordering of the core tensors (or modes).

When the $k$-th TR rank of the tensor ring decomposition $\{ R_1, R_2, ..., R_N \}$ becomes 1 ($R_k = 1$), the connection of that part is cut, and the decomposition becomes a TT decomposition. At this time, the order of the core tensors rotates as $\{\ten{G}_{k+1}, ..., \ten{G}_N, \ten{G}_1, ..., \ten{G}_k\}$. The starting point is $\ten{G}_{k+1}$, and the end point is $\ten{G}_k$.
In this sense, the TT decomposition is a special case of the TR decomposition.

\paragraph{Fully-connected tensor network}
{\em Fully-connected tensor network (FCTN)} \cite{zheng2021fully} has a network structure in which all core tensors are connected to each other.
FCTN neutrally captures all binary relations between modes.
In addition, since connections are cut when the ranks are 1, it can be considered as a generalization of TT decomposition and TR decomposition.
However, hierarchical Tucker is not included because it has intermediate (relay) nodes.

\newpage

\section*{Acknowledgment}
This work was supported in part by the Japan Society for the Promotion of Science (JSPS) KAKENHI under Grant 23K28109.

\bibliographystyle{abbrv}

\end{document}